%% file: neurips_2026.tex
\documentclass{article}

\usepackage[preprint]{neurips_2026}

\usepackage[utf8]{inputenc} % allow utf-8 input
\usepackage[T1]{fontenc}    % use 8-bit T1 fonts
\usepackage{hyperref}       % hyperlinks
\usepackage{url}            % simple URL typesetting
\usepackage{booktabs}       % professional-quality tables
\usepackage{amsfonts}       % blackboard math symbols
\usepackage{nicefrac}       % compact symbols for 1/2, etc.
\usepackage{microtype}      % microtypography
\usepackage{xcolor}         % colors
\usepackage{graphicx}
\usepackage{enumitem}
\usepackage{wrapfig}
\usepackage{comment}

\input{custom_package}

\title{\ours{}: A DSM-Grounded Benchmark for Evaluating Psychiatric Diagnostic Capability of Large Language Models}

\newcommand\CoauthorMark{\footnotemark[\arabic{footnote}]}
\newcommand\CorrMark{\footnotemark[\arabic{footnote}]}
 \author{Hoyun Song$^1$\thanks{Equal Contribution}\hspace{3mm} Migyeong Kang$^2$\CoauthorMark\hspace{3mm} Jisu Shin$^1$\hspace{3mm} Jihyun Kim$^2$\hspace{3mm} Chanbi Park$^3$\\
 \bf Hangyeol Yoo$^4$\hspace{3mm} Jihyun An$^5$\hspace{3mm} Alice Oh$^1$\hspace{3mm} Jinyoung Han$^2$\thanks{Corresponding Authors}\hspace{3mm} KyungTae Lim$^1$\CorrMark\\
 $^1$KAIST \hspace{3mm} $^2$Sungkyunkwan University\\
 $^3$Dongguk University Medical Center\hspace{3mm} $^4$Seoultech\hspace{3mm}$^5$Samsung Medical Center \\
 \texttt{\{hysong, ktlim\}@kaist.ac.kr} \hspace{2.5mm}
 \texttt{gy77@g.skku.edu} \hspace{2.5mm}
 \texttt{jinyounghan@skku.edu}
 }

\begin{document}

\maketitle

\input{Texts/0Abstract}

\input{Texts/1Introduction}

\input{Texts/2RelatedWork}

\input{Texts/3-1MentalKG}

\input{Texts/3-2MentalBench}

\input{Texts/4Experiments}

\input{Texts/5-1Disease_Analysis}

\input{Texts/5-2Error_Analysis}

\input{Texts/6Conclusion}

\section*{Acknowledgements}
We used AI assistants, including ChatGPT\footnote{\url{https://chatgpt.com/}}, Gemini\footnote{\url{https://gemini.google.com}} and Grammarly\footnote{\url{https://app.grammarly.com/}}, to support the writing and coding processes.

\bibliographystyle{plain}
\bibliography{references}
\clearpage

%%%%%%%%%%%%%%%%%%%%%%%%%%%%%%%%%%%%%%%%%%%%%%%%%%%%%%%%%%%%

\appendix

\startcontents[appendices]
\printcontents[appendices]{}{1}{\section*{Appendix}}
\clearpage

\input{Texts/7Limitations}

\input{Texts/Appendix_0scope_and_expert}

\input{Texts/Appendix_1KG}

\input{Texts/Appendix_2Benchmark}

\input{Texts/Appendix_4ResultandAnalysis}
\input{Texts/Appendix_3Experiments}
%%%%%%%%%%%%%%%%%%%%%%%%%%%%%%%%%%%%%%%%%%%%%%%%%%%%%%%%%%%%

\clearpage
\input{checklist.tex}

\end{document}

%% file: custom_package.tex
\usepackage{kotex}   % hangeulwriting
\usepackage{comment}

\newcommand{\kgeg}[3]{
    \small
    \item \textbf{#1}: #2
        \begin{itemize}[nosep]
            #3
        \end{itemize}
}
\newcommand{\sublist}[1]{
    \item \texttt{#1}
}

\usepackage{booktabs}
\usepackage{tabularx}

\usepackage{tcolorbox}
\tcbuselibrary{breakable}
\tcbset{
  mybox/.style={
    colback=gray!5,
    colframe=black!60,
    boxrule=0.5pt,
    arc=2mm,
    outer arc=2mm,
    left=6pt,
    right=6pt,
    top=6pt,
    bottom=6pt,
    breakable
  }
}

\newcommand{\ours}{\textsc{MentalBench}}
\newcommand{\kg}{\textsc{MentalKG}}

\usepackage{graphicx}
\usepackage{wrapfig}

\usepackage{enumitem}
\usepackage[utf8]{inputenc}

\usepackage{booktabs}
\usepackage{xcolor}

\usepackage{array}
\usepackage{makecell}
\usepackage{amsmath}
\usepackage{tabularx, makecell, threeparttable}
\usepackage{amsmath}

\usepackage[dvipsnames]{xcolor}  % ForestGreen 사용

\usepackage{multirow}

\usepackage{algorithm}
\usepackage{algpseudocode}

\usepackage[table]{xcolor}
\usepackage{adjustbox}

\usepackage{booktabs}
\usepackage[table]{xcolor}
\usepackage{tabularx}
\usepackage{ragged2e}
\usepackage[table]{xcolor} % 색상 및 테이블 색상 사용
\usepackage{colortbl}     % \rowcolor, \columncolor 등

\tcbuselibrary{listings} % 이 부분을 추가하세요

\usepackage{listings}

\usepackage{expl3}
\usepackage{xparse}
\usepackage{pifont}        
\newcommand{\Vmark}{\ding{51}}   % check sign

\ExplSyntaxOn

\clist_new:N \g_work_green_clist
\clist_new:N \g_work_black_clist

\NewDocumentCommand{\work}{mmm}
 {
   #1
   
   \clist_gset:Nn \g_work_green_clist { #2 }
   \clist_gset:Nn \g_work_black_clist { #3 }
   
   \int_step_inline:nn { 23 }
    {
      & % 다음 셀로 이동
      
      \clist_if_in:NnTF \g_work_black_clist { ##1 }
       {
         \textcolor{gray!60}{\Vmark}
       }
       {
         \clist_if_in:NnTF \g_work_green_clist { ##1 }
          {
            \textcolor{green!60!black}{\Vmark}
          }
          { }
       }
    }
   \\ 
 }

\ExplSyntaxOff

\usepackage{titletoc}

%% file: Texts/0Abstract.tex
\begin{abstract}
% We introduce \ours{}, a benchmark for evaluating whether large language models (LLMs) can make DSM-grounded psychiatric diagnostic decisions while calibrating their diagnostic commitment under varying levels of clinical ambiguity. Existing mental health benchmarks largely rely on social media data or supportive dialogue settings, limiting their ability to assess whether models can apply formal diagnostic criteria and differential diagnostic rules. At the core of \ours{} is \kg{}, a psychiatrist-built and validated knowledge graph encoding DSM-5 diagnostic criteria and differential diagnostic rules for 23 psychiatric disorders. Using \kg{} as an expert-curated logical backbone, we generate 24,750 synthetic clinical cases that systematically vary in information completeness and diagnostic complexity, enabling DSM-grounded evaluation. Our experiments show that although state-of-the-art LLMs perform well on noise-free queries that probe DSM-5 knowledge, they struggle to calibrate their confidence in diagnostic decision-making when distinguishing between disorders with overlapping symptoms. These findings reveal a diagnostic-commitment failure that is not captured by conventional mental health benchmarks focused on symptom classification, social-media signals, or supportive response generation. All resources are publicly released.\footnote{Code: \url{https://github.com/HoyunS/MentalBench}, Data: \url{https://hf.co/datasets/hysong/MentalBench}}

Large language models (LLMs) have attracted growing interest as supportive tools for psychiatric assessment and clinical decision support. However, existing mental health benchmarks largely rely on social media data or supportive dialogue settings, limiting their ability to assess whether models can apply formal diagnostic criteria and differential diagnostic rules. In this paper, we introduce \ours{}, a benchmark for evaluating whether LLMs can make DSM-grounded psychiatric diagnostic decisions under varying levels of clinical ambiguity. At the core of \ours{} is \kg{}, a psychiatrist-built and validated knowledge graph encoding DSM-5 diagnostic criteria and differential diagnostic rules for 23 psychiatric disorders. Using \kg{} as an expert-curated logical backbone, we generate 24,750 synthetic clinical cases that systematically vary in information completeness and diagnostic complexity, enabling DSM-grounded evaluation. Our experiments show that although state-of-the-art LLMs perform well on noise-free queries that probe DSM-5 knowledge, they struggle to calibrate their confidence when distinguishing between disorders with overlapping symptoms. These findings raise concerns about the reliability of LLMs as psychiatric decision-support tools and highlight the need for more evaluation that reflects the diverse challenges in real-world psychiatric diagnosis.  All resources are publicly released.\footnote{Code: \url{https://github.com/HoyunS/MentalBench}, Data: \url{https://hf.co/datasets/hysong/MentalBench}}

% All resources are publicly released.\footnote{Code: \url{https://github.com/HoyunS/MentalBench}, Data: \url{https://hf.co/datasets/hysong/MentalBench}}

\end{abstract}

%% file: Texts/1Introduction.tex
\section{Introduction\label{sec:intro}}
% Motivation: mental health에 LLM 많이 쓰인다. 근데 사실 이거 되게 복잡하다. / mental health diagnosis에는 두 가지 어려움 있다 1. 정보 불완전성 2. 심텀 오버랩 속에서의 주진단 감별. / 따라서 clinical knowledge in dsm-5를 외우는 것만으로는 충분하지 않고 잘 활용할 줄 알아야 한다.
% 하지만 기존의 mental health benchmark는 이러한 능력을 잘 측정하지 못하고 있다. 기존 연구 ~ from social media post ... 단순 cls task, no robust evidence, 그 외 지적한 문제점 / 정보 불완전성, 감별진단을 다루는 benchmark를 만들기 어려운 이유는 DSM-5가 매우 dense knowledge가 dependent하게 unstructred text 형태로 있기 때문에 전문가의 개입 없이는 이를 체계화(?) 하기 어렵기 때문이다.
% 이러한 문제를 해결하기 위해, 우리가 전문가와 함께 kg 만들어 체계화했다.

% The global burden of mental disorders continues to increase~\cite{world2022world}, motivating interest in whether large language models (LLMs) can support mental health–related analytical tasks~\cite{kim2025large, tong2025clinical, yang2023towards, yang2024mentallama}. Prior work suggests that LLMs can identify clinically relevant signals in text and generate plausible diagnostic hypotheses by leveraging mental health knowledge acquired during pretraining~\cite{li2024zero, omar2025exploring, song2025rationale, wang2024explainable}. However, psychiatric diagnosis is a high-stakes, criteria-driven process, and it remains unclear whether existing evaluations meaningfully assess the clinical soundness of these models in their rigorous application of diagnostic criteria.

The global burden of mental disorders continues to increase~\cite{world2022world}, motivating growing interest in the use of large language models (LLMs) as supportive tools for psychiatric diagnosis and clinical decision support~\cite{kim2025large, tong2025clinical, yang2023towards, yang2024mentallama}. Prior work suggests that LLMs can identify clinically relevant signals in text and generate plausible diagnostic hypotheses, demonstrating promising capabilities in mental health analysis~\cite{li2024zero, omar2025exploring, song2025rationale, wang2024explainable}. However, psychiatric diagnosis is a high-stakes, criteria-driven process requiring the rigorous application of standardized diagnostic criteria, and it remains unclear whether existing evaluations meaningfully assess the clinical soundness of LLMs in clinical contexts.  

% A central difficulty in applying diagnostic criteria arises from their inherently multi-criteria and boundary-sensitive decision structure. Diagnostic judgments are governed by standardized frameworks such as DSM-5~\cite{american2013diagnostic}, which require integrating symptom combinations, duration constraints, functional impairment, exclusion rules, and distinctions between clinically similar disorders~\cite{clark2017three, hyman2010diagnosis, kendell2003distinguishing}. In practice, these judgments are further complicated by incomplete or selectively reported patient information~\cite{feng2021comparison, mcbride2018occult}. As a result, evaluating diagnostic performance requires benchmarks that go beyond factual recall of diagnostic criteria and instead assess how models make clinically sound decisions when diagnostic boundaries are unclear.

A central difficulty in evaluating the psychiatric diagnostic capability of LLMs arises from the inherently multi-criteria and boundary-sensitive decision structure of psychiatric diagnosis. Psychiatric decision-making is governed by standardized frameworks such as DSM-5~\cite{american2013diagnostic}, which require integrating multi-faceted clinical factors, including symptom combinations, duration constraints, and functional impairments~\cite{clark2017three, hyman2010diagnosis, kendell2003distinguishing}. This challenge is particularly pronounced in differential diagnosis, where multiple psychiatric disorders with overlapping symptom profiles must be distinguished based on fine-grained criteria such as the intensity, duration, and temporal co-occurrence of individual symptoms~\cite{newson2021poor, kendell2003distinguishing}. Consequently, evaluating the psychiatric diagnostic capability of LLMs requires assessing not only whether models possess psychiatric knowledge, but also whether they can rigorously apply diagnostic criteria to resolve complex and ambiguous clinical scenarios.

Despite this need, existing mental health benchmarks remain limited. Importantly, this gap reflects not a lack of interest, but the inherent difficulty in constructing high-quality psychiatric diagnostic data~\cite{arnaout2026responsible, gongsurvey, ive2022leveraging}. Strict privacy regulations, the absence of objective biomarkers, and the complexity of formal diagnostic logic make large-scale curation of expert-validated datasets exceptionally challenging~\cite{ive2022leveraging, vedanta2024psychsynth}. Consequently, most benchmarks rely on social media data and self-reported labels~\cite{cohan2018smhd, macavaney2018rsdd, yates-etal-2017-depression}, which are ill-suited for evaluating DSM-grounded diagnostic judgments or for systematic differential diagnosis. As a result, they fail to provide the controlled scenarios necessary to isolate whether LLMs can accurately resolve clinical boundaries between competing diagnoses under inherent clinical ambiguity.

% To address this gap, we introduce \ours{}, a DSM-grounded benchmark designed to evaluate whether LLMs can rigorously apply psychiatric diagnostic criteria across diverse clinical scenarios. \ours{} consists of 24,750 question-answering pairs that evaluate whether LLMs can accurately identify psychiatric disorders under two core clinical challenges: information incompleteness and diagnostic complexity. The former reflects fragmented and selectively reported patient information in real-world clinical settings, whereas the latter captures differential diagnosis among disorders with overlapping symptoms and fine-grained diagnostic boundaries. To ensure reliable and clinically grounded evaluation, \ours{} is built upon \kg{}, a psychiatrist-built and validated knowledge graph that explicitly encodes DSM-5 diagnostic criteria and differential diagnostic rules for 23 psychiatric disorders. By structurally encoding complex diagnostic criteria and differential diagnostic logic, \kg{} enables \ours{} to systematically evaluate whether LLMs can rigorously apply psychiatric diagnostic reasoning under varying levels of clinical complexity.

To address this gap, we introduce \kg{} and \ours{}, a DSM-grounded logical backbone and its corresponding benchmark to assess whether LLMs can rigorously apply psychiatric diagnostic criteria across diverse clinical scenarios.  \kg{} is a psychiatrist-built and validated knowledge graph that explicitly encodes DSM-5 diagnostic criteria and differential diagnostic rules for 23 psychiatric disorders. By encoding these complex logics, \kg{} enables systematic control over DSM-5 knowledge and supports assessment across varying levels of clinical complexity. Built upon this foundation, \ours{} comprises 24,750 question-answer pairs that test model performance on two core clinical challenges: information incompleteness and diagnostic complexity. The former reflects fragmented and selectively reported patient information in real-world clinical settings, whereas the latter captures differential diagnosis among disorders with overlapping symptoms and fine-grained diagnostic boundaries. Through these challenges, \ours{} provides a systematic assessment of whether LLMs can rigorously apply psychiatric diagnostic reasoning across varying levels of clinical complexity.

Our findings reveal several critical limitations in the capabilities of current LLMs. First, while LLMs perform well on structured clinical descriptions with explicit diagnostic cues, their performance degrades substantially when diagnostic evidence is conveyed through fragmented, narrative-style patient language, suggesting difficulties in mapping real-world symptom expressions onto formal diagnostic criteria. Second, models exhibit diminished performance on underrepresented disorders in prior studies, indicating that current LLMs struggle to maintain reliable diagnostic performance across diverse psychiatric conditions. Finally, models struggle with differential diagnosis in clinically ambiguous settings, often failing to resolve diagnostic boundaries between multiple plausible disorders and a single definitive diagnosis, raising concerns about the reliability of LLMs in real-world settings where the number of plausible diagnoses is inherently uncertain.

\begin{itemize}[leftmargin=*,topsep=-2px,partopsep=0px]
    \item We construct \kg{}, an expert-built knowledge graph that transforms dense DSM-5 text into a structured representation encoding diagnostic criteria and differential diagnosis rules for 23 psychiatric disorders.
    \item We introduce \ours{}, a benchmark derived from expert-curated logic that varies information completeness and diagnostic complexity to assess the capability of LLMs to navigate diagnostic decisions under clinical ambiguity.
    \item Our experimental results demonstrate that current LLMs struggle to rigorously apply the clinical logic required for DSM-grounded differential diagnosis.
\end{itemize}

%% file: Texts/2RelatedWork.tex
\section{Related Work}
\paragraph{LLMs in Mental Health Diagnosis.}
Recent studies demonstrate that LLMs can identify clinically meaningful patterns in text and suggest potential diagnoses~\cite{ozgun2025trustworthy, song2025rationale, xiao2025mentrasuite, yang2023towards, yang2024mentallama}. Despite promising results, the studies demonstrate limitations in the systematic application of formal medical criteria to reach a diagnostic conclusion. For example, failing to identify key clinical signals~\cite{kim2025large, song2025rationale, xu2024mental, yang2023towards} or frequently confusing disorders with overlapping symptoms, such as depression and bipolar disorder~\cite{hengle2024still, lee-etal-2024-detecting-bipolar, ozgun2025trustworthy, srivastava2025towards}, indicates that models often lack the rigorous analytical reasoning needed to interpret psychiatric criteria. These findings underscore a significant deficiency in systematic reasoning rooted in structured medical knowledge, which is indispensable for maintaining clinical safety. Therefore, in this work, we present a benchmark to assess the clinical validity of model decisions, emphasizing rigorous execution of diagnostic logic rather than surface-level associations.

\paragraph{Benchmarks for Mental Health Diagnosis.}
Current datasets for assessing psychiatric diagnostic abilities primarily focus on classifying social media data~\cite{coppersmith2015clpsych, ji2022suicidal, kabir2023deptweet, li2024zero, pirina2018identifying, shing2018expert, song2023simple, wang2024explainable}. While these assessment methods address critical domains, they are not primarily designed to evaluate the rigorous application of formal psychiatric criteria essential for professional diagnosis~\cite{arnaout2026responsible, bucur-etal-2026-survey, shah2025advancing}. Consequently, the extent to which models can apply the diagnostic complexity and dense logical dependencies of the DSM-5 remains largely understudied.

%In general medicine, question-answering (QA) benchmarks are commonly employed to systematically assess domain knowledge~\cite{jin2021disease, jin2019pubmedqa, pal2022medmcqa}. However, mental health-related cases are only a small part of these resources and do not capture the unique challenges of psychiatric diagnosis, such as ambiguous symptom descriptions and overlapping conditions. While recent mental health benchmarks have adopted QA formats, they predominantly prioritize empathy, coping strategies, and broad mental health support~\cite{alhuzali2024mentalqa, fouda2026psychiatrybench, jin2023psyeval, liu2025psychbench, racha2025mhqa}.
Recent mental health benchmarks have adopted question-answering (QA) to assess LLMs' abilities in empathy, coping strategies, and broad mental health support~\cite{alhuzali2024mentalqa, fouda2026psychiatrybench, jin2023psyeval, liu2025psychbench, racha2025mhqa}. However, these frameworks often focus on small subsets of domain knowledge because they prioritize broader clinical objectives, resulting in less emphasis on the systematic application of rigorous diagnostic logic. To fill this gap, we introduce \ours{}, a benchmark designed to evaluate whether models can effectively navigate DSM-5 criteria under conditions of diagnostic ambiguity and complexity.

\paragraph{Challenges in Clinically Grounded Data Synthesis.}
Constructing clinically grounded psychiatric evaluation data is challenging because each existing source poses unique methodological challenges.
Most computational mental health studies rely on textbook cases~\cite{jin2021disease, li2021mlec, pal2022medmcqa, singhal2025toward} or social media data~\cite{coppersmith2015clpsych, ji2022suicidal, pirina2018identifying, shing2018expert}. However, textbook materials are highly likely to appear in training data, blurring the boundary between clinical logic and memorized patterns~\cite{fouda2026psychiatrybench, siam2025benchmarking}. Social media data often lacks expert-validated reliability and covers a limited range of disorders~\cite{ernala2019methodological, harrigian2020models}. Furthermore, acquiring real-world clinical records remains exceptionally challenging due to strict privacy regulations and the scarcity of expert labels~\cite{ive2022leveraging, vedanta2024psychsynth}. While some recent work has turned to synthetic data generation to address data scarcity, such approaches often struggle to maintain clinical grounding~\cite{kang2026synsym, murtaza2023synthetic}. We address these limitations by leveraging \kg{}, an expert-built knowledge graph, to establish a clinically grounded data synthesis pipeline.

%% file: Texts/3-1MentalKG.tex
\input{Figures/3Method/graph_construction}

\section{\label{sec:3_knowledge graph} \kg{}}
Psychiatric diagnosis is defined by pervasive symptom overlap and intricate exclusionary logic, creating entangled dependencies across disorders. Differential diagnosis is uniquely challenging in this environment, requiring the resolution of these dependencies through strict temporal and exclusionary rules. For instance, distinguishing schizoaffective disorder from major depressive disorder requires determining whether psychotic symptoms persist independently for at least two weeks. Since sound diagnosis depends on precise differentiation of clinical features, a rigorous evaluation framework is essential for determining whether LLMs demonstrate clinical rigor or merely rely on surface-level pattern recognition.
Constructing a benchmark with clinical rigor requires grounding in the DSM-5. However, its dense, unstructured text makes it difficult for non-experts to clearly extract critical elements such as symptom combinations, duration constraints, and exclusion rules. To address this, we introduce \kg{} to formalize these dependencies. It transforms unstructured text into a knowledge graph, providing a logical backbone for DSM-grounded clinical assessment.

\subsection{Structural Formalization of Diagnostic Criteria}

\paragraph{Graph Topology and Composition.} To transform the dense DSM-5 text into a structured format, we conduct expert-driven manual extraction and formalization of clinical entities. As illustrated in \autoref{fig:graph_schema}, the \kg{} schema models the hierarchical and relational dependencies among clinical entities. Quantitatively, the constructed graph comprises 23 Disorder nodes, 23 Symptom Group nodes, 84 Symptom nodes, and 87 Differential Diagnosis nodes. These entities are interconnected via specific relations, including 67 mandatory edges, 91 included-in (hierarchical) edges, and 174 differential diagnosis edges, forming a dense network of clinical dependencies.

\paragraph{Diagnostic Rules and Constraints.}
A core feature of \kg{} is the integration of complex diagnostic conditions directly into the graph attributes. As shown in \autoref{fig:graph_schema}, disorder nodes are enriched with essential constraints, including mandatory symptoms, duration constraints, and minimum symptom thresholds. Furthermore, differential diagnosis nodes encode the exclusionary logic necessary for resolving symptom overlap. Each of these nodes defines a triggering condition that prompts the differential logic, a key difference separating the disorders, and the specific discriminating rules required to reach a definitive conclusion.

\paragraph{Symptom Diversification via Subtypes.}
A common limitation of graph-based generation is the tendency to produce rigid, textbook-style phrasing. To mitigate this, we integrate symptom subtypes into the schema. By defining varied manifestations for each core symptom, \kg{} ensures that the generated cases exhibit high linguistic and symptomatic diversity. This diversification provides a robust foundation for evaluating LLMs across a wide spectrum of clinical presentations, moving beyond repetitive patterns and enabling a more rigorous assessment. Further specifications and representative examples for each node type are provided in Appendix~\ref{appen:kg}.

\subsection{Expert-Driven Construction}
The entire construction process for \kg{} is carried out in collaboration with two experts.\footnote{Detailed professional qualifications for each expert involved are provided in Appendix~\ref{appen:expert}.} Specifically, the psychiatrist validated the following aspects of the graph: (1) the clinical appropriateness of symptom subtypes within symptom nodes, (2) the alignment of the formalized diagnostic criteria with DSM-5 standards, and (3) the clinical validity of the differential diagnosis logic. Through iterative cycles of expert review and refinement, \kg{} can structurally encode the complex diagnostic criteria and differential-diagnostic logic of the DSM-5 rather than simply serving as a collection of disorder--symptom connections. A detailed validation process is provided in Appendix~\ref{appen:kg_ev}.

%% file: Figures/3Method/graph_construction.tex
\begin{figure}[t]
    \centering
    \includegraphics[width=0.95\linewidth]{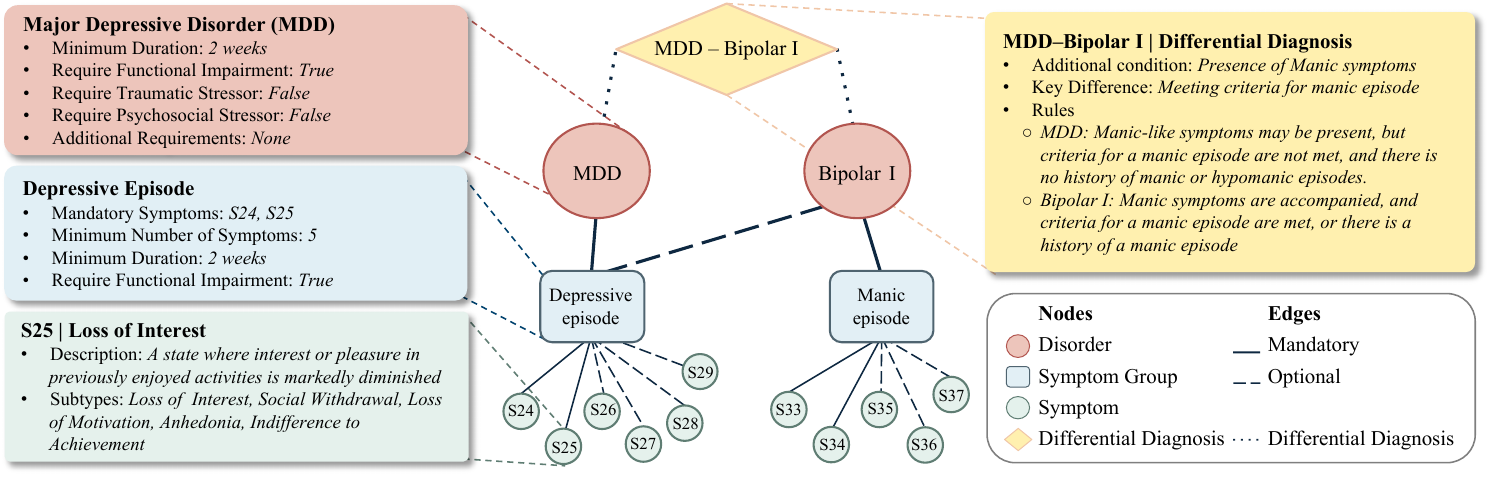}
    %\vspace{-0.1in}
    \caption{\label{fig:graph_schema} Example of the \kg{} schema. It demonstrates the relational dependencies among disorders, symptom groups, and symptoms, as well as directional differential diagnoses. It also details the granular clinical attributes embedded in nodes and edges, including discriminating rules, symptom subtypes, and diagnostic constraints (e.g., duration and thresholds).}
\end{figure}

%% file: Texts/3-2MentalBench.tex
\section{\label{sec:4_mentalbench}\ours{}}
% Leveraging \kg{} as a structured foundation, we generate \ours{}, a large-scale question-answering benchmark comprising 24,750 synthetic clinical cases that present specific patients’ psychiatric symptom profiles and their corresponding diagnostic options. \ours{} is defined by three core attributes: (1) \textbf{Scenario Diversity}: By systematically varying information completeness and diagnostic complexity, the benchmark enables comprehensive evaluation across a broad spectrum of clinically critical scenarios. (2) \textbf{Scalable Clinical Composition}: The framework integrates diverse symptom patterns across 23 psychiatric disorders and 87 differential diagnoses, providing a large-scale evaluation of the clinical landscape. (3) \textbf{Clinical Soundness}: Grounded in the psychiatrist-built \kg{}, the dataset undergoes expert-in-the-loop validation to ensure the highest level of clinical rigor.
Leveraging \kg{} as a structured foundation, we generate \ours{}, a large-scale question-answering benchmark comprising 24,750 synthetic clinical cases. We designed this benchmark to evaluate models on two primary challenges: information incompleteness and diagnostic complexity. To ensure a comprehensive evaluation of each disease's diverse manifestations and their differential diagnoses, we generate these cases using a multi-stage data-synthesis pipeline. Furthermore, to guarantee clinical soundness, domain experts explicitly designed the seed generation, option design, and ground-truth assignment processes, followed by rigorous validation of the final outputs.

\input{Figures/3Method/Type_example}
\subsection{Categorization of Clinical Evaluation Scenarios}
The design of \ours{} focuses on two primary challenges: \textbf{information incompleteness} and \textbf{diagnostic complexity}. The former stems from expression diversity and selective disclosure by patients~\cite{feng2021comparison, mcbride2018occult}, whereas the latter arises from clinical ambiguity in differential diagnosis~\cite{mcallister2016clinical}. \autoref{fig:Type_comparison} illustrates the distinct features of four case types that differ along these two challenges.

\paragraph{Information Incompleteness.}
\textit{Types 1} and \textit{2} are created using single-disease nodes to assess whether the models understand the full set of features for each disease. The primary challenge lies in varying information density, which tests the model's ability to identify missing evidence and maintain diagnostic accuracy despite fragmented and unclear narratives. Accordingly, as illustrated in \autoref{fig:Type_comparison}, \textit{Type 1} presents professional summaries in which diagnostically relevant information is explicitly organized and complete. In contrast, \textit{Type 2} preserves the minimum evidence required for the target diagnosis but presents it through fragmented, colloquial, and selectively emphasized self-reports, requiring models to recover diagnostic cues from less structured patient narratives. An analysis of the linguistic divergence resulting from these presentation styles is provided in Appendix~\ref{appen:linguistic}.

\paragraph{Diagnostic Complexity.}
\textit{Types 3} and \textit{4} are derived from differential diagnosis nodes to evaluate whether the models can accurately apply diagnostic criteria when diseases share symptoms. The core challenge lies in the differential diagnosis itself. To determine the differential diagnosis, it is essential to identify all candidate disorders based on the observed symptoms. If a discriminatory rule exists, it should be used to narrow the list to a single diagnosis. To this end, \textit{Type 3} constructs controlled ambiguity cases in which the shared symptom profile satisfies the common diagnostic evidence for competing disorders, but the discriminating evidence required to rule out one candidate is absent. As a result, multiple diagnoses remain consistent with the presented evidence. In contrast, \textit{Type 4} includes the relevant discriminating evidence, requiring models to narrow the candidate set to a single definitive diagnosis. This challenge requires models to adjust diagnoses when differentiating disorders with overlapping symptoms, using precise criteria. An analysis of the diagnostic complexity arising from semantic overlap is provided in Appendix~\ref{appen:complexity}.

\subsection{Benchmark Construction Pipeline}
To ensure both clinical validity and structural diversity, we develop a pipeline that transforms expert knowledge from \kg{} into a diverse set of evaluation cases. The pipeline consists of four sequential stages as illustrated in \autoref{fig:qa_framework}.

% 증상 조합을 전문가 검증 하에 엄밀하게 추출했다
\paragraph{Seed Generation.} 
This stage focuses on sampling the clinical profile from \kg{} to establish the diagnostic foundation of each case. \textit{Type 1} seeds are constructed to strictly satisfy all DSM-5 requirements, while \textit{Type 2} seeds are intentionally designed to be incomplete, falling below diagnostic thresholds. To maintain clinical rigor, two experts validate these seeds to ensure that the minimum requirements for a correct diagnostic determination are present, even when specific information is omitted. For differential diagnosis cases, we first instantiate profiles that satisfy the shared triggering conditions of a differential-diagnosis pair. For \textit{Type 3} seeds, we omit the discriminating evidence needed to resolve the pair, yielding multiple diagnoses that remain consistent with the case. For \textit{Type 4} seeds, we include the relevant discriminating evidence, yielding a single diagnosis under the encoded differential rule. Experts reviewed both settings to ensure that the resulting labels align with the intended diagnostic logic. In total, we generated 5 unique seeds per disease for \textit{Types 1, 3} and \textit{4}, and 10 seeds per disease for \textit{Type 2}. Further details on the seed generation and expert validation process are provided in Appendix~\ref{appen:seed_generation}.

% 표현을 다양하게 하기 위해, 각 증상에 대한 subtype을 다양하게 뽑고, patient profile도 다양하게 했다.
\paragraph{Profile Augmentation.} 
To enhance expressive diversity and avoid repetitive patterns, we enrich the clinical seeds by incorporating disease subtypes from \kg{} and synthesizing detailed patient profiles. Specifically, we generate five distinct cases per seed by sampling varied symptom manifestations defined by the subtypes in \kg{}. Each case is assigned a patient demographic identity, including age, gender, occupation, and marital status, drawn from 300 unique combinations. This process yields 5 distinct patient profiles per seed, ensuring that \ours{} captures a broad spectrum of clinical presentations beyond rigid textbook descriptions. Details on profile augmentation are provided in Appendix~\ref{appen:demographic}.

% generator bias가 없도록 3개의 생성 모델을 사용해서 생성했다.
\paragraph{Data Synthesis.} 
To reduce generator bias, we employ a multi-model synthesis framework that uses three distinct LLMs to generate the clinical cases. This ensemble approach ensures that the evaluation data does not overfit to the linguistic style or structural tendencies of any single model, as discussed in Appendix~\ref{appendix:bias}. Each patient profile is converted into a natural-language narrative tailored to four task types: formal third-person descriptions and medical-chart-style summaries for \textit{Type 1}, and colloquial first-person narratives for \textit{Types 2, 3} and \textit{4}. During generation, we prompted models to ensure that clinical cases accurately reflect the clinical information in the seeds, including symptom presentation, duration, and constraints. Detailed prompts for case generation are provided in Appendix~\ref{appen:prompts}. As a result, we generate 24,750 clinical cases in total, comprising 1,725 for \textit{Type 1}, 3,450 for \textit{Type 2}, 6,525 for \textit{Type 3}, and 13,050 for \textit{Type 4} cases. Detailed statistics on the distribution of the generated samples are provided in Appendix~\ref{appen:statistics}.

% This synthesis is followed by an expert-in-the-loop refinement process to ensure both clinical accuracy and stylistic authenticity.

\input{Figures/3Method/qa_gen_framework}

\paragraph{Option and Ground Truth Generation.}
The final stage defines the options and gold labels. Each case includes 4 diagnostic options. \textit{Types 1, 2} and \textit{4} are single-answer tasks. \textit{Type 3} is a controlled-ambiguity task in which the correct label is a set of diagnoses consistent with the evidence. In this benchmark, the set contains two diagnoses that match the paired differential-diagnosis structure encoded in \kg{}. Distractors are selected from \kg{} based on symptomatic overlap to rigorously assess discriminative capability, and experts review each problem to ensure clinical validity. Further details on option design and distractors are provided in Appendix~\ref{appendix:option}.

\input{Tables/Dataset/expert_validation}

\subsection{Expert Validation}
To rigorously assess the quality of the synthesized narratives, two experts evaluated 220 scenarios, randomly and evenly sampled across the four data types (\textit{Type 1--4}), using a 5-point Likert scale. This assessment focuses on three critical dimensions: \textit{Linguistic Naturalness (LN)}, \textit{Diagnostic Validity (DV)}, and \textit{Clinical Realism (CR)}. As shown in \autoref{tab:expert_valid}, \ours{} achieved consistently high ratings across all metrics, confirming its quality for evaluating LLMs in complex psychiatric contexts. Detailed descriptions of the evaluation criteria and comprehensive validation results are provided in Appendix~\ref{appendix: expert validation}.

%% file: Figures/3Method/Type_example.tex
\begin{figure}
    \centering
    \includegraphics[width=\linewidth]{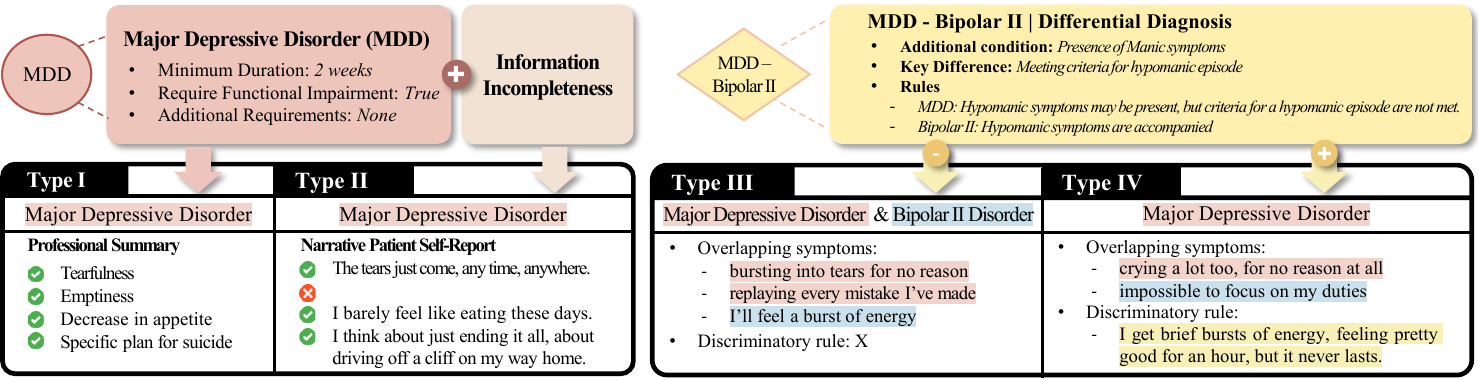}
    \vspace{-0.1in}
    \caption{\label{fig:Type_comparison} Examples of generated clinical cases illustrating four evaluation scenarios. \textit{Types 1–2} assess information incompleteness by contrasting medical charts with colloquial narratives, reflecting expression diversity and selective disclosure. \textit{Types 3–4} evaluate diagnostic complexity through differential diagnosis. Detailed examples of each type are provided in Appendix~\ref{appen:example}.}
\end{figure}

%% file: Figures/3Method/qa_gen_framework.tex
\begin{figure}[t]
    \centering
    \includegraphics[width=1\linewidth]{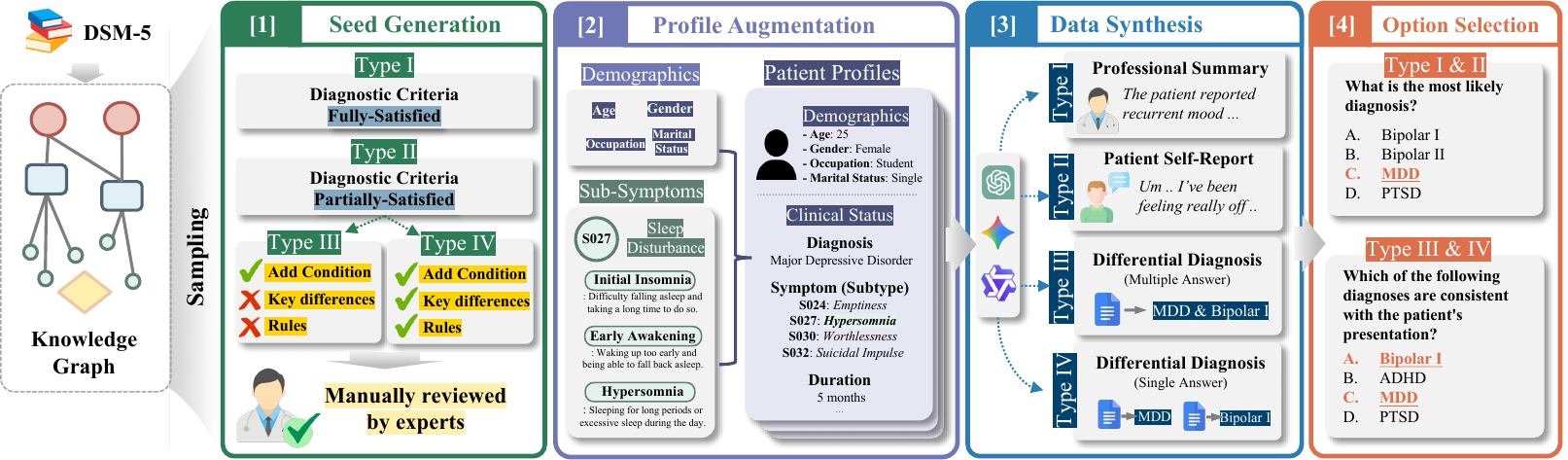}
    \vspace{-0.1in}
    \caption{\label{fig:qa_framework}Overview of the clinical case generation pipeline for constructing \ours{}.}
\end{figure}

% Our pipeline consists of three stages: (1) Symptom Profile Construction, where concrete symptom configurations are sampled from \kg{}; (2) Patient Profile Instantiation, which assigns demographic and specific symptom manifestations; and (3) Clinical Case Generation, which synthesizes clinical cases based on patient profiles. These processes are applied under two clinical scenarios: Single-Disease Identification and Differential Diagnosis.

%% file: Tables/Dataset/expert_validation.tex
\begin{wraptable}{r}{0.48\textwidth}
\centering
\scriptsize
\vspace{-2em}
\caption{\label{tab:expert_valid}Expert validation results by two experts. }
\begin{tabular}{lrrrrrrr}
\hline
\multirow{2}{*}{\textbf{Type}} & \multicolumn{2}{c}{\textbf{LN}} & \multicolumn{2}{c}{\textbf{DV}} & \multicolumn{2}{c}{\textbf{CR}} \\ 
\cline{2-3} \cline{4-5} \cline{6-7}
 & \textbf{E1} & \textbf{E2} & \textbf{E1} & \textbf{E2} & \textbf{E1} & \textbf{E2} \\ \hline
Type 1 & 4.87 & 4.65 & 5.00 & 4.52 & 4.83 & 4.57 \\
Type 2 & 4.96 & 5.00 & 4.91 & 4.30 & 4.78 & 4.91 \\
Type 3 & 4.98 & 5.00 & 4.20 & 4.02 & 4.91 & 4.99 \\
Type 4 & 4.93 & 5.00 & 4.70 & 4.61 & 4.79 & 5.00 \\ \hline
\textbf{Overall} & 4.94 & 4.91 & 4.70 & 4.36 & 4.83 & 4.87 \\ \hline
\end{tabular}
\end{wraptable}

%% file: Texts/4Experiments.tex
\section{Experiments and Analysis}
% \subsection{Experimental Setup}
\paragraph{Experimental Setup.}
We conduct a comprehensive evaluation of diverse LLMs on \ours{} to assess their psychiatric diagnostic capabilities in clinical settings. 
We employ closed-source models from the GPT, Gemini, and Claude families, as well as open-source models including Qwen 2.5, Qwen 3, Gemma 3, Llama-3.1, and MentaLLaMA. Notably, MentaLLaMA~\cite{yang2024mentallama} is a specialized model fine-tuned from Llama-2 for the mental health domain. For \textit{Types 1} and \textit{2}, models are instructed to select a single answer, whereas for \textit{Types 3} and \textit{4}, they are directed to choose one or more answers. Details of the model versions, prompts, and settings are in Appendix~\ref{appen:setup}. We measure accuracy using an exact match metric, recognizing a prediction as correct only if it identifies the complete set of ground-truth answers.

\input{Tables/Experiments/main}

\subsection{Diagnostic Capability Assessment}
\paragraph{Main Results.} \autoref{tab:main_results} presents results across diverse model families on \ours{}. It reveals a clear hierarchy: proprietary models generally outperform open-source models, and accuracy tends to increase with model scale. This suggests that model scale and model family are associated with better performance, but the gains are uneven across scenario types.

The results also highlight the benchmark's exceptional difficulty, as even top frontier models achieve a maximum score of only 62.69. The challenge is especially clear in \textit{Type 3} and \textit{Type 4} scenarios, where performance often drops below 20. These steep declines show that \ours{} exposes a diagnostic ambiguity that is not directly targeted by many existing mental-health benchmarks. Notably, MentaLLaMA underperforms despite its domain specialization, highlighting that relying on the older Llama 2 architecture is insufficient and underscoring the need for new domain-specific models built on advanced architectures.

% average 폭 type 1이랑 2 -> 어떤 모델이 incomplete 정보에 더 약한가

\paragraph{Impact of Information Incompleteness.} 
Our results show that model scale significantly affects robustness to incomplete information. Shifting from structured charts (\textit{Type 1}) to subjective narratives (\textit{Type 2}) causes a steep $10.58$-point average performance drop for models under 30B, whereas larger models (over 30B) are resilient, with only a $3.74$-point average gap. 
%This contrast suggests that increased model scale enables robust semantic inference from partial clinical data, whereas smaller ones remain constrained to rigid pattern matching that fails to generalize amid narrative noise.
This contrast suggests that increased scale enables robust semantic inference, whereas smaller models rely on surface-level associations that fail to resolve the semantic overlap analyzed in Appendix~\ref{appen:linguistic}.

\paragraph{Diagnostic Complexity and Divergent Failure Modes.}
Different model types exhibit distinct error patterns in differential diagnosis. Open-source models have an average positive gap of $22.74$ (\textit{Type 3} $-$ \textit{Type 4}), primarily because they do not effectively enforce exclusion rules in clear cases (\textit{Type 4}). Conversely, closed-source models perform worse in ambiguous situations (\textit{Type 3}), with an average gap of $-39.60$. This reveals a diagnostic-commitment trade-off: open-weight models often fail to exclude plausible distractors in unambiguous cases, whereas proprietary models often under-retain plausible alternatives in ambiguous cases.

%% file: Tables/Experiments/main.tex
\begin{table}[tb!]
\centering
\caption{\label{tab:main_results}Main evaluation results (accuracy) across four distinct types on \ours{}. Overall scores represent the weighted average proportional to the sample size of each type.}
\tiny
\begin{minipage}[t]{0.48\textwidth}
\centering
\setlength{\tabcolsep}{5pt}
\begin{tabular}{lrrrrr}
\toprule
\textbf{\textsc{Model}} & \textbf{Type 1} & \textbf{Type 2} & \textbf{Type 3} & \textbf{Type 4} & \textbf{Overall} \\
\midrule

\textbf{Qwen 2.5 7B}  & 85.85 & 76.87 & 50.70 & 11.59 & 36.12 \\
\textbf{Qwen 2.5 14B} & 94.15 & 82.26 & 44.78 & 31.83 & 46.55 \\
\textbf{Qwen 2.5 32B} & 92.87 & 83.04 & 49.20 & 30.89 & 47.08 \\
\textbf{Qwen 2.5 72B} & 93.85 & 83.10 & 52.75 & 18.25 & 41.66 \\
\midrule

\textbf{Gemma 3 4B}  & 79.65 & 64.90 & 34.04 & 18.41 & 33.34 \\
\textbf{Gemma 3 12B} & 92.81 & 78.32 & 48.28 & 15.79 & 38.50 \\
\textbf{Gemma 3 27B} & 92.87 & 84.09 & 52.51 & 9.20 & 36.85 \\
\midrule
\textbf{Llama-3.1 8B}  & 84.99 & 71.01 & 40.12 & 14.18 & 33.60 \\
\textbf{Llama-3.1 70B} & 91.94 & 80.81 & 39.75 & 34.97 & 47.09 \\
\midrule

\textbf{MentaLLaMA 7B}  & 30.90 & 30.67 & 2.19 & 25.98 & 19.65 \\
\textbf{MentaLLaMA 13B} & 52.17 & 46.75 & 18.10 & 16.17 & 23.45 \\

\bottomrule
\end{tabular}
\end{minipage}
\hspace{4mm}
\begin{minipage}[t]{0.48\textwidth}
\centering
\setlength{\tabcolsep}{5pt}
\begin{tabular}{lrrrrr}
\toprule
\textbf{\textsc{Model}} & \textbf{Type 1} & \textbf{Type 2} & \textbf{Type 3} & \textbf{Type 4} & \textbf{Overall} \\
\midrule
\textbf{Qwen 3 8B}  & 89.74 & 77.77 & 55.40 & 2.53 & 33.09 \\
\textbf{Qwen 3 14B} & 90.78 & 75.48 & 52.26 & 14.09 & 38.14 \\
\textbf{Qwen 3 32B} & 90.20 & 83.13 & 48.03 & 27.37 & 44.86 \\
\textbf{Qwen 3 235B} & 85.33 & 84.17 & 54.19 & 28.63 & 47.06 \\
\midrule
\textbf{GPT 4o}  & 74.61 & 79.97 & 40.25 & 48.30 & 52.43 \\
\textbf{GPT 5-mini} & 71.01 & 88.46 & 9.01 & 77.72 & 60.63 \\
\textbf{GPT 5.1} & 94.38 & 90.43 & 22.48 & 70.30 & 62.17 \\
\midrule
\textbf{Claude Haiku-4.5}  & 94.20 & 87.39 & 29.87 & 65.00 & 60.89 \\
\textbf{Claude Sonnet-4.5} & 94.32 & 91.16 & 14.99 & 74.84 & \textbf{62.69} \\
\midrule
\textbf{Gemini 2.5-flash}  & 93.39 & 85.27 & 27.94 & 60.96 & 57.90 \\
\textbf{Gemini 2.5-pro} & 93.91 & 88.20 & 35.16 & 59.82 & 59.65 \\
\bottomrule
\end{tabular}
\end{minipage}
\end{table}

%% file: Texts/5-1Disease_Analysis.tex
\subsection{Disease-level Analysis}
\input{Figures/5Experiments/Type2_heatmap}

% underrepresented disorder일수록 information incompleteness에 취약하다

\paragraph{Information Completeness in Underrepresented Disorders.}
We examine disease-specific accuracy of LLMs, focusing on individual disorders. \autoref{fig:heatmap} compares accuracy across 23 disorders between \textit{Type 1} and \textit{Type 2}. The result indicates that the \textit{D003}–\textit{D007} cluster (ADHD and Schizophrenia spectrum) collapses during the transition to \textit{Type 2}. This drop aligns with the coverage analysis in Appendix~\ref{appen:scope}, which identifies that disorders in this cluster--particularly Schizoaffective Disorder (\textit{D006}, \textit{D007})--have been largely excluded from prior research. This correlation suggests that disorders that are underrepresented are more likely to be affected by incomplete information.

\paragraph{Symptomatic Overlap Drives Diagnostic Complexity.}
\input{Figures/5Experiments/sankey_diagram_diff_diag}

To investigate the impact of symptomatic overlap on model decision-making, we analyze results in the unambiguous scenarios (\textit{Type 4}) by disease category. We focus on disorder groups that share overlapping dominant symptoms: \textit{depressive episodes} and \textit{psychotic features}. \autoref{fig:error_diff_diag} visualizes the mapping between ground-truth diagnoses and model predictions, revealing significant entanglement within symptomatically similar clusters.
% specifically, those characterized by depressive episodes and those defined by psychotic features. 

In the depressive cluster (bottom), we observe frequent misclassification between Major Depressive Disorder (MDD) and Adjustment Disorder, as well as between MDD and Bipolar Disorders. These error patterns align with prior research highlighting the challenge of distinguishing between disorders that share depressive episodes~\cite{ozgun2025trustworthy}. Notably, models struggle significantly to correctly identify Persistent Depressive Disorder (PDD), a failure likely driven by their inability to resolve fine-grained temporal criteria, particularly the specific duration required to distinguish it from the more dominant MDD class.
% These error patterns align with prior research highlighting the challenge of distinguishing unipolar from bipolar depression~\cite{ozgun2025trustworthy}.

% \input{Figures/5Experiments/sankey_diagram_diff_diag}

Even more critical confusion emerges in the psychotic cluster (top), where boundaries between Schizophrenia, Schizoaffective Disorder, and Bipolar I with Psychotic Features are blurred. Unlike depressive disorders, this group has been underrepresented in prior computational studies---specifically, Schizoaffective Disorder has rarely been evaluated in existing benchmarks. The high level of mutual confusion observed here suggests that current models lack the ability to differentiate these conditions based on fine-grained differential criteria (e.g., the concurrence of mood episodes with psychosis), underscoring an urgent need for future work to address these complex diagnoses.
% 증상이 공유되는 애들끼리는 다른 진단기준들 (기간, ..) 들의 검토가 필요한데 모델들이 그걸 잘 못한다 

%% file: Figures/5Experiments/Type2_heatmap.tex
% \begin{wrapfigure}{R}{0.42\linewidth}
\begin{comment}
\begin{wrapfigure}{R}{0.48\linewidth}
    \vspace{-1.5em}
    \centering
    \includegraphics[width=0.95\linewidth]{Figures/5Experiments/Accuracy_Heatmap_TypeII_cropped.pdf}
    \vspace{-0.9em}
   
    \caption{\label{fig:heatmap} \small Heatmap of diagnostic accuracy across 23 mental disorders for each model, for \textit{Type 2}}
    \vspace{-0.5em}
\end{wrapfigure}
\end{comment}

\begin{figure}[h]
    \centering
    \includegraphics[width=0.95\linewidth]{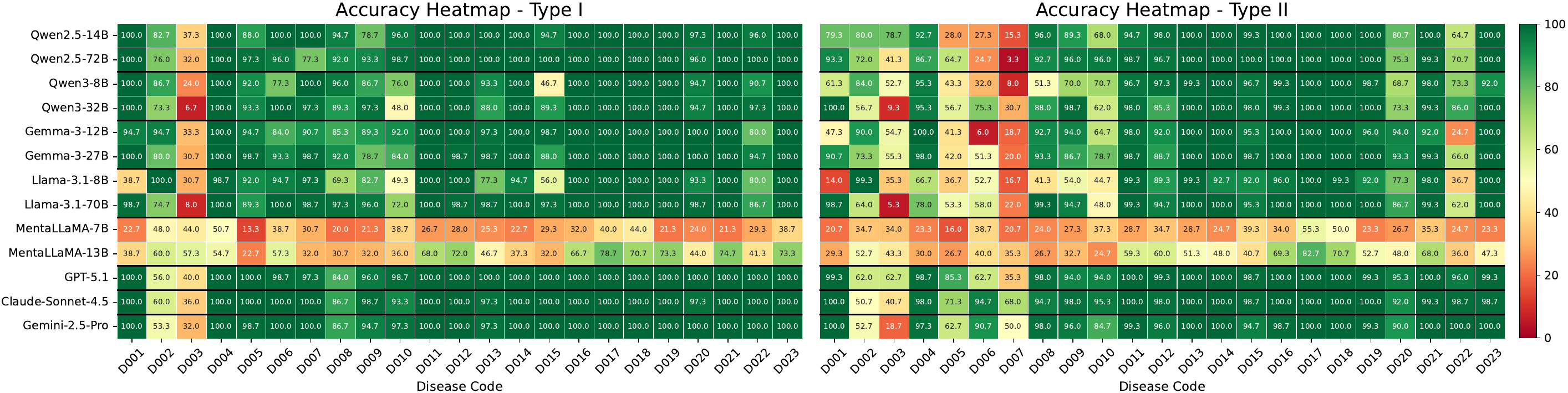}
    \caption{\label{fig:heatmap}Heatmap of diagnostic accuracy across 23 mental disorders for each model, for \textit{Types 1} and \textit{2}. Greener cells indicate higher performance, whereas red indicates lower accuracy.}
\end{figure}

%% file: Figures/5Experiments/sankey_diagram_diff_diag.tex
% \begin{wrapfigure}{R}{0.42\linewidth}
\begin{wrapfigure}{R}{0.5\linewidth}
    \vspace{-1em}
    \centering
    \includegraphics[width=0.95\linewidth]{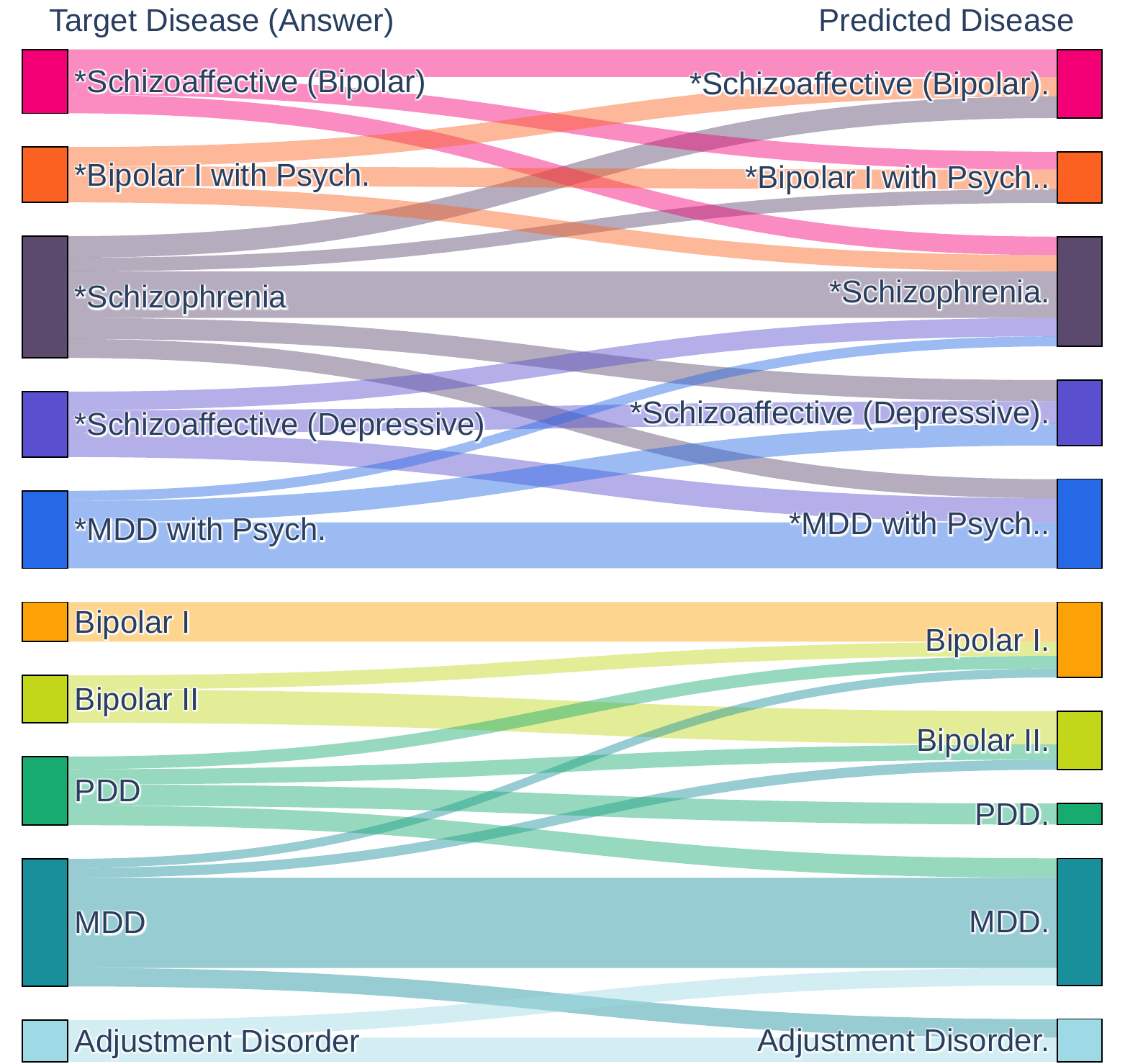}
    \caption{\label{fig:error_diff_diag}Mapping of ground truth (left) to predicted diagnoses (right) for \textit{Type 4}.}
\end{wrapfigure}

%% file: Texts/5-2Error_Analysis.tex
\subsection{Error Pattern Analysis\label{sec:error_analysis}}

\paragraph{Models Struggle to Calibrate Their Diagnostic Decisions.} 
%\paragraph{Balancing Strictness with Inclusivity is Challenging.} 
To investigate the reasoning failures driving the performance divergence observed in \autoref{tab:main_results}, we conduct a fine-grained error analysis on \textit{Type 3 (Ambiguous)} and \textit{Type 4 (Unambiguous)} scenarios. We categorize errors into three distinct types: (1) Over-diagnosis, where the model predicts supersets of the ground truth (i.e., failing to rule out distractors); (2) Under-diagnosis, where the model predicts subsets (i.e., missing valid comorbidities); and (3) Incorrect, where the model predicts unrelated diagnoses.

\autoref{fig:error_analysis} illustrates the distribution of these error types, showing a fundamental failure to calibrate diagnostic commitment. Open-source models exhibit excessive commitment in \textit{Type 4}, leading to widespread Over-diagnosis because they fail to apply exclusionary rules. Conversely, proprietary models appear rigidly calibrated toward decisiveness in \textit{Type 3}, resulting in frequent Under-diagnosis (insufficient commitment) by forcing single-label predictions even when clinical evidence supports multiple conditions. This divergence highlights that current models lack the adaptive reasoning required to achieve this calibration, instead exhibiting biases toward either overly expansive or excessively restrictive diagnostic tendencies, depending on the model architecture. Analysis using quantitative metrics is provided in Appendix~\ref{appen:metric_analysis}.

% \autoref{fig:error_analysis} illustrates the distribution of these error types across representative models from each family. The results show a fundamental difference in constraint handling. Open-source models struggle to adhere to exclusionary rules in Type 4, leading to widespread Over-diagnosis as they fail to filter out distractors. Conversely, proprietary models appear over-calibrated toward decisiveness in Type 3, resulting in frequent Under-diagnosis by forcing single-label predictions even when clinical evidence supports multiple conditions. This highlights that current models struggle to simultaneously optimize for exclusionary strictness and diagnostic inclusivity.

\input{Figures/5Experiments/Error_analysis}

%\paragraph{LLMs Lack Stable Decision Boundaries.}
\paragraph{Models Lack Intrinsic Constraint Awareness.}
% 제약조건에 대한 인지가 사람처럼 잘 되지 않음
Motivated by these calibration failures, we investigate the impact of prompt constraints to determine whether the performance gap stems from a lack of clinical knowledge or an inability to infer decision boundaries. Specifically, we compare performance in the \textit{Type 4} setting across three instruction levels, where the prompt explicitly directs the model to `\textit{choose all applicable diagnoses}' (\textit{Multiple}), `\textit{choose one or more answers}' (\textit{One-or-more}), or `\textit{choose only one answer}' (\textit{Single}). The prompts are listed in Appendix~\ref{appen:prompts}.

\autoref{fig:prompt_analysis} presents the performance results for each prompting strategy. While all models perform robustly under the \textit{Single}-answer constraint, they struggle significantly in the \textit{Multiple}- and \textit{One-or-more} settings. 
These results suggest that model failures are not solely driven by a lack of clinical knowledge. Instead, the significant improvement under the explicit constraints (\textit{Single}-answer) suggests that models struggle to determine the appropriate diagnostic boundaries from clinical evidence without external guidance. Ultimately, this dependency reveals a critical limitation: current models cannot autonomously calibrate their diagnostic scope, a fundamental requirement in real-world scenarios where the number of conditions is never predefined.

%% file: Figures/5Experiments/Error_analysis.tex
\begin{figure*}[t]
    \centering
    \begin{minipage}{0.49\linewidth}
        \centering
        \includegraphics[width=\linewidth]{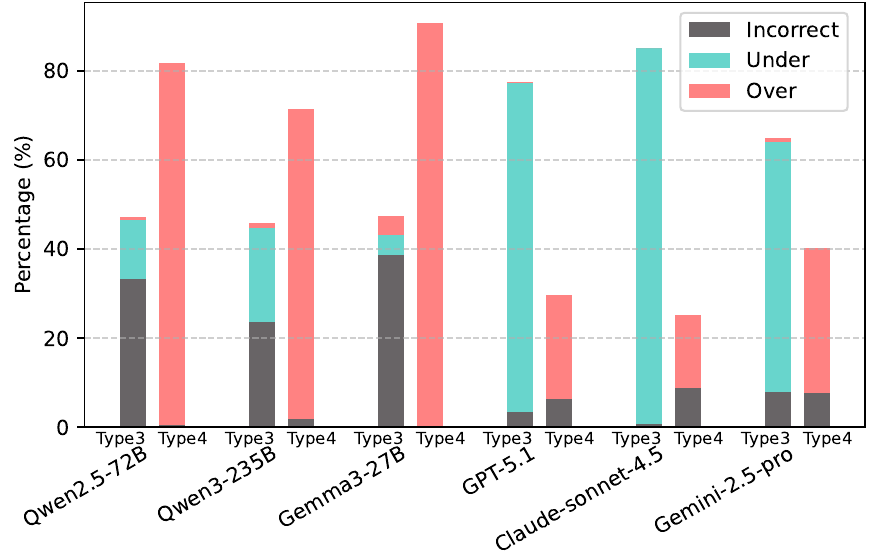}
        \caption{\label{fig:error_analysis} Distribution of error types across models for \textit{Type 3} and \textit{Type 4} tasks.}
    \end{minipage}\hfill
    \begin{minipage}{0.49\linewidth}
        \centering
        \includegraphics[width=\linewidth]{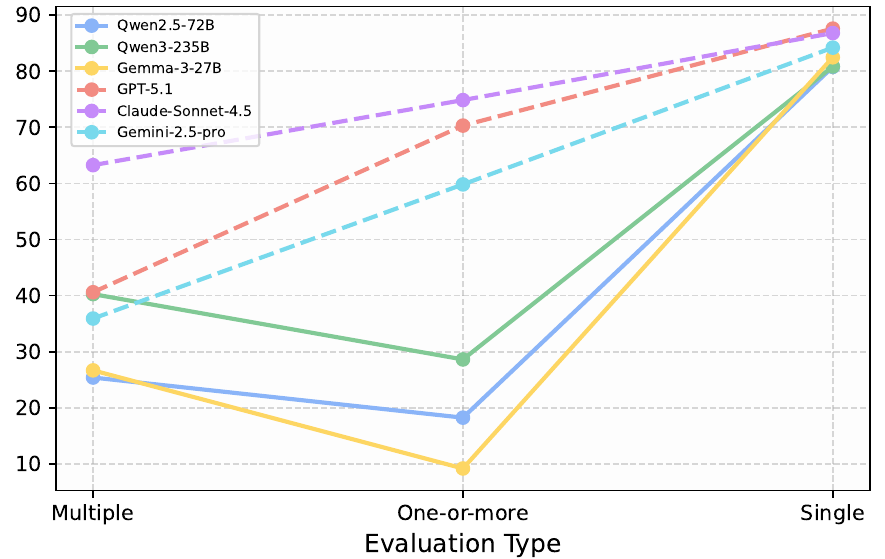}
        \caption{\label{fig:prompt_analysis} Impact of explicit prompt instructions for output boundaries on model performance.}
    \end{minipage}
    
\end{figure*}

%% file: Texts/6Conclusion.tex
\section{Conclusion}
We introduce \ours{}, a benchmark grounded in \kg{} for rigorously evaluating psychiatric diagnostic reasoning. While our experiments confirm that LLMs possess robust latent knowledge of DSM-5 criteria, they exhibit critical diagnostic failures when applied to ambiguous clinical scenarios. We find that models struggle to handle information incompleteness, with performance collapsing when shifting from professional summaries to subjective patient narratives. Furthermore, models fail to calibrate their diagnostic decisions, exhibiting biases toward either overly expansive or excessively restrictive diagnostic tendencies, and lack the intrinsic awareness of constraints to define decision boundaries without explicit prompting. These findings confirm that \ours{} poses a substantial diagnostic challenge to current state-of-the-art models. By establishing a rigorous evaluation framework, \ours{} advocates that future research prioritize adaptive, robust decision-making, which is essential for real-world psychiatry.

%% file: Texts/7Limitations.tex
\section{Discussions}

\subsection{\label{limitation}Limitations}
Despite its contributions, this study has several limitations. 

First, \ours{} comprises synthetic clinical scenarios generated using expert-curated diagnostic logic. Although this design enables controlled evaluation of DSM-grounded ambiguity, it does not fully capture the heterogeneity, comorbidity, sociocultural context, or longitudinal uncertainty of real clinical encounters. Furthermore, while we introduce symptom subtypes and prompting strategies to encourage diverse patient expressions, the generated cases remain synthetic approximations and may not fully reflect the variability and complexity of real-world psychiatric presentations. 

Second, the clinical cases in \ours{} are constructed as descriptive scripts rather than interactive dialogues. While real psychiatric diagnosis involves iterative clinician-patient conversations, our benchmark is intentionally designed to evaluate models’ ability to apply clinically grounded diagnostic criteria under controlled and interpretable scenarios. Future work will leverage the constructed knowledge graph to simulate doctor–patient interactions and assess additional diagnostic capabilities such as follow-up question selection. 

Third, all clinical cases are generated exclusively in English, limiting generalizability across languages and cultural contexts where psychiatric symptoms may be expressed differently. However, the knowledge graph encodes DSM-5–based diagnostic logic in a language-independent manner, making its structure applicable across diverse linguistic and cultural settings. This allows future extensions to multilingual and culturally diverse benchmarks through prompt-level control.

Finally, our evaluation focuses on models’ final diagnostic outputs and observable error patterns, without directly examining whether their internal reasoning aligns with DSM-5 diagnostic criteria. While this design enables systematic analysis of how models apply diagnostic knowledge at the output level, it does not capture whether their intermediate reasoning processes follow clinically grounded diagnostic logic. In future work, we plan to augment \ours{} with explicit, criterion-level diagnostic reasoning, allowing the benchmark to more directly assess models’ diagnostic reasoning behaviors.

% 와해된 언어를 못함 (못하는 척을 못하는 이슈). 와해된 언어보다는 좀 희안한 용어를 만드는 식으로 simulate. 이런 건 llm의 한계.
% - 전문가 평가 결과 → 임상적 타당성 좀 더 높일 수 있게 
% - 영어로만 생성되어서 언어/문화적 다양성 제한
% - 모델 추론 과정에 대한 reasoning 없는 것 → future work
% - 실제 인터뷰는 대화 방식 → 우리는 단편적인 narrative → future work

\subsection{Ethical Considerations\label{ethics}}

\ours{} includes only clinically grounded synthesized clinical cases and does not contain any real patient data, medical records, or personally identifiable information. While some synthesized cases describe severe psychiatric symptoms, including suicidal ideation or self-harm, such content is presented as clinically descriptive symptom narratives focusing on patients’ reported experiences, without including explicit descriptions of suicidal methods or self-harm behaviors.

\kg{} does not reproduce DSM-5 text verbatim. Instead, it encodes expert-derived and paraphrased diagnostic logic, including symptom thresholds, temporal constraints, and differential-diagnostic relations, for research evaluation purposes. The released benchmark consists of synthetic case descriptions, labels, and evaluation metadata, and is not intended to substitute for copyrighted clinical manuals or professional diagnostic assessments.

The benchmark is not intended for clinical diagnosis, triage, risk assessment, treatment recommendation, or patient-facing deployment. Rather, \ours{} is designed to evaluate whether LLMs can apply DSM-grounded diagnostic reasoning and appropriately calibrate diagnostic commitment under controlled ambiguity. Furthermore, although \ours{} is designed to reflect a wide range of clinically realistic diagnostic scenarios, it does not capture the full complexity of real-world psychiatric practice, as discussed in the limitations. Consequently, benchmark performance should not be interpreted as evidence of real-world clinical reliability or deployment readiness. 

% First, MENTALBENCH is not intended to evaluate deployable diagnostic systems or replace clinical assessment. Rather, it provides a controlled testbed for studying whether LLMs can apply formal diagnostic logic and calibrate diagnostic commitment under systematically varied ambiguity. 그러므로 우리 벤치마크의 결과를 그대로 뭐 그런 임상 시스템에서의 성능으로 바로 이해하면 안됨 

% \textcolor{red}{리밋, 에띡, AI 사용에 대해서 써야 하는지 아닌지 확인. limitation은 써야 하는 것 같음 (checklist에 있음.)}
%Checklist에서
%- positive and negative societal impacts
%- Does the paper describe safeguards that have been put in place for responsible release of data or models that have a high risk for misuse (e.g., pre-trained language models, image generators, or scraped datasets)?

%% file: Texts/Appendix_0scope_and_expert.tex
\section{Scope of Disorders \label{appen:scope}}
In this section, we provide a comprehensive description of the constructed \kg{} and its constituent elements, with a particular focus on the scope of mental health conditions covered.

\subsection{Limited Scope of Prior Benchmarks}
% The comprehensive scope of our framework stands in marked contrast to prior studies, as detailed in \autoref{tab:related_work}. NLP in mental health has predominantly concentrated on high-prevalence conditions, most notably MDD, GAD, and BD. While some studies extend to Eating Disorders or ADHD, their coverage often lacks diagnostic granularity. As indicated by the gray checkmarks in the table (\textcolor{gray!60}{\Vmark}), previous work frequently aggregates distinct conditions into broad categories (e.g., treating ADHD as a monolithic class without distinguishing subtypes) or overlooks the specific phenomenological nuances required for clinical differentiation. Furthermore, diagnostically challenging conditions (e.g., MDD with psychotic symptoms) are largely absent in existing benchmarks. \ours{} is the first study to address this gap by providing fine-grained coverage across a wide spectrum of 23 distinct disorders, moving beyond broad categorizations to enable a deep, expert-level evaluation of psychiatric reasoning.
\autoref{tab:related_work} presents a comparative analysis of disorder coverage across prominent mental health datasets and benchmarks established in recent years. As illustrated, prior research has predominantly concentrated on high-prevalence conditions, specifically Major Depressive Disorder (MDD), Generalized Anxiety Disorder (GAD), and PTSD. While these conditions are critical, this narrow focus leaves a significant portion of the DSM-5 diagnostic spectrum unexplored in the context of LLM evaluation.

Notably, complex psychopathological profiles and distinct subtypes remain largely underrepresented. For instance, distinctions within Schizoaffective Disorders (Bipolar vs. Depressive types) and specific subtypes of Bipolar Disorder (I vs. II vs. with Psychotic features) are rarely explicitly modeled in existing datasets. Furthermore, conditions such as Delusional Disorder, Specific Phobias, and various Feeding and Eating Disorders (e.g., Binge-Eating Disorder) often lack dedicated evaluation benchmarks.

\input{Tables/Appendix/Knowledge_Graph/related_work_table}

\input{Tables/Appendix/Knowledge_Graph/disorder_list}

\subsection{Scope of \ours{}}
\autoref{tab:disorder_list_full_name} presents the complete list of 23 disorders included in our work. To ensure a holistic evaluation of psychopathological reasoning, we selected representative disorders across a wide spectrum of diagnostic categories defined in the DSM-5. Our coverage encompasses Anxiety Disorders (e.g., GAD, Specific Phobia), Obsessive-Compulsive and Related Disorders (e.g., OCD, Body Dysmorphic Disorder), Trauma- and Stressor-Related Disorders (e.g., PTSD, Adjustment Disorder), Mood Disorders (e.g., MDD, BD I and II), the Schizophrenia Spectrum (e.g., Delusional Disorder, Schizophrenia), Feeding and Eating Disorders (e.g., Anorexia Nervosa, Binge-Eating Disorder), and Attention-Deficit/Hyperactivity Disorder (ADHD).

Crucially, our selection criteria prioritize the replication of authentic challenges encountered in real-world clinical settings, specifically addressing the diagnostic ambiguities posed by symptom overlap and subtle phenomenological distinctions. Since precise differentiation among these conditions is a prerequisite for identifying the primary pathology and determining appropriate therapeutic interventions, we deliberately included disorders where accurate diagnosis hinges on fine-grained clinical details.

For instance, we explicitly distinguish between the three presentations of ADHD---Combined, Inattentive, and Hyperactive---to test the model's ability to identify specific behavioral nuances. Furthermore, we address the complex comorbidities found at the intersection of mood and psychotic symptoms, which represent a significant diagnostic hurdle. To capture this, we include Schizoaffective Disorder (both Bipolar and Depressive types) alongside Mood Disorders with Psychotic Features (BD I with psychotic features and MDD with psychotic features). While these conditions often present with identical symptoms, accurate diagnosis relies on analyzing the duration, concurrence of symptoms, and severity. By including these confounding cases, we evaluate whether LLMs can distinguish fine-grained DSM-grounded diagnostic patterns under controlled ambiguity

Regarding the target demographic, our benchmark focuses on adult clinical narratives. Consequently, we generally exclude neurocognitive and neurodevelopmental disorders (e.g., Alzheimer’s Disease) and conditions primarily diagnosed in early childhood (e.g., Conduct Disorder). However, we make a strategic exception for ADHD. Although classified as a neurodevelopmental disorder with childhood onset, ADHD is frequently first diagnosed in adulthood and serves as a critical differential diagnosis for various mood and psychotic disorders. Therefore, its inclusion is essential for a robust evaluation of adult mental health conditions.

\section{Expert Background and Qualifications\label{appen:expert}}
To ensure clinical rigor and minimize subjective bias during validation, we collaborated with two mental health experts from distinct yet complementary professional domains. As detailed in \autoref{tab:expert_profile}, the team comprises a board-certified psychiatrist and a nationally licensed clinical psychologist, both with extensive experience in diagnostic criteria and differential diagnosis.

\input{Tables/Appendix/Experts/expert_profile}

By integrating the perspectives of a medical psychiatrist and a clinical psychologist, we ensured that the expert-grounded validation reflects a comprehensive understanding of mental health and mitigates the potential bias of a single professional viewpoint. This multidisciplinary validation supports the clinical fidelity of MentalKG and MentalBench as controlled research benchmarks.

%% file: Tables/Appendix/Knowledge_Graph/related_work_table.tex
\newcommand{\rot}[1]{%
  \rule{0pt}{2.5cm}%
  \rotatebox{90}{%
    \parbox{2.5cm}{\raggedright #1}%
  }%
}

\begin{table*}[ht!]
    \centering
    \caption{\label{tab:related_work} Comparison of mental health disorder coverage across related works and \ours{}. Green checkmarks (\textcolor{green!60!black}{\Vmark}) denote disorders explicitly handled by the work, while gray checkmarks (\textcolor{gray!60}{\Vmark}) indicate implicit or partial coverage.}
    \resizebox{\textwidth}{!}{
    \scriptsize
    \begin{tabular}{p{0.2\textwidth}*{23}{c}}
    \toprule
        \multicolumn{1}{r}{\textbf{Disorder Name}} & 
            \rot{ADHD (Combined)} & 
            \rot{ADHD (Inattentive)} &
            \rot{ADHD (Hyperactive)} &
            \rot{Delusional Disorder} & 
            \rot{Schizophrenia} & 
            \rot{Schizoaffective Disorder (Bipolar Type)} &
            \rot{Schizoaffective Disorder (Depressive Type)} & 
            \rot{Bipolar 1} &
            \rot{Bipolar 2} & 
            \rot{Bipolar 1 w/ Psy.} & 
            \rot{Generalized Anxiety Disorder} & 
            \rot{Specific Phobia} &
            \rot{MDD} & 
            \rot{Persistent Depressive Disorder} & 
            \rot{MDD w/ Psy.} & 
            \rot{OCD} &
            \rot{Body Dysmorphic Disorder} & 
            \rot{PTSD} & 
            \rot{Acute Stress Disorder} & 
            \rot{Adjustment Disorder} &
            \rot{Anorexia Nervosa} & 
            \rot{Bulimia Nervosa} & 
            \rot{Binge-Eating Disorder} \\
        \multicolumn{1}{r}{\textbf{Disorder No.}} & 1 & 2 & 3 & 4 & 5 & 6 & 7 & 8 & 9 & 10 & 11 & 12 & 13 & 14 & 15 & 16 & 17 & 18 & 19 & 20 & 21 & 22 & 23 \\
        % Disorder No. & D001 & D002 & D003 & D004 & D005 & D006 & D007 & D008 & D009 & D010 & D011 & D012 & D013 & D014 & D015 & D016 & D017 & D018 & D019 & D020 & D021 & D022 & D023 \\
        \multicolumn{24}{l}{\textbf{Work}} \\
        \midrule
        \work{Coppersmith et al. \cite{coppersmith2015adhd}}{1,5,8,11,13,16,18,23}{2,3,9,14,21,22} % ADHD, Anxiety, BPD, Bipolar, Depression, Eating, OCD, PTSD, Schizophrenia, Seasonal Affective
        \work{MDDL \cite{shen-2017-depression}}{13}{14}
        \work{RSDD \cite{yates-etal-2017-depression}}{13}{14}
        \work{SMHD~\cite{cohan2018smhd}}{1,11,8,13,16,18,5,23}{2,3,9,14,21,22} % ADHD, Anxiety, Autism, Bipolar, BPD, Eating, OCD, PTSD, Schizophrenia, Seasonal Affective
        \work{Gaur et al. \cite{gaur2018let}}{8,11, 13, 5}{9, 14}
        \work{Pirina \& {\c{C}}{\"o}ltekin\cite{pirina-coltekin-2018-identifying}}{13}{14}
        \work{Murarka et al. \cite{murarka-etal-2021-classification}}{13,11,18,1,8}{14,2,3,9} % Depression, Anxiety, PTSD, ADHD, Bipolar
        \work{Kayalvizhi\&Thenmozhi\cite{kayalvizhi-2022-data}}{13}{14} % depression
        \work{DepSign \cite{poswiata-perelkiewicz-2022-opi}}{13}{14}
        \work{Ji et al. \cite{ji2022suicidal}}{8,11,13,18}{9,14}
        \work{PsySym \cite{zhang2022symptom}}{13,11,1,8,16,18,23}{2,3,14,21,22} % Depression, Anxiety, ADHD, Bipolar Disorder, OCD, PTSD, Eating Disorder
        \work{D4 \cite{yao-etal-2022-d4}}{13}{14}   % depression, suicidal risk
        \work{Tutun et al. \cite{tutun2023ai}}{5,11,12,13,16,17}{14} % anxiety, depression, hostility, interpersonal sensitivity, obsessive-compulsive, paranoid, phobic anxiety, psychoticism, somatization, ... (focusing on symptoms)
        \work{Song et al. \cite{song2023simple}}{8,11,13}{9, 14}
        \work{KoMOS \cite{kang2024cure}}{11,13,23}{14, 21,22}
        \work{MentalRiskES \cite{marmol-romero-etal-2024-mentalriskes}}{11,13,23}{14,21,22}
        \work{eRisk \cite{parapar-2025-erisk}}{1,13,21,23}{2,3,14,22}
        \work{MHQA~\cite{racha2025mhqa}}{11,13,16,18}{14} % anxiety, depression, trauma, obsessive and compulsive issue
        \work{Mentalchat16K \cite{xu2025mentalchat16k}}{11,13,18,23}{14}
        \work{\ours (\textbf{Ours})}{1,2,3,4,5,6,7,8,9,10,11,12,13,14,15,16,17,18,19,20,21,22,23}{}
    \bottomrule
    \end{tabular}
    }

\end{table*}

%% file: Tables/Appendix/Knowledge_Graph/disorder_list.tex
\begin{table}[ht!]
    \centering
    \caption{\label{tab:disorder_list_full_name}List of disorders covered in \ours{}}
    \resizebox{\linewidth}{!}{
    \small
    \begin{tabular}{cp{0.33\linewidth}p{0.6\linewidth}}
    \toprule
        \textbf{No.} & \textbf{Abbreviation} & \textbf{Full Name} \\
    \midrule
        1 & ADHD (Combined) & Attention-Deficit/Hyperactivity Disorder (Combined Presentation) \\
        2 & ADHD (Inattentive) & Attention-Deficit/Hyperactivity Disorder (Predominantly Inattentive Presentation) \\
        3 & ADHD (Hyperactive) & Attention-Deficit/Hyperactivity Disorder (Predominantly Hyperactive/Impulsive Presentation) \\
        4 &  & Delusional Disorder \\
        5 &  & Schizophrenia \\
        6 & Schizoaffective (Bipolar Type) & Schizoaffective Disorder (Bipolar Type)\\
        7 & Schizoaffective (Depressive Type) & Schizoaffective Disorder (Depressive Type) \\
        8 & BD 1 & Bipolar I Disorder \\
        9 & BD 2 & Bipolar II Disorder \\
        10 & BD 1 w/ Psych. & Bipolar I Disorder with Psychotic Features\\
        11 & GAD & Generalized Anxiety Disorder \\
        12 &  & Specific Phobia \\
        13 & MDD & Major Depressive Disorder \\
        14 & PDD & Persistent Depressive Disorder \\
        15 & MDD w/ Psych. & Major Depressive Disorder with Psychotic Features \\
        16 & OCD & Obsessive-Compulsive Disorder \\
        17 &  & Body Dysmorphic Disorder \\
        18 & PTSD & Posttraumatic Stress Disorder \\
        19 &  & Acute Stress Disorder \\
        20 & AD & Adjustment Disorder \\
        21 &  & Anorexia Nervosa \\
        22 &  & Bulimia Nervosa \\
        23 &  & Binge-Eating Disorder \\
    \bottomrule
    \end{tabular}
    }

\end{table}

%% file: Tables/Appendix/Experts/expert_profile.tex
\begin{table}[th!]
\centering
\small
\caption{\label{tab:expert_profile}
% Professional background of psychiatric experts involved in knowledge graph construction and benchmark validation. All experts are licensed psychiatrists with direct clinical experience relevant to diagnostic criteria and differential diagnosis.
Professional background of mental health experts involved in knowledge graph construction and benchmark validation. The team consists of a board-certified psychiatrist and a licensed clinical psychologist with direct clinical experience relevant to diagnostic criteria and differential diagnosis.
}
\begin{tabularx}{\columnwidth}{lX}
\toprule
\textbf{Expert} & \textbf{Professional Background} \\
\midrule
Expert 1 &
Board-certified psychiatrist with extensive clinical experience.
Evaluated the naturalness, validity, and realism of the constructed benchmark based on DSM-5 standards and clinical expertise. \\

Expert 2 &
% Clinical psychologist 
% Board-certified psychiatrist specializing in psychotic and bipolar disorders.
% Reviewed and refined diagnostic rules and ambiguity resolution cases in the benchmark construction process. \\
Nationally licensed clinical psychologist.
Provided advisory support on aligning the benchmark with real-world clinical challenges. Reviewed the diagnostic criteria within the knowledge graph for DSM-5 compliance and evaluated the naturalness, validity, and realism of the benchmark. \\

\bottomrule
\end{tabularx}

\end{table}

%% file: Texts/Appendix_1KG.tex
\section{Details on Knowledge Graph}
\label{appen:kg}

\subsection{Further Details on Nodes}
\paragraph{Symptom Nodes.}
\kg{} comprises 84 symptom nodes ($S$) covering the diagnostic criteria for 23 mental disorders. Each symptom node contains two attributes: a textual description describing the clinical meaning of the symptom, and a set of subtypes representing its specific manifestations. 

\paragraph{Symptom Group Nodes.}
The DSM-5 often organizes symptoms into clinically relevant groups, such as \textit{depressive} and \textit{manic episodes}. To represent this hierarchy, we introduce 23 symptom group nodes ($G$), which aggregate related symptom nodes. 

\paragraph{Disorder Nodes.}
\kg{} contains 23 disorder nodes ($D$) corresponding to specific mental disorders in DSM-5, which we select in consultation with psychiatrists based on two criteria: (1) high clinical prevalence, (2) need for differential diagnosis from other disorders. We map DSM-5 diagnostic criteria for each disorder in \kg{} to a structured graph representation, connecting disorder nodes ($D$) directly to symptom group nodes ($G$) to represent the symptoms required for diagnosis (details are in \autoref{fig:graph_schema}). For more fine-grained diagnostic requirements, such as minimum symptom thresholds and duration constraints, we encode them as internal attributes of the disorder nodes. Through this process, the natural language descriptions of DSM-5 are effectively transformed into structurally formalized information in the graph. 

\paragraph{Differential Diagnosis Nodes.}
In clinical practice, patients' self-reports rarely perfectly match the diagnostic criteria of a single disorder. Challenges often arise when symptoms overlap, or clinical evidence is inconclusive; for instance, a patient presenting with depression who also exhibits manic features requires careful differentiation between \textit{Major Depressive Disorder} and \textit{Bipolar II Disorder}. To address this, we introduce differential diagnosis nodes ($D_d$) connecting a primary suspicion ($D_1$) to an alternative ($D_2$) via a directed edge. This directionality captures scenarios where a patient resembles $D_1$, but specific confounding contexts (e.g., ambiguous duration or shared symptoms) necessitate the evaluation of $D_2$. 

Each node internally encodes the differential diagnostic criteria using the following three attributes: (1) \textit{Triggering Condition}, which specifies additional symptoms that prompt the need for differential diagnosis; (2) \textit{Key Difference}, which represents the core points of divergence between the two disorders; and (3) \textit{Discriminating Rules}, which specify conditional decision rules for determining the final diagnosis. For example, when depressive episodes co-occur with elevated mood, differential diagnosis between MDD and BD becomes necessary. The key difference lies in whether manic symptoms collectively satisfy the criteria for a Manic Episode: if so, BD is diagnosed; otherwise, MDD. \autoref{tab:diff_graph_logic} illustrates this example.

\subsection{Examples of Knowledge Graph}

As mentioned in Section~\ref{sec:3_knowledge graph}, \kg{} comprises four distinct types of nodes. To facilitate a clearer understanding of the graph's architecture, we visualize the full schematic structure of \kg{} in \autoref{fig:kg_full} and illustrate a representative subgraph of \kg{} in \autoref{fig:kg}. Below, we detail the specific attributes encoded within each node type, provided with their descriptions and examples.

\input{Figures/1Introduction/kg_overview}
\input{Figures/Appendix/knowledge_graph_full}
\clearpage

\paragraph{Disorder.}
Each disorder node encapsulates the core diagnostic prerequisites and exclusion criteria defined in the DSM-5.

\input{Tables/Appendix/KG_examples/disorder}

\paragraph{Symptom Group.}
These nodes represent logical clusters of symptoms (e.g., episodes) that serve as intermediate diagnostic layers.
\input{Tables/Appendix/KG_examples/symptom_group}

\paragraph{Symptom.}
Symptom nodes capture atomic clinical observations and their diverse phenotypic manifestations.
\input{Tables/Appendix/KG_examples/symptom}

\paragraph{Differential Diagnosis.}
These nodes encode the directional logic required to distinguish between confounding disorders.
\input{Tables/Appendix/KG_examples/differential_diagnosis}

For understanding, we provide more examples in \autoref{tab:diff_graph_logic}.
\input{Tables/Appendix/KG_examples/differential_diagnosis_tab}

\subsection{Details on Expert Validation\label{appen:kg_ev}}
% 구축한 \kg{}의 임상적 신뢰성을 보장하기 위해, 그래프 구축은 전문가와의 논의를 통해 진행되었다. 구체적으로, 그래프 구축된 그래프에 대해 전문가는 다음과 같은 항목에 대한 검증을 수행하였다. 
To ensure the clinical reliability of the constructed \kg{}, the entire development process is conducted in collaboration with mental health experts. Specifically, the experts perform a rigorous validation focusing on three key aspects:

% (1) the clinical appropriateness of symptom subtypes within symptom nodes 증상 노드에 포함된 해당 증상의 manifestation 이 해당 노드를 잘 표현하고 있는지 검증, 이 과정에서 의미가 명확하지 않거나, 다른 노드들과 겹치는 subtype들은 제거되거나 재배치되었음. 예를 들어, 절망감(S000)은 우울감(S000) 노드의 subtype이었으나, 지속성우울장애의 진단기준 구현을 위해 별도의 증상 노드로 분리됨 
\paragraph{1) Clinical Appropriateness of Subtypes within Symptom Nodes.} 
Experts verify whether the phenotypic manifestations (\textbf{Subtypes}) listed under each symptom node accurately represented the core clinical concept of that node. During this process, subtypes that are semantically ambiguous or exhibited significant overlap with other nodes are removed or reassigned. For instance, \textit{Hopelessness} (S083), initially categorized as a subtype of the \textit{Depressed Mood} (S024) node, is separated into a distinct symptom node to accurately implement the specific diagnostic criteria for \textit{Persistent Depressive Disorder}.

% (2) the alignment of the formalized diagnostic criteria with DSM-5 standards: 증상 노드와 증상 그룹 노드 간의 연결과 내재된 attribute로 구현된 각 질병에 대한 진단 기준이 dsm-5의 것을 잘 반영하는지 검토하였으며, 이 과정에서 각 질병간의 boundary를 명확하게 하기 위한 additional requirements의 주입 등이 이루어졌음. 
\paragraph{2) Alignment with DSM-5 Standards.}
Experts review the structural connections between Symptom and Symptom Group nodes, as well as the logic encoded in node attributes, to ensure they faithfully reflect the formal diagnostic criteria of the DSM-5. During this review, experts manually encoded \textit{Additional Requirements} (e.g., exclusion criteria) into disorder nodes to sharpen the decision boundaries between clinically similar conditions if needed.

% (3) the clinical validity of the differential diagnosis logic: dsm-5내의 감별진단 규칙은 구조화되지 않은 텍스트로 표현되어 있으며 우리는 이를 logic형태로 변환함. 전문가 검토를 통해 각 감별규칙이 실제 두 질환간의 혼동 양상 및 판별 룰을 적절히 반영하고 있는지 검토하였으며, 이 과정에서 모호한 규칙은 수정되거나 삭제되었으며, 보다 명확한 판별 룰을 위한 조건들이 추가되었음. 
\paragraph{3) Clinical Validity of Differential Diagnosis Logic.} 
Since DSM-5 differential diagnosis guidelines are originally presented as unstructured text, we converted them into structured logical rules. Experts evaluate the discriminating rules within differential diagnosis nodes to ensure they appropriately capture real-world patterns of diagnostic confusion and the discrimination logic between paired disorders. Through this iterative review, ambiguous rules are refined or discarded, and specific conditional contexts are added to establish clearer discrimination criteria.

% 이러한 과정은 우리의 지식그래프가 23개의 증상에 걸친 DSM-5의 진단체계 및 감별진단을 높은 임상적 수준에서 구조화된 형태로 변환할 수 있게 함.
This rigorous validation process ensures that \kg{} successfully transforms the complex diagnostic systems and differential diagnoses of the DSM-5 across 23 disorders into a structured format with a high level of clinical fidelity.

%% file: Figures/1Introduction/kg_overview.tex
% \begin{figure}
%     \centering
%     % \includegraphics[width=\columnwidth]{Figures/1Introduction/overview_of_kg.pdf}
%     \includegraphics[width=\columnwidth]{Figures/1Introduction/figure8_cropped.pdf}
%     \caption{\label{fig:kg} \small
%     A subgraph of \kg{} illustrating the diagnostic structure of Major Depressive Disorder (MDD), Bipolar I Disorder, and Bipolar II Disorder.} 
%     % \vspace{-0.2in}
% \end{figure}

\begin{figure*}[bht!]
  \centering
    \includegraphics[width=0.65\linewidth]{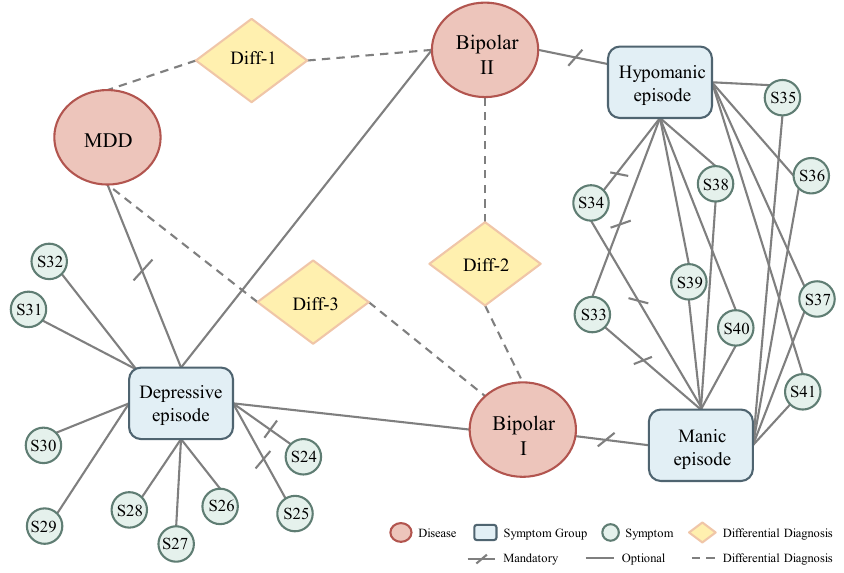}
  \caption{\label{fig:kg} 
    A subgraph of \kg{} illustrating the diagnostic structure of Major Depressive Disorder (MDD), Bipolar I Disorder, and Bipolar II Disorder.} 
\end{figure*}

%% file: Figures/Appendix/knowledge_graph_full.tex
\begin{figure*}[thb!]
    \centering
    \includegraphics[width=\textwidth]{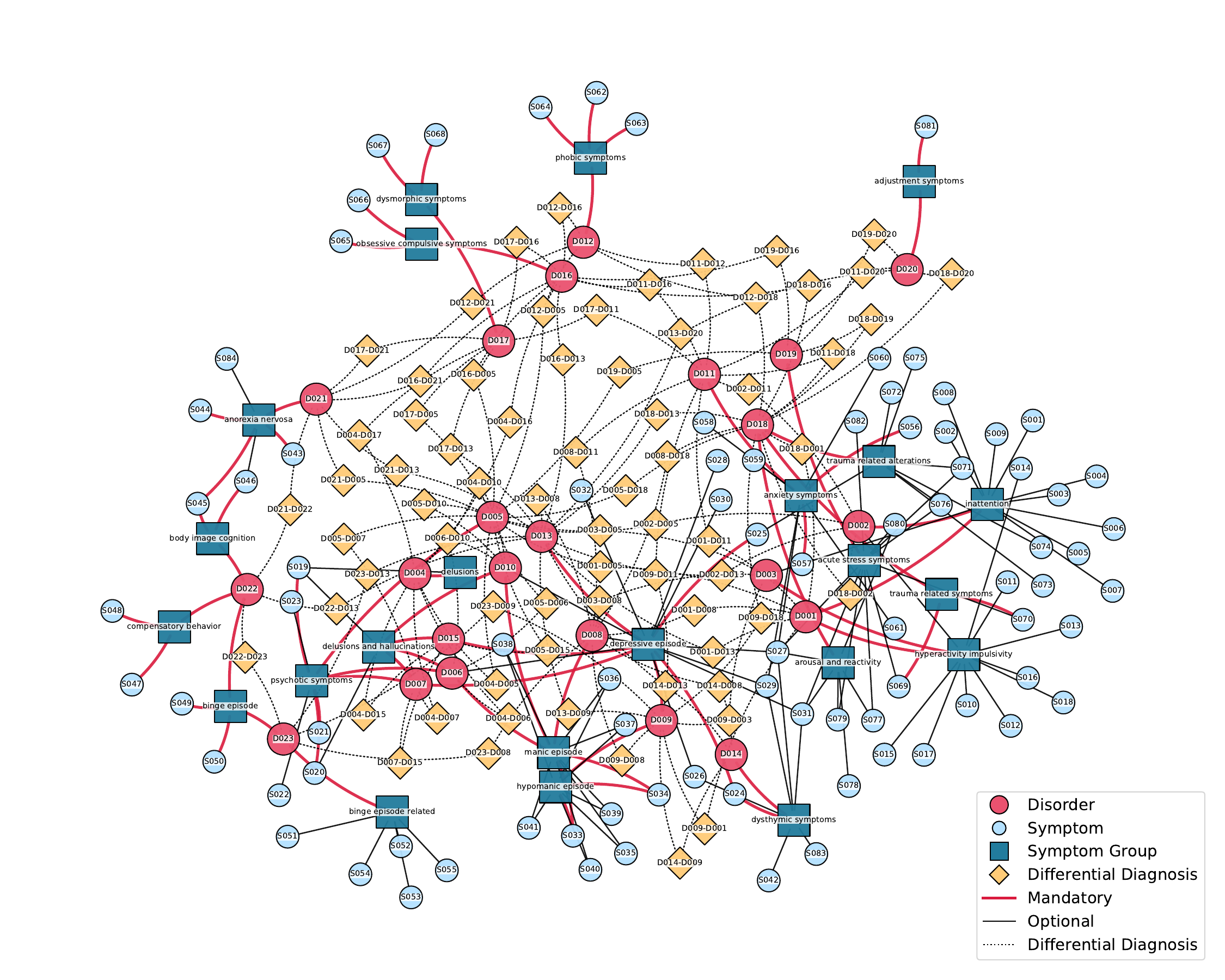}
    \caption{\label{fig:kg_full} The full topology of our constructed \kg{}.}
    
\end{figure*}

%% file: Tables/Appendix/KG_examples/disorder.tex
\begin{itemize}[itemsep=0pt]
    \kgeg{Name}{The standardized clinical designation of the condition}{
        \sublist{Major Depressive Disorder}
        }
    \kgeg{Code}{Unique identifier within the Knowledge Graph}{
        \sublist{D013}
    }
    \kgeg{Minimum Duration}{The mandatory temporal threshold required for the diagnosis}{\sublist{None}}
    \kgeg{Maximum Duration}{The upper temporal limit for the condition's validity}{\sublist{None}}
    \kgeg{Functional Impairment}{Indicates whether the condition entails clinically significant distress or impairment}{\sublist{true}}
    \kgeg{Traumatic Stressor}{Specifies if exposure to a traumatic event is a prerequisite}{\sublist{False}}
    \kgeg{Psychosocial Stressor}{Specifies if a specific psychosocial stressor is a prerequisite}{\sublist{False}}
    \kgeg{Additional Requirements}{Supplementary exclusion criteria or specific conditions necessary for a valid diagnosis}{\sublist{None}}
\end{itemize}

%% file: Tables/Appendix/KG_examples/symptom_group.tex
\begin{itemize}[itemsep=0pt]
    \kgeg{Name}{The standardized label for the clinical symptom cluster}{\sublist{Depressive Episode}}
    \kgeg{Minimum Duration}{The mandatory temporal threshold required to satisfy the episode criteria}{\sublist{2 weeks}}
    \kgeg{Maximum Duration}{The upper temporal limit for the episode's validity }{\sublist{None}}
    \kgeg{Functional Impairment}{Indicates whether the symptom cluster entails clinically significant distress or impairment}{\sublist{True}}
    \kgeg{Minimum Number of Required Symptoms}{The quantitative threshold of total symptoms required for diagnosis}{\sublist{5}}
    \kgeg{Required Core Symptoms}{Logical constraint specifying essential symptoms that must be present}{\sublist{one of the core symptoms}}
    % \kgeg{Must-Included Symptoms}{}{\sublist{S024, S025}}
    % \kgeg{Symptoms}{}{\sublist{S024, S025, S026, S027, S028, S029, S030, S031, S032}}
\end{itemize}

%% file: Tables/Appendix/KG_examples/symptom.tex
\begin{itemize}[itemsep=0pt]
    \kgeg{Name}{The standardized clinical label of the symptom}{
        \sublist{Loss of Interest}}
    \kgeg{Code}{Unique identifier of the symptom within the Knowledge Graph}{\sublist{S025}}
    \kgeg{Category}{The category of the symptom}{\sublist{Depressive Episode}}
    \kgeg{Description}{Formal clinical definition and description characterizing the core experience}{
    \sublist{A state where interest or pleasure in previously enjoyed activities is markedly diminished}
    }
    \kgeg{Subtypes}{Specific behavioral or emotional manifestations (phenotypes)}{
            \sublist{Loss of Interest: Reduced interest in previously enjoyed hobbies or leisure activities.}
            \sublist{Social Withdrawal: Reduced interest or effort in social interactions and maintaining relationships.}
            \sublist{Loss of Motivation: Lack of motivation to initiate or complete work or tasks.}
            \sublist{Anhedonia: Inability to experience positive emotions such as happiness, satisfaction, love, or pleasure.}
            \sublist{Indifference to Achievement: Insensitivity or indifference to personal achievements or success.}
    }
\end{itemize}

%% file: Tables/Appendix/KG_examples/differential_diagnosis.tex
\begin{itemize}[itemsep=0pt]
    \kgeg{Disorder}{The primary disorder ($D_1$) initially suspected based on presenting symptoms}{
        \sublist{Major Depressive Disorder}
        }
    \kgeg{Differential Target}{The alternative disorder ($D_2$) that necessitates evaluation due to overlapping features or ambiguity}{
        \sublist{Adjustment Disorder}
    }
    \kgeg{Code}{Unique identifier for the directed differential diagnosis edge ($D_1 \rightarrow D_2$)}{\sublist{D013-D020}}
    \kgeg{Triggering Condition}{}{
        \sublist{Experience of psychosocial stress}
    }
    \kgeg{Key Difference}{The specific clinical context or symptom overlap that initiates the differential process}{
        \sublist{Meeting criteria for manic episode and history}
    }
    \kgeg{Discriminating Rules}{The critical distinguishing factor or exclusion criteria between the two disorders}{
        \sublist{\textbf{MDD} (D013): Depressive symptoms may appear after a psychosocial stressor, and the full criteria for a major depressive episode are met}
        \sublist{\textbf{AD} (D020): Depressive symptoms appeared after a psychosocial stressor, but the criteria for a major depressive episode are not met}
    }
    
\end{itemize}

%% file: Tables/Appendix/KG_examples/differential_diagnosis_tab.tex
\begin{table*}[ht!]
\centering
\caption{\label{tab:diff_graph_logic}Differential diagnosis logic for Major Depressive Disorder (MDD) derived from the knowledge graph structure. The table schema corresponds to the differential nodes connecting MDD to Bipolar disorders.}
\scriptsize
\begin{tabularx}{\textwidth}{l p{3cm} p{3.5cm} X}
\toprule
\textbf{\begin{tabular}[c]{@{}l@{}}Differential\\ Target\end{tabular}} & 
\textbf{Triggering Condition} & 
\textbf{Key Difference} & 
\textbf{Discriminating Rules} \\
\midrule
\textbf{Bipolar I} & 
Presence of Manic Symptoms & 
Meeting criteria for manic episode and history & 
\textbf{MDD (D013):} Manic-like symptoms present, but criteria are not met, and no history of mania. \newline
\textbf{Bipolar I (D008):} Manic symptoms present, criteria for manic episode met, or history of mania exists. \\
\midrule
\textbf{Bipolar II} & 
Presence of Hypomanic Symptoms & 
Meeting criteria for hypomanic episode and history & 
\textbf{MDD (D013):} Hypomanic-like symptoms present, but criteria are not met, and no history of hypomania. \newline
\textbf{Bipolar II (D009):} Hypomanic symptoms present, and history of hypomania exists. \\
\bottomrule
\end{tabularx}

\end{table*}

%% file: Texts/Appendix_2Benchmark.tex
\section{Details on Benchmark}
\subsection{Linguistic Divergence by Presentation Style\label{appen:linguistic}}
To compare the linguistic differences between the professional summaries (\textit{Type 1}) and the patient self-reports in (\textit{Type 2}), we analyze how these distinct presentation styles pose unique challenges for diagnostic evaluation. \autoref{fig:tsne_by_code} visualizes the sentence embeddings of \textit{Type 1} and \textit{Type 2} samples across 23 disorders using t-SNE, with each category color-coded to highlight its semantic distribution. The embedding space analysis highlights three key insights into the linguistic characteristics and challenges within the dataset.

\paragraph{Stylistic Discrepancy.} Even for the same psychiatric disorder, \textit{Type 1} and \textit{Type 2} samples are positioned in entirely different regions of the embedding space. This indicates that the stylistic shift from formal medical language to informal speech creates a gap, posing a major challenge for models to apply their internal knowledge across different levels of linguistic formality.
\paragraph{Clarity through Formal Terminology.} \textit{Type 1} samples form highly distinguished and well-separated clusters. The use of structured medical summaries and explicit clinical keywords enables clear, distinct descriptions of disorders, making the diagnostic task in this format more amenable to straightforward discrimination.
\paragraph{Semantic Overlap and Clinical Ambiguity.} In contrast, \textit{Type 2} samples exhibit substantial overlap and form blurred clusters. Because different mental health conditions often share similar colloquial descriptions, this overlap poses a significant challenge for models in accurately resolving clinical boundaries between competing diagnoses.

\clearpage
\input{Figures/Appendix/Language/tsne_figs}
\clearpage

\subsection{\label{appen:complexity}Analysis of Semantic Overlap in Differential Diagnosis}
\autoref{fig:tsne_type34} visualizes the sentence embeddings of \textit{Type 3} (controlled ambiguity) and \textit{Type 4} (definitive diagnosis) samples. Unlike the clear stylistic discrepancy observed between professional summaries and colloquial narratives, the embedding space for \textit{Types 3} and \textit{4} exhibits high semantic convergence. The red (\textit{Type 3}) and blue (\textit{Type 4}) clusters are heavily interleaved across the entire distribution.

This visual overlap highlights the intrinsic complexity of the differential diagnosis task. The proximity in the embedding space confirms that the transition from a multiple-answer scenario (\textit{Type 3}) to a single-answer scenario (\textit{Type 4}) is driven by specific, high-level clinical logic rather than by a shift in the data distribution. Consequently, models must perform precise DSM-grounded inference to determine whether the discriminatory evidence required to resolve the diagnostic boundary is present or absent.

%To analyze the diagnostic complexity of \textit{Type 4} (derived from differential diagnosis nodes) relative to \textit{Type 2} (derived from single disease nodes), we compare t-SNE visualizations of sentence embeddings for both types (in \autoref{fig:tsne_type24}). The visual distributions reveal a stark contrast in semantic convergence between the two types. As shown in the upper subplot, \textit{Type 2} exhibits distinct, well-defined clusters for each disorder category. This indicates that narratives based on single disease nodes are characterized by clearly delineated semantic boundaries, with the clinical features of each condition uniquely represented. In contrast, the distribution in \textit{Type 4} is much more scattered and interleaved across the embedding space. This increased dispersion reflects the high degree of semantic overlap inherent in differential diagnosis, where multiple conditions share closely related symptomatic presentations. This indicates that \textit{Type 4} inherently contains logical complexity beyond what is captured by semantic distribution, requiring models to navigate diagnostic boundaries through rigorous DSM-based reasoning.

\input{Figures/Appendix/Language/tsne_type3vs4}

\clearpage
%\input{Figures/Appendix/Language/tsne_type2vs4}
%\clearpage

\subsection{Details on Seed Generation\label{appen:seed_generation}}

\paragraph{Seed Generation Principles.}
The seed generation process aims to construct structured clinical profiles that serve as the diagnostic foundation of each case. Using \kg{} as the logical backbone, we sample diagnostic components encoded in the graph, including mandatory symptoms, minimum symptom thresholds, duration constraints, functional impairment requirements, and additional diagnostic conditions. These components are selectively controlled according to the objective of each case type, allowing the benchmark to systematically vary information completeness and diagnostic complexity.

\noindent\textbf{Type 1.}
Seeds are designed to represent professional summaries in which diagnostically relevant information has been sufficiently collected through structured clinical interviews. Accordingly, each seed is constructed to explicitly satisfy the DSM-5 criteria of the target disorder. Mandatory symptoms are always included, symptom counts are sampled at or above the required diagnostic threshold, and additional components such as duration, functional impairment, and diagnostic conditions are explicitly specified. For example, if a disorder requires at least six symptoms, the seed contains six or more symptoms within a constrained upper range.

\noindent\textbf{Type 2.}
Seeds are designed to reflect patient self-report scenarios, where individuals typically describe their condition through chief complaints rather than systematically reporting every diagnostic criterion. To avoid overly dense and unrealistic narratives, Type 2 preserves the core characteristics of the disorder while intentionally reducing information completeness. This reduction is primarily applied to the symptom-count criterion. For example, when a disorder requires at least six symptoms, Type 2 seeds may include approximately three to six symptoms. However, mandatory symptoms, duration constraints, and additional diagnostic conditions are preserved, as removing these elements could substantially weaken the clinical plausibility of the disorder. For instance, a depressive profile without depressed mood or loss of interest, or with symptoms lasting only a very brief duration, would no longer plausibly represent depressive pathology.

\noindent\textbf{Type 3 \& Type 4.}
Seeds are constructed from the Differential Diagnosis nodes in \kg{} to model diagnostic complexity. We first instantiate a Type 2-style profile for the target disorder and then augment it with the triggering condition associated with a differential-diagnosis pair, creating a scenario in which competing diagnoses remain clinically plausible. In Type 3, the discriminating evidence required to resolve the differential diagnosis is intentionally omitted, allowing multiple diagnoses to remain consistent with the presented evidence. In Type 4, we further incorporate the key differences and discriminative rules encoded in the differential-diagnosis node. These additional clues resolve the ambiguity and support a single diagnosis under the encoded rule.

Overall, this seed generation strategy allows \ours{} to move beyond simple symptom combination by explicitly controlling which diagnostic evidence is present, incomplete, or discriminative. This design enables the benchmark to represent both incomplete patient disclosures and clinically grounded differential-diagnosis scenarios.

\paragraph{Expert Validation Procedure.}
After generation, all seeds underwent manual review and refinement by two experts. The validation focused on whether each seed was clinically coherent, whether it appropriately represented the intended target disorder, and whether the assigned gold label was consistent with the provided diagnostic options. 

For \textbf{Type 1}, most seeds were retained without substantial modification because they were generated to fully satisfy the encoded diagnostic criteria. For \textbf{Type 2}, some seeds required refinement because undersampling could occasionally remove clinically representative symptom patterns, produce overly atypical presentations, or combine symptoms that rarely co-occur in practice. In such cases, experts revised the symptom composition while preserving the intended level of information incompleteness.

For \textbf{Type 3} and \textbf{Type 4}, experts additionally examined whether the added differential-diagnosis conditions were clinically compatible with the base profile. They verified that Type 3 seeds preserved the intended ambiguity between competing diagnoses and that Type 4 seeds contained sufficient discriminating evidence to support a single diagnosis. Through this iterative review process, we ensured that the final seeds remained clinically plausible while faithfully reflecting the DSM-grounded diagnostic logic encoded in \kg{}.

\paragraph{Seed Examples.} 
To further illustrate the construction process, we provide representative seed examples for each case type. Specifically, Tables~\ref{tab:type1_seed_example} and \ref{tab:type2_seed_example} present example seeds for Major Depressive Disorder (MDD) corresponding to Type 1 and Type 2, respectively, demonstrating how information completeness is controlled across the two settings.

In addition, Tables~\ref{tab:type3_seed_example} and \ref{tab:type4_seed_example} present representative differential-diagnosis seeds involving Major Depressive Disorder and Bipolar I Disorder. These examples illustrate how Type 3 preserves diagnostic ambiguity by omitting discriminating evidence, whereas Type 4 resolves the ambiguity through the inclusion of key differential diagnostic clues and discriminative rules.

% Type 1 (MDD) 
\input{Tables/Appendix/seed_examples/seed_type1}

% Type 2 (MDD)
\input{Tables/Appendix/seed_examples/seed_type2}

% Type 3 (MDD & Bipolar I)
\input{Tables/Appendix/seed_examples/seed_type3}

% Type 4 (MDD & Bipolar I) -> MDD 
\input{Tables/Appendix/seed_examples/seed_type4}

\clearpage

\subsection{Details on Profile Augmentation\label{appen:demographic}}
\paragraph{Subtypes from \kg{}.}
Each symptom node in \kg{} is associated with specific "subtypes," which represent behavioral or emotional manifestations (phenotypes) curated by clinical experts. These subtypes provide detailed descriptions of how a particular symptom manifests across clinical contexts.

This hierarchical structure is essential for ensuring the diversity of the generated data. During the profile augmentation process, even when the same set of symptoms is selected for a profile, the framework can generate heterogeneous clinical scenarios by sampling different subtypes. This mechanism enables the benchmark to capture the diverse and nuanced ways psychiatric symptoms present in real-world patients. Examples of subtypes are shown below:

\textbf{Negative\_Symptoms (S023)}
\begin{itemize}
    \item \textbf{Diminished Emotional Expression:} A state of reduced emotional expression in facial expressions, eye contact, intonation of speech, and movements.
    \item \textbf{Avolition:} A decrease in motivated self-initiated purposeful activities.
    \item \textbf{Alogia:} A state where the quantity of spontaneous speech is significantly reduced.
    \item \textbf{Anhedonia:} A decrease in the ability to experience pleasure or interest in enjoyable activities.
    \item \textbf{Asociality:} A state where interest in social activities or interpersonal relationships is significantly reduced.
\end{itemize}

\textbf{Loss\_of\_Interest (S025)}
\begin{itemize}
    \item \textbf{Loss of Interest:} Reduced interest in previously enjoyed hobbies or leisure activities.
    \item \textbf{Social Withdrawal:} Reduced interest or effort in social interactions and maintaining relationships.
    \item \textbf{Loss of Motivation:} Lack of motivation to initiate or complete work or tasks.
    \item \textbf{Anhedonia:} Inability to experience positive emotions such as happiness, satisfaction, love, or pleasure.
    \item \textbf{Indifference to Achievement:} Insensitivity or indifference to personal achievements or success.
\end{itemize}

\textbf{Death\_Suicide (S032)}
\begin{itemize}
    \item \textbf{Recurrent Thoughts of Death:} Passive wishes for death (e.g., ``I wish I would not wake up'') without active suicidal intent.
    \item \textbf{Suicidal Ideation:} Recurring thoughts such as ``I want to die'' or ``I want to kill myself,'' without a specific suicide plan.
    \item \textbf{Suicidal Impulse:} Sudden and intense urges to engage in potentially lethal actions (e.g., jumping from a height or running into traffic) without a concrete plan.
    \item \textbf{Specific Suicide Plan:} Formulating specific methods or scenarios regarding when, where, and how to attempt suicide.
    \item \textbf{Suicide Attempt:} Actual self-harm behaviors intended to end one’s life, or preparatory actions such as hoarding pilㄱls or writing a will.
\end{itemize}

\textbf{Excessive\_Anxiety\_Worry (S056)}
\begin{itemize}
    \item \textbf{Chronic Worry:} Worry occurs more days than not for at least 6 months.
    \item \textbf{Generalized Worry:} Recurring thoughts such as ``I want to die'' or ``I want to kill myself,'' without a specific suicide plan.
    \item \textbf{Catastrophizing:} Jumping to conclusions that minor mistakes or events will lead to irreversible disasters.
    \item \textbf{Intolerance of Uncertainty:} Unable to tolerate future uncertainty, trapped in 'what if' questions.
\end{itemize}

\paragraph{Demographic Sampling.}
We transform the previously constructed symptom profiles into concrete patient profiles by assigning demographic information. Specifically, we consider four demographic attributes: age, gender, occupation, and marital status. To generate natural and realistic demographic combinations, we use Gemini-2.5-Flash to produce 300 plausible combinations of these four attributes. Examples of the generated demographic combinations are shown below. 

\begin{itemize}
\item \{"Age": 59, "Gender": "Male", "Occupation": "Farmer", "Marital Status": "Married"\},
\item \{"Age": 26, "Gender": "Female", "Occupation": "Emergency Room Nurse", "Marital Status": "Single"\}.
\end{itemize}

For each symptom profile, we randomly sample one of the pre-generated demographic combinations and assign it to the profile. This procedure enables the construction of diverse clinical scenarios grounded in realistic patient demographics.

\subsection{\label{appen:statistics}Data Statistics}
\input{Tables/Dataset/data_stat}

As mentioned in Section~\ref{sec:4_mentalbench}, \ours{} comprises 24,750 clinical cases and their corresponding question-answer pairs. The detailed statistics of the generated data are presented in \autoref{tab:data_stats}.

For \textbf{Type 1 and Type 2}, the generation process covers all 23 distinct disorders. We first generate 5 and 10 clinical seeds per disorder, respectively. For each seed, we create 5 diverse patient profiles. Clinical cases are then synthesized by pairing each seed with a profile. To ensure linguistic and structural diversity, each profile-seed combination is processed by 3 different LLMs (generators), yielding the final case sets (e.g., $23 \times 5 \times 5 \times 3 = 1,725$ cases for Type 1).

For \textbf{Type 3 and Type 4}, the generation is anchored in the 87 differential diagnoses. Type 3 models ambiguous clinical cases where the presenting symptoms cannot fully differentiate between two conditions, resulting in dual correct answers (multiple answers). Type 4, on the other hand, builds upon the base seeds of Type 3 by adding specific discriminatory rules to resolve the diagnostic ambiguity. Because we synthesize distinct clinical cases that definitively point to either Disorder A or Disorder B for each differential pair, the generation targets $87 \times 2$ conditions. Following the same multiplier for seeds, profiles, and generators, this process yields 13,050 cases for Type 4.

\subsection{\label{appendix:option}Option Design and Distractor Generation}
Option design for each question is critical for ensuring the reliability of the benchmark. Since \ours{} aims to measure whether LLMs can accurately distinguish between disorders with overlapping symptoms, option construction must simultaneously satisfy two conditions: (1) each question must admit a clearly defined set of correct answer(s), ensuring that no alternative diagnosis outside this set can be considered equally plausible given the presented case, and (2) each question must include at least one distractor that models are likely to confuse with the correct answer, while each distractor still has a clear clinical basis for being incorrect. To this end, we construct distractors based on \kg{} using symptom overlap as the primary criterion, and apply type-specific selection strategies across case types.

To facilitate more effective option construction, we categorize candidate disorders into three levels based on their relevance to the target disorder. This relevance is quantified using the hierarchical structure of \kg{}, which encodes relationships among disorders, symptom groups, and individual symptoms.

\begin{itemize}
    \item \textbf{High}: Disorders that share at least one symptom group with the target disorder's symptom profile (e.g., Major Depressive Disorder (MDD) and Bipolar I/II disorders, which share depressive episodes).
    
    \item \textbf{Moderate}: Disorders that do not share symptom groups but share at least one individual symptom with the target disorder (e.g., Post-Traumatic Stress Disorder (PTSD) and Generalized Anxiety Disorder (GAD), which may share symptoms such as anxiety).
    
    \item \textbf{Low}: Disorders that do not share any symptoms with the target disorder (e.g., Schizophrenia and Adjustment Disorder).
\end{itemize}

\paragraph{Options for Types 1 and 2.}
Types 1 and 2 are designed to require the model to select the single most appropriate diagnosis given a clinical scenario describing a specific disorder. For each question, the three distractors are selected as the top-ranked disorders based on the relevance criteria defined above, excluding the ground-truth diagnosis. As a result, all options consist of disorders that are symptomatically similar to the target disorder, requiring the model to perform fine-grained diagnostic reasoning to identify the correct answer. 

However, when this strategy is applied uniformly across all disorders, we observe that for certain disorder pairs---particularly those distinguished by subtle criteria such as episode duration or temporal thresholds---some distractors may also be considered plausible answers for a given clinical case. To address this issue, we introduce exception rules that explicitly exclude specific distractors for selected disorders. These exceptions are determined in consultation with a psychiatrist to ensure that each question maintains a clearly defined correct answer. The list of such exceptions is provided below.

\begin{itemize}
    \item \textbf{D007} (Schizoaffective Disorder - Depressive Type)
    \begin{itemize}
        \item Exclude: \textbf{D006} (Schizoaffective Disorder - Bipolar Type)
    \end{itemize}

    \item \textbf{D009} (Bipolar II Disorder)
    \begin{itemize}
        \item Exclude: \textbf{D008} (Bipolar I Disorder)
    \end{itemize}

    \item \textbf{D011} (Generalized Anxiety Disorder)
    \begin{itemize}
        \item Exclude: \textbf{D006} (Schizoaffective Disorder - Bipolar Type), 
        \textbf{D007} (Schizoaffective Disorder - Depressive Type), 
        \textbf{D008} (Bipolar I Disorder), 
        \textbf{D009} (Bipolar II Disorder), 
        \textbf{D010} (Bipolar I Disorder - Psychotic Features)
    \end{itemize}

    \item \textbf{D018} (Posttraumatic Stress Disorder)
    \begin{itemize}
        \item Exclude: \textbf{D019} (Acute Stress Disorder)
    \end{itemize}

    \item \textbf{D019} (Acute Stress Disorder)
    \begin{itemize}
        \item Exclude: \textbf{D018} (Posttraumatic Stress Disorder)
    \end{itemize}
\end{itemize}

\paragraph{Options for Types 3 and 4.}
Types 3 and 4 are designed to evaluate differential diagnosis between two clinically similar disorders encoded in MentalKG. Accordingly, each question is constructed around a paired differential-diagnosis structure consisting of two candidate disorders, denoted as A and B, both of which are always included in the option set.

For Type 3, the scenario presents shared clinical evidence that keeps both A and B consistent with the available information, while omitting or underspecifying the discriminating evidence required to rule out either candidate. Thus, both A and B are treated as correct answers. For Type 4, the scenario includes the relevant discriminating evidence, making only one of the paired disorders correct while the other serves as a clinically motivated distractor.

The remaining two options are selected based on the relevance criteria defined above, with one drawn from the Moderate category and one from the Low category. This design encourages models to focus on resolving the differential-diagnosis pair while reducing the chance that additional distractors become inadvertently plausible. Unlike Types 1 and 2, which use single-disorder distractor filtering, Types 3 and 4 rely on the paired differential-diagnosis structure to determine the primary competing diagnosis and answer set.

\subsection{\label{appendix: expert validation}Expert Validation}
To ensure the clinical reliability and quality of the generated scenarios, we conduct a rigorous human evaluation involving domain experts. One psychiatrist and one clinical psychologist independently reviewed a sample of 220 scenarios generated by our framework. The evaluation set is constructed to provide balanced coverage across disorders: for Types 1 and 2, we include one scenario per disorder (23 each), while for Types 3 and 4, we include one scenario per differential diagnosis pair (87 each). For Type 4, we sample one instance per differential diagnosis pair with a randomly selected target label, resulting in a total of 220 evaluated cases. 

\paragraph{Evaluation Guidelines.}
\input{Tables/Appendix/Experts/expert_guideline}

Experts were asked to rate each generated clinical scenario on a 5-point Likert scale across the following three dimensions. The detailed scoring rubric provided to the experts is shown in Table~\ref{tab:expert_guidelines}.

\begin{itemize} 
    \item \textbf{Linguistic Naturalness}: Whether the symptom expressions in the generated scenario are natural and plausible in real clinical contexts. For example, in patient self-reports, symptoms should be expressed in everyday language, such as ``I do not even enjoy the games I used to play,'' rather than as a direct clinical label such as ``loss of interest.'' 
    
    \item \textbf{Diagnostic Validity}: Whether the generated scenario sufficiently supports the gold diagnosis and appropriately rules out incorrect options. For example, if Major Depressive Disorder is provided as the correct answer, the case should be more consistent with MDD than with other options such as Persistent Depressive Disorder, Bipolar I Disorder, or Schizoaffective Disorder (Depressive Type). 
    
    \item \textbf{Clinical Realism}: Whether the presented symptom combination and manifestation are realistic in clinical practice. For example, insomnia, appetite loss, and concentration difficulties are commonly co-occurring symptoms in patients with Major Depressive Disorder.
\end{itemize}

\paragraph{Evaluation Result.}

\autoref{tab:expert_valid} summarizes the results. The generated scenarios receive consistently high ratings from both experts, with mean scores exceeding 4.3 across all dimensions. Notably, \textit{Linguistic Naturalness} achieves near-perfect scores (4.94 and 4.91), and \textit{Clinical Realism} is also rated highly (4.83 and 4.87), demonstrating that our LLM-based generation pipeline successfully captures the subtleties of patient narratives.

In contrast, \textit{Diagnostic Validity} receives comparatively lower scores (4.70 and 4.36) than the other two dimensions, while still maintaining a high level of overall clinical validity. This trend is particularly pronounced in Type 3 scenarios, where multiple clinically plausible diagnoses are intentionally incorporated. To better understand this pattern, we further analyze the distribution of expert scores and conduct a case-level analysis of scenarios with lower ratings. 

\paragraph{Score Distribution.}
\input{Figures/Appendix/expert_score_distribution}
Figure~\ref{fig:score_distribution} presents the distribution of Diagnostic Validity scores across different case types. Types 1, 2, and 4 exhibit highly concentrated distributions at the upper end of the scale, with most cases receiving scores of 4 or 5 from both experts. In particular, Expert 2 assigns a score of 5 to nearly all cases, while Expert 1 also rates the majority of cases as 4 or higher, indicating that these scenarios are generally diagnostically well-grounded.

In contrast, Type 3 shows a more dispersed distribution, with scores spanning 3 to 5. Nevertheless, most cases still receive scores of 3 or higher. According to our evaluation rubric, a score of 3 indicates that the scenario remains aligned with the target diagnosis, although it may also be consistent with an alternative disorder. As a result, the question–answer pair remains valid, while still introducing controlled ambiguity that challenges fine-grained diagnostic reasoning.

\subsection{\label{appen:example}Representative Examples}

\input{Tables/Appendix/mentalbench_examples/type12}
\input{Tables/Appendix/mentalbench_examples/type3}
\input{Tables/Appendix/mentalbench_examples/type4}

To provide a qualitative understanding of the benchmark structure, we present representative examples from all four evaluation types in \ours{}. Table~\ref{tab:case_1_2_examples} illustrates the contrast between structured clinical summaries (Type 1) and subjective patient narratives (Type 2) under different levels of information completeness. Specifically, Type 1 presents clinically organized summaries with explicit diagnostic evidence, whereas Type 2 consists of incomplete and colloquial symptom descriptions that mimic real-world patient self-reports.

Tables~\ref{tab:case_3_examples} and~\ref{tab:case_4_examples} present representative examples of differential diagnosis scenarios with varying levels of diagnostic ambiguity. Type 3 introduces additional conditions associated with differential diagnosis, resulting in clinically ambiguous scenarios where multiple candidate disorders remain diagnostically plausible. In contrast, Type 4 incorporates explicit discriminating rules that resolve diagnostic ambiguity, leading to a single definitive diagnosis despite overlapping symptom profiles.

%% file: Figures/Appendix/Language/tsne_figs.tex
% \clearpage
\begin{figure*}[ht!]
    \centering
    \includegraphics[width=0.9\textwidth]{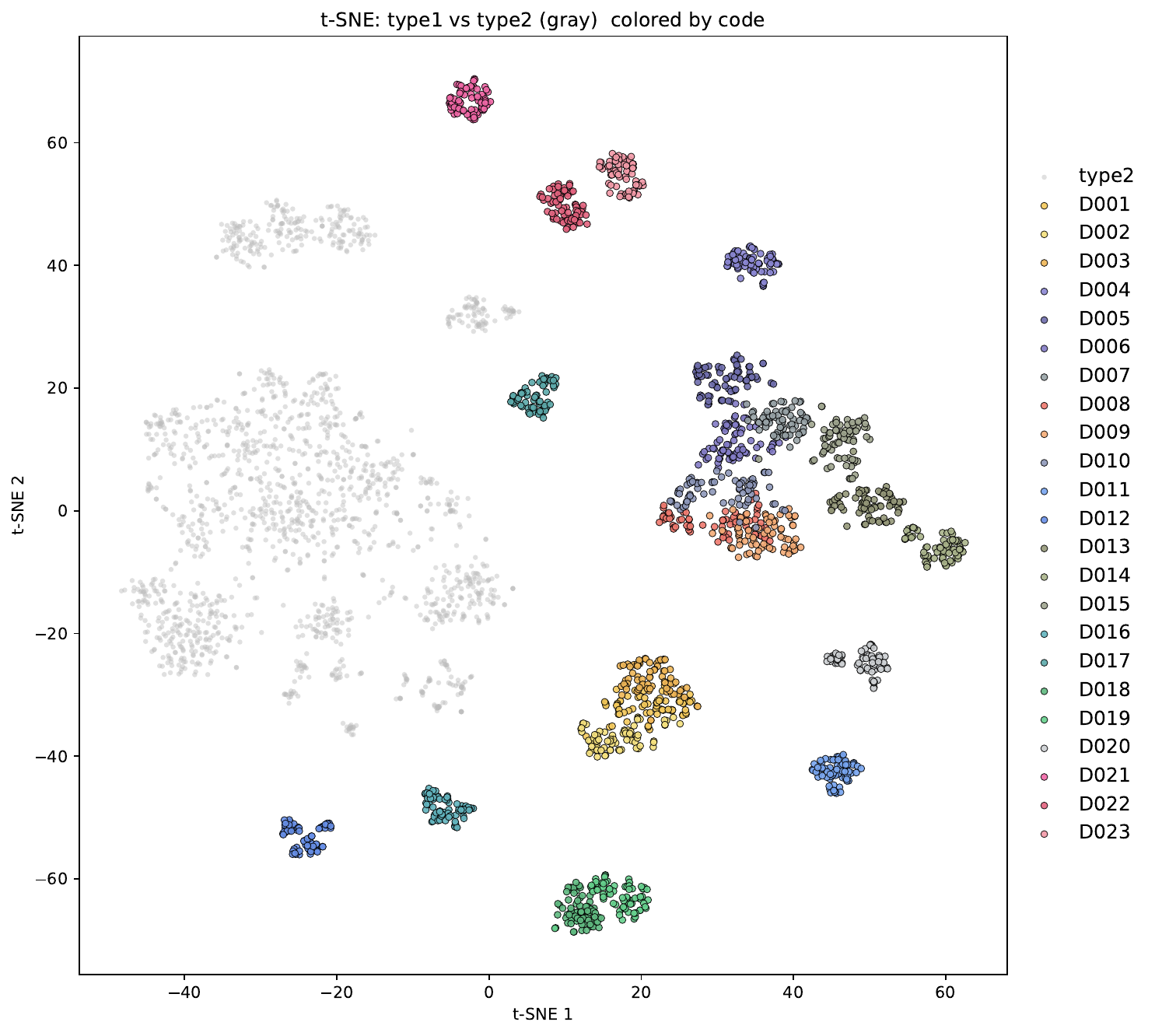}
    % \\[1em]
    \includegraphics[width=0.9\textwidth]{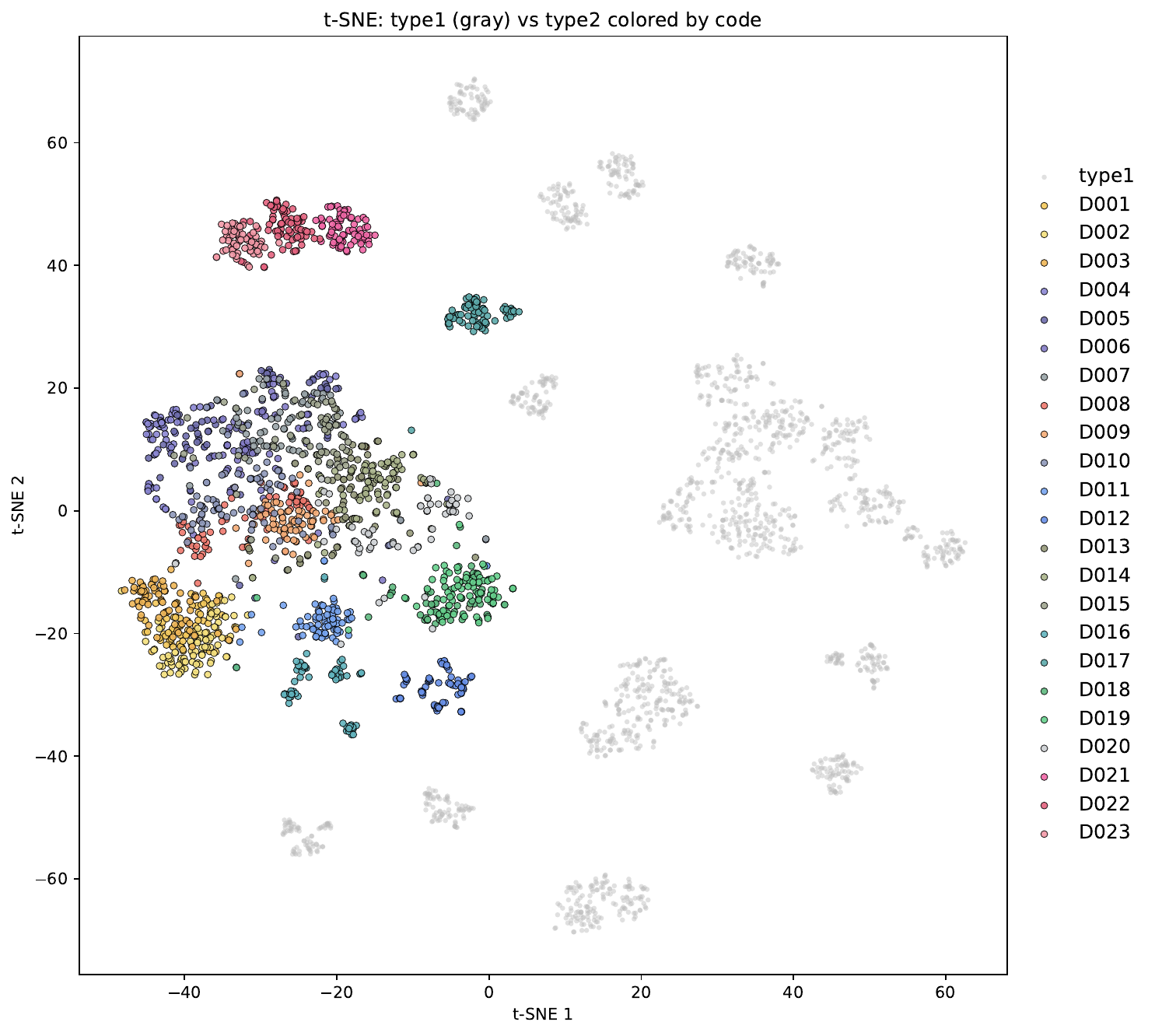}
    \caption{t-SNE visualization of sentence embeddings across 23 disorder categories, illustrating the linguistic divergence between Type 1 (upper) and Type 2 (lower) samples.}
    \label{fig:tsne_by_code}
\end{figure*}

%% file: Figures/Appendix/Language/tsne_type3vs4.tex
\begin{figure*}[ht!]
    \centering
    \includegraphics[width=0.9\textwidth]{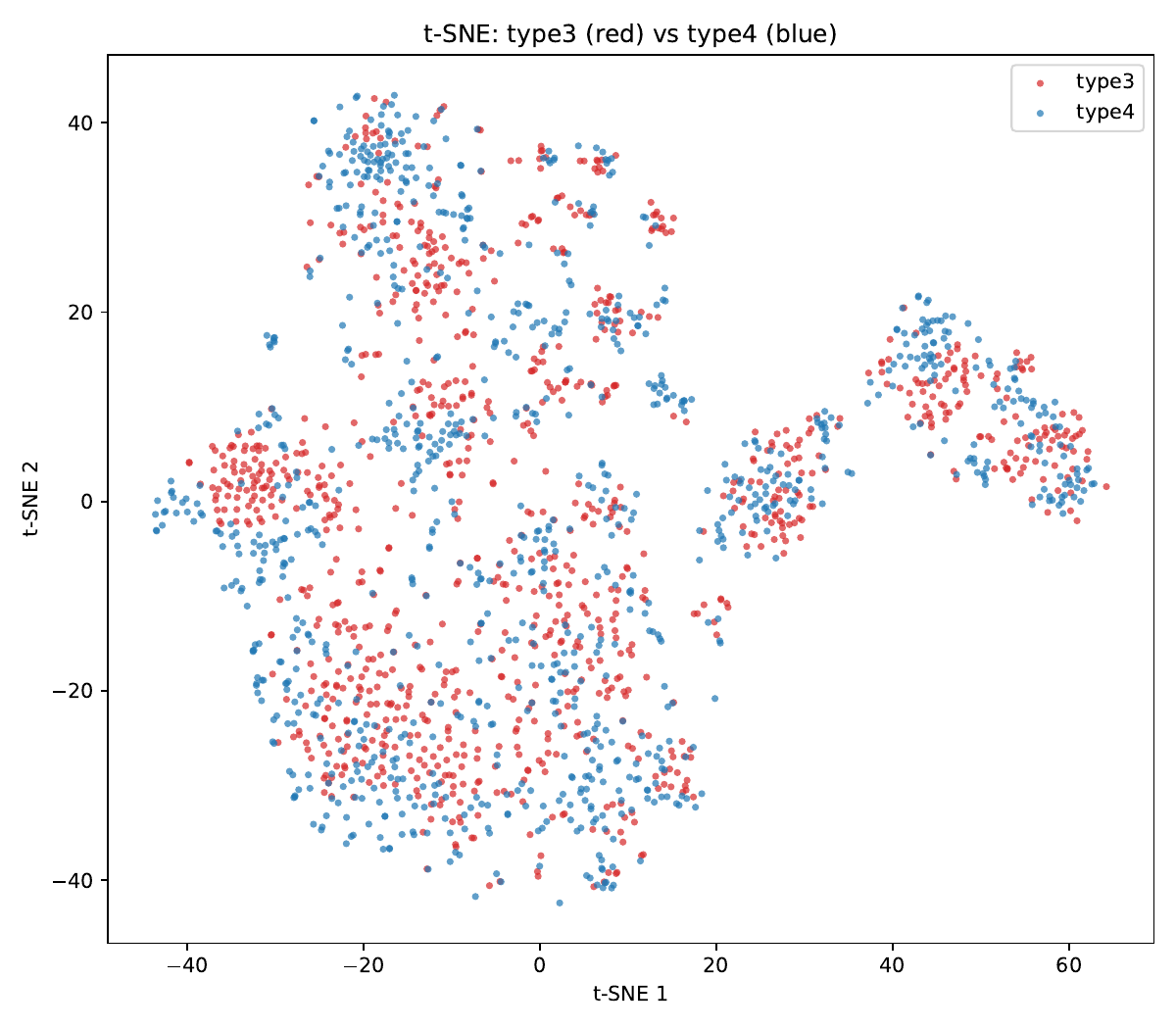}
    \caption{\label{fig:tsne_type34}t-SNE visualization of sentence embeddings for Type 3 (red) and Type 4 (blue) samples, demonstrating substantial semantic convergence.}
\end{figure*}

%% file: Tables/Appendix/seed_examples/seed_type1.tex
\definecolor{bgcolor}{RGB}{248,248,246}
\definecolor{keycolor}{RGB}{120,40,40}

\lstdefinelanguage{json}{
    morestring=[b]",
}

\lstdefinestyle{diagnosticjson}{
    language=json,
    basicstyle=\ttfamily\footnotesize\color{black},
    backgroundcolor=\color{bgcolor},
    frame=single,
    framerule=0.3pt,
    rulecolor=\color{black!20},
    breaklines=true,
    showstringspaces=false,
    keepspaces=true,
    columns=fullflexible,
    xleftmargin=6pt,
    xrightmargin=6pt,
    stringstyle=\color{black},
    literate=
     *{question}{{{\color{keycolor}\bfseries question}}}{8}
      {option}{{{\color{keycolor}\bfseries option}}}{6}
}

\begin{table}[ht!]
\centering

\begin{lstlisting}[
style=diagnosticjson,
]
{
  "question": {
    "disease_code": "D013",
    "disease_name": "Major Depressive Disorder",

    "sampled_features": {
      "sampled_symptom_groups": {
        "depressive_episode": {
          "functional_impairment_required": true,

          "symptoms": [
            "S024", "Depressed_Mood",
            "S025", "Loss_of_Interest",
            "S026", "Weight_Appetite_Change",
            "S027", "Sleep_Disturbance",
            "S028", "Psychomotor_Changes",
            "S030", "Worthlessness_Guilt",
            "S031", "Diminished_Thinking",
            "S032", "Death_Suicide"
          ],

          "duration": "1 years and 2 months",
          "name": "depressive_episode"
        }
      }
    }
  },

  "option": {
    "options": [
      "Schizoaffective Disorder (Bipolar Type)",
      "Schizoaffective Disorder (Depressive Type)",
      "Bipolar I Disorder",
      "Major Depressive Disorder"
    ],

    "answer_index": [3]
  }
}
\end{lstlisting}

\caption{Example of Type 1 seed for Major Depressive Disorder}
\label{tab:type1_seed_example}

\end{table}

%% file: Tables/Appendix/seed_examples/seed_type2.tex
\definecolor{bgcolor}{RGB}{248,248,246}
\definecolor{keycolor}{RGB}{120,40,40}

\lstdefinelanguage{json}{
    morestring=[b]",
}

\lstdefinestyle{diagnosticjson}{
    language=json,
    basicstyle=\ttfamily\footnotesize\color{black},
    backgroundcolor=\color{bgcolor},
    frame=single,
    framerule=0.3pt,
    rulecolor=\color{black!20},
    breaklines=true,
    showstringspaces=false,
    keepspaces=true,
    columns=fullflexible,
    xleftmargin=6pt,
    xrightmargin=6pt,
    stringstyle=\color{black},
    literate=
     *{question}{{{\color{keycolor}\bfseries question}}}{8}
      {option}{{{\color{keycolor}\bfseries option}}}{6}
}

\begin{table}[ht!]
\centering

\begin{lstlisting}[
style=diagnosticjson,
]
{
  "question": {
    "disease_code": "D013",
    "disease_name": "Major Depressive Disorder",

    "sampled_symptom_groups": {
      "depressive_episode": {
        "functional_impairment_required": true,

        "symptoms": [
          "S024", "Depressed_Mood",
          "S028", "Psychomotor_Changes",
          "S030", "Worthlessness_Guilt",
          "S032", "Death_Suicide"
        ],

        "duration": "5 months",
        "name": "depressive_episode"
      }
    }
  },

  "option": {
    "options": [
      "Schizoaffective Disorder (Depressive Type)",
      "Major Depressive Disorder",
      "Bipolar I Disorder",
      "Schizoaffective Disorder (Bipolar Type)"
    ],

    "answer_index": [1]
  }
}
\end{lstlisting}

\caption{Example of Type 2 seed for Major Depressive Disorder}
\label{tab:type2_seed_example}

\end{table}

%% file: Tables/Appendix/seed_examples/seed_type3.tex
\definecolor{bgcolor}{RGB}{248,248,246}
\definecolor{keycolor}{RGB}{120,40,40}

\lstdefinelanguage{json}{
    morestring=[b]",
}

\lstdefinestyle{diagnosticjson}{
    language=json,
    basicstyle=\ttfamily\footnotesize\color{black},
    backgroundcolor=\color{bgcolor},
    frame=single,
    framerule=0.3pt,
    rulecolor=\color{black!20},
    breaklines=true,
    showstringspaces=false,
    keepspaces=true,
    columns=fullflexible,
    xleftmargin=6pt,
    xrightmargin=6pt,
    stringstyle=\color{black},
    literate=
     *{question}{{{\color{keycolor}\bfseries question}}}{8}
      {option}{{{\color{keycolor}\bfseries option}}}{6}
}

\begin{table}[ht!]
\centering

\begin{lstlisting}[
style=diagnosticjson,
]
{
  "question": {
    "disease_code": "D013",
    "disease_name": "Major Depressive Disorder",

    "sampled_symptom_groups": {
      "depressive_episode": {
        "functional_impairment_required": true,

        "symptoms": [
          "S024", "Depressed_Mood",
          "S028", "Psychomotor_Changes",
          "S030", "Worthlessness_Guilt",
          "S032", "Death_Suicide"
        ],

        "duration": "5 months",
        "name": "depressive_episode"
      }
    },

    "differential_diagnosis": {
      "additional_condition": "Presence of Manic Symptoms"
    }
  },

  "option_both": {
    "options": [
      "Persistent Depressive Disorder",
      "Major Depressive Disorder",
      "Bipolar I Disorder",
      "Body Dysmorphic Disorder"
    ],

    "answer_index": [1, 2]
  }
}
\end{lstlisting}

\caption{Example of Type 3 seed with differential diagnosis between Major Depressive Disorder and Bipolar I Disorder}
\label{tab:type3_seed_example}

\end{table}

%% file: Tables/Appendix/seed_examples/seed_type4.tex
\definecolor{bgcolor}{RGB}{248,248,246}
\definecolor{keycolor}{RGB}{120,40,40}

\lstdefinelanguage{json}{
    morestring=[b]",
}

\lstdefinestyle{diagnosticjson}{
    language=json,
    basicstyle=\ttfamily\footnotesize\color{black},
    backgroundcolor=\color{bgcolor},
    frame=single,
    framerule=0.3pt,
    rulecolor=\color{black!20},
    breaklines=true,
    showstringspaces=false,
    keepspaces=true,
    columns=fullflexible,
    xleftmargin=6pt,
    xrightmargin=6pt,
    stringstyle=\color{black},
    literate=
     *{question}{{{\color{keycolor}\bfseries question}}}{8}
      {option}{{{\color{keycolor}\bfseries option}}}{6}
}

\begin{table}[ht!]
\centering

\begin{lstlisting}[
style=diagnosticjson,
]
{
  "question": {
    "disease_code": "D013",
    "disease_name": "Major Depressive Disorder",

    "sampled_symptom_groups": {
      "depressive_episode": {
        "functional_impairment_required": true,

        "symptoms": [
          "S024", "Depressed_Mood",
          "S028", "Psychomotor_Changes",
          "S030", "Worthlessness_Guilt",
          "S032", "Death_Suicide"
        ],

        "duration": "5 months",
        "name": "depressive_episode"
      }
    },

    "differential_diagnosis": {
      "additional_condition": "Presence of Manic Symptoms",
      "key_difference": "Meeting criteria for manic episode and history",
      "rule": [
        "Manic-like symptoms may be present, but criteria for a manic episode are not met, and there is no history of manic or hypomanic episodes."
      ]
    }
  },

  "option_a": {
    "options": [
      "Persistent Depressive Disorder",
      "Major Depressive Disorder",
      "Bipolar I Disorder",
      "Body Dysmorphic Disorder"
    ],

    "answer_index": [1]
  }
}
\end{lstlisting}

\caption{Example of Type 4 seed with differential diagnosis between Major Depressive Disorder and Bipolar I Disorder}
\label{tab:type4_seed_example}

\end{table}

%% file: Tables/Dataset/data_stat.tex
\begin{table}[ht]
\centering
\scriptsize
\caption{\label{tab:data_stats}Statistics of the generated data samples}
\begin{tabular}{lrrrrrr}
\hline
\textbf{Type} & \textbf{Disorder} & \textbf{Seed} & \textbf{Profile} & \textbf{Generator} & \textbf{Case} \\ \hline
Type 1 & 23    & 5  & 5 & 3 & 1,725  \\
Type 2 & 23    & 10 & 5 & 3 & 3,450  \\
Type 3 & 87    & 5  & 5 & 3 & 6,525  \\
Type 4 & 87*2  & 5  & 5 & 3 & 13,050 \\ \hline
\textbf{Total} & -- & -- & -- & -- & \textbf{24,750} \\ \hline
\end{tabular}
\end{table}

%% file: Tables/Appendix/Experts/expert_guideline.tex
\begin{table*}[t]
\centering
\small
\begin{tabular}{p{3cm} p{0.8cm} p{8.8cm}}
\toprule
\textbf{Dimension} & \textbf{Score} & \textbf{Description} \\
\midrule

\multirow{5}{*}{\textbf{Linguistic Naturalness}}
& 1 & Expressions are unnatural or impossible in real clinical settings. \\
& 2 & Mechanical or textbook-like expressions with unclear meaning. \\
& 3 & Clearly artificial, but the meaning is understandable. \\
& 4 & Minor awkwardness, but generally natural. \\
& 5 & Fully natural and indistinguishable from real patient reports. \\
\midrule

\multirow{5}{*}{\textbf{Diagnostic Validity}}
& 1 & Does not meet diagnostic criteria; inconsistent with the target diagnosis. \\
& 2 & Missing core symptoms or contains conflicting evidence. \\
& 3 & Aligned with the target diagnosis but closer to another disorder. \\
& 4 & Mostly aligned, but could still be interpreted as another disorder. \\
& 5 & Fully supports the target diagnosis without ambiguity. \\
\midrule

\multirow{5}{*}{\textbf{Clinical Realism}}
& 1 & Clinically implausible (e.g., impossible symptom combinations). \\
& 2 & Unrealistic progression or symptom patterns. \\
& 3 & Some unrealistic or overly textbook-like aspects. \\
& 4 & Rare but plausible in real clinical settings. \\
& 5 & Highly realistic and representative of real-world cases. \\
\bottomrule
\end{tabular}
\caption{Detailed scoring rubric used for expert evaluation.}
\label{tab:expert_guidelines}
\end{table*}

%% file: Figures/Appendix/expert_score_distribution.tex
\begin{figure*}[h]
    \centering
    \includegraphics[width=1.0\textwidth]{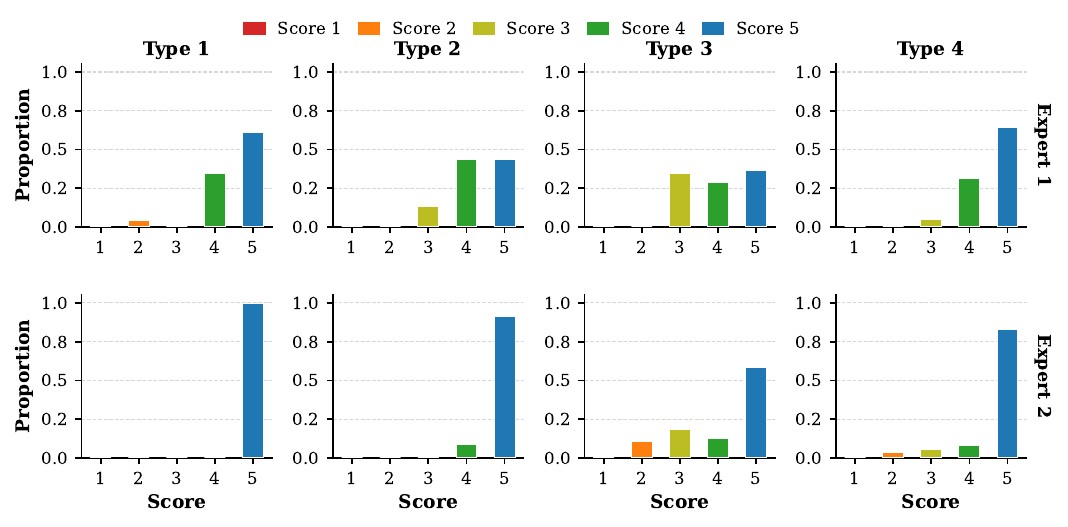}
    \caption{\label{fig:score_distribution} Distribution of Diagnostic Validity scores across case types for both experts.}
\end{figure*}

%% file: Tables/Appendix/mentalbench_examples/type12.tex
\begin{table*}[t]
    \centering
    \small
    \renewcommand{\arraystretch}{1.2}
    \caption{Representative examples of Type 1 and Type 2 cases in \ours{}.}
    \label{tab:case_1_2_examples}

    % \begin{tabular}{p{2.0cm} p{1.0cm} p{1.0cm} p{8.5cm}}
    \begin{tabularx}{\textwidth}{
        p{2.0cm}
        p{1.0cm}
        p{1.0cm}
        >{\raggedright\arraybackslash}X
    }
        \toprule
        \makecell[l]{\textbf{Target} \textbf{disease}}
        & \textbf{Model}
        & \textbf{Type}
        & \textbf{Generated case} \\
        \midrule

        \multirow{2}{*}{\makecell[l]{D005\\(Schizophrenia)}}
        & \multirow{2}{*}{Gemini}
        & Type I
        & A 29-year-old single male sound engineer presents with a 14-month history of significant functional decline. For the past four months, he has exhibited prominent psychotic phenomena, including grandiose delusions characterized by beliefs of possessing extraordinary abilities. He also experiences auditory hallucinations, specifically hearing voices in the absence of external stimuli. Motoric disturbances are noted, with episodes of catalepsy where he maintains an induced posture against gravity. Additionally, he demonstrates avolition, manifesting as a marked decrease in engagement in purposeful, self-initiated activities. These symptoms have led to considerable impairment in his occupational and social functioning over the past 14 months.\\
        \cline{3-4}
        &
        & \rule{0pt}{2.6ex} Type II
        & It's been a rough few months, doc. Like, two months now, things have just felt... off. I can't shake this feeling that my old firm, maybe even my ex-wife, is out to get me. I see their cars, or cars that look just like theirs, everywhere I go. I even hear things sometimes, like whispers about me. And the taste... that's the weirdest part. This metallic, bitter taste, like old pennies, is just constantly in my mouth, no matter what I eat or drink. It's making it impossible to focus on anything, especially my consulting work. I'm usually so sharp, but now, it’s like my brain is just… fuzzy. I'm afraid to leave the house some days, afraid of what 'they' might do. \\

        \midrule

        \multirow{2}{*}{\makecell[l]{D011\\(Generalized \\Anxiety Disorder)}}
        & \multirow{2}{*}{GPT-5}
        & Type I
        & A 32-year-old single female simultaneous interpreter presents with a 7-month history of pervasive anxiety and associated somatic symptoms. She reports excessive, wide-ranging worry occurring on most days, experienced as difficult to regulate despite deliberate efforts to distract or reassure herself. The patient describes persistent internal tension and nervousness, with heightened startle response to minor stimuli. She endorses ongoing fatigue that is disproportionate to activity level and not relieved by usual periods of rest or sleep, contributing to reduced stamina and decreased capacity to sustain her usual work pace. She notes frequent jaw clenching and tension, particularly during periods of heightened worry, associated with jaw discomfort and intermittent headaches. The symptoms have been present for at least 7 months, with no discrete precipitating event identified, and occur independently of specific situational triggers. They have led to meaningful functional impairment, including reduced efficiency and concentration at work and withdrawal from some social and leisure activities due to exhaustion and preoccupation with worries. There is no indication that occupational status or marital status temporally coincides with symptom onset. \\

        \cline{3-4}
        &
    
        & \rule{0pt}{2.6ex}Type II
        & I work nights as a security guard at a warehouse, so I’m used to being alert, but this is different. For about eight months now, I feel like my mind never shuts off. I’m constantly going over what could go wrong—at work, with my wife’s health, the bills, my kids—even when nothing in particular is happening. It’s there most days, all day. Even when I’m doing my rounds or watching the monitors, these thoughts just push their way in. I’ll suddenly catch myself staring at the screen, not really seeing it because I’m thinking about, say, the car breaking down or losing my job. It makes it hard to focus. Physically, I feel like I’m wound tight, like a spring that won’t unwind. I can’t relax, even on my days off. I pace around the house, can’t sit through a full TV show. My muscles are always sore—my back, shoulders, legs—like I’ve been tensing them without realizing it. I get tired very easily, but when I try to rest, my body just won’t let me fully relax. \\

        \bottomrule
    \end{tabularx}
\end{table*}

%% file: Tables/Appendix/mentalbench_examples/type3.tex
\begin{table*}[t]
    \centering
    \small
    \renewcommand{\arraystretch}{1.2}
    \caption{Representative examples of Type 3 cases in \ours{}. }
    \label{tab:case_3_examples}

    % \begin{tabular}{p{2.0cm} >{\centering\arraybackslash}p{1.6cm} p{0.8cm} p{8.2cm}}
    \begin{tabularx}{\textwidth}{
            p{2.0cm}
            >{\centering\arraybackslash}p{1.6cm}
            p{1.0cm}
            >{\raggedright\arraybackslash}X
        }
        \toprule
        \makecell[l]{\textbf{Target} \textbf{disease}}
        & \makecell[l]{\textbf{Differential}\\\textbf{diagnosis}}
        & \textbf{Model}
        & \textbf{Generated case} \\
        \midrule

        \multirow{2}{*}{\makecell[l]{D005\\(Schizophrenia)}}
        & \multirow{2}{*}{\makecell[l]{D015\\(Major \\Depressive\\Disorder\\with\\Psychotic\\Features)}}
        & \multirow{2}{*}{Gemini}
        & \textbf{\textit{Additional Condition.}}
      Presence of depressive symptoms.\\
        % \cline{4-4}
        &
        &
        & \textbf{\textit{Case.}}
        It's been... rough, these past two months. I feel like my senses are playing tricks on me. Sometimes I smell this really strong, foul odor, like something's burning or rotting, but no one else smells it. It’s disorienting, especially at work.And on top of that, I’ve had this growing sense that there’s something special about me, a purpose I'm meant to fulfill. Like I have this incredible power to heal that goes beyond medicine, or that I’m destined for something truly significant, more than just being a nurse. It's exhilarating and terrifying at the same time. I’m so tired, all the time. Getting out of bed feels like a monumental effort, and I just don’t find joy in anything anymore. My appetite is gone, and I’ve lost weight without trying. It's hard to focus, both on my patients and just in general. I try to put on a brave face, but inside, I just feel this immense hopelessness swirling around. I don't know what's happening to me. \\

        \midrule

        \multirow{2}{*}{\makecell[l]{D005\\(Schizophrenia)}}
        & \multirow{2}{*}{\makecell[l]{D018\\(Post\\Traumatic\\Stress\\Disorder)}}
        & \multirow{2}{*}{Gemini}
        & \textbf{\textit{Additional Condition.}}
      Experience of Traumatic Event.\\
        % \cline{4-4}
        &
        &
        & \textbf{\textit{Case.}}
        I'm glad to finally be here. Things have been… off for about two months now. It started subtly, but it’s gotten to the point where I just can't shake it. There’s this woman, you see, a colleague, she’s been sending me signs. Little glances, a certain way she smiles. I know she cares for me, more than just a work friend. It’s clear as day. Sometimes, I also feel this crawly sensation, like bugs or something under my skin. It’s not always there, but it’s enough to make me uncomfortable. My wife, she says I’m imagining things, but it feels so real. Honestly, I'm just looking for some answers, some relief. \\

        \midrule
        
        \multirow{2}{*}{\makecell[l]{D011\\(Generalized \\Anxiety Disorder)}}
        & \multirow{2}{*}{\makecell[l]{D016\\(Obsessive\\Compulsive\\Disorder)}}
        & \multirow{2}{*}{GPT-5}
        & \textbf{\textit{Additional Condition.}}
      None. \\
        % \cline{4-4}
        &
        & 
        & \textbf{\textit{Case.}}
        I’ve been feeling on edge for a few months now, maybe three or four. It’s like my brain never really shuts up. I’ll start off thinking about the dinner service—whether the deliveries will be on time—and then it jumps to whether we’re making enough money, then to my wife’s health, then to what would happen if one of us couldn’t work. It just snowballs from one thing to another, even if nothing specific has actually gone wrong. It’s not like there’s one particular thing I’m stuck on; it just hops around: the restaurant, our bills, my parents getting older, whether I messed something up on last night’s menu. By the time I’m home, my body feels like it’s been clenched all day. My shoulders and neck are rock hard, my back aches, even my forearms and legs feel sore, like I’ve overworked them, even on slower days. When I lie down at night, I notice how stiff I am. I try to stretch or breathe it out, but I can’t fully relax, and the tension and the random “what if” thoughts make it hard to just switch off. \\

      %   \midrule
        
      %   \multirow{2}{*}{\makecell[l]{D011\\(Generalized \\ Anxiety Disorder)}}        
      %   & \multirow{2}{*}{\makecell[l]{D020\\(Adjustment\\Disorder)}}
      %   & \multirow{2}{*}{GPT-5}
      %   & \textbf{\textit{Additional Condition.}}
      % Experience of Psychosocial Stress.\\
      %   % \cline{4-4}
      %   &
      %   &
      %   & \textbf{\textit{Case.}}
      %   It’s hard to explain, but for a few months now I feel like my nerves are always buzzing. I jump at every little noise in the kitchen – if someone drops a pan behind me, my heart just shoots up into my throat. Even at home, a car door slamming outside makes me tense up. While I’m at the sink, I’ll suddenly start thinking, “If I break this plate, they’ll get rid of me… then I can’t pay the rent… I’ll lose the flat,” and it just spirals from there, even though nothing’s actually happened. Those thoughts just cut in while I’m trying to count plates or concentrate on not burning myself, and then I lose track of what I’m doing. My shoulders and the back of my neck feel like tight knots most days. Sometimes the tension crawls up into my head and I get this dull ache that doesn’t really go away, even when I’m off work. Things have been a bit difficult lately, with changes at work and at home, and I feel like I’m not handling it as well as I used to, but I’m not sure if that’s all this is or something else. \\

        \bottomrule
    \end{tabularx}
\end{table*}

%% file: Tables/Appendix/mentalbench_examples/type4.tex
\begin{table*}[t]
    \centering
    \small
    \renewcommand{\arraystretch}{1.4}
    \caption{Representative examples of Type 4 cases in \ours{}.}
    \label{tab:case_4_examples}

    % \begin{tabular}{p{2.0cm} >{\centering\arraybackslash}p{1.6cm} p{1.0cm} p{8.2cm}}
    \begin{tabularx}{\textwidth}{
        p{2.0cm}
        >{\centering\arraybackslash}p{1.6cm}
        p{1.0cm}
        >{\raggedright\arraybackslash}X
    } 
        \toprule
        \makecell[l]{\textbf{Target} \textbf{disease}}
        & \makecell[c]{\textbf{Differential}\\\textbf{diagnosis}}
        & \textbf{Model}

        & \textbf{Generated case} \\
        \midrule

        \multirow{4}{*}{\makecell[l]{D005\\(Schizophrenia)}}
        & \multirow{4}{*}{\makecell[l]{D015\\(Major \\Depressive\\Disorder\\with\\Psychotic\\Features)}}
        & \multirow{4}{*}{Gemini}

        & \textbf{\textit{Diagnosis.}} D005 (Schizophrenia) \\
        &
        &
        & \textbf{\textit{Discriminative Rule.}}
        Depressive symptoms may be accompanied but appear for a relatively short period compared to the total course of the illness, and psychotic symptoms appear continuously regardless of depressive symptoms.\\
        % \cline{4-4}
        &
        & 
        & \textbf{\textit{Scenario.}}
        It all started about two months ago. I've been feeling so low, just completely drained and unable to find enjoyment in anything, even the things I used to love about nursing. It's been hard to even get out of bed some mornings, and I’ve really retreated from my friends. My appetite's gone, and I’m exhausted all the time, but sleep doesn’t help. The worst part is this conviction I have that I’m being specifically chosen by a higher power to heal the world. I know it sounds strange, especially since I'm just a nurse, but it feels so real, like I have this extraordinary gift, this crucial mission. And recently, there's this smell, too – a strong, sweet, decaying scent that no one else seems to notice. I caught a whiff of it during rounds yesterday, and it’s just constantly there, following me. This feeling of being profoundly sad and hopeless comes and goes, but the special calling and the smell, they're always there, every single day, no matter how I'm feeling otherwise. \\
        \cline{4-4}
        &
        &
        & \textbf{\textit{Diagnosis.}} D015 (Major Depressive Disorder with Psychotic Features)\\
        &
        &
        & \textbf{\textit{Discriminative Rule.}}
        Depressive symptoms are accompanied, and psychotic symptoms appear only during the period of the major depressive episode.\\
        &
        &
        & \textbf{\textit{Scenario.}}
        It's been really tough lately, doctor. For the past two months, I’ve felt like I’m moving through quicksand. I barely sleep, and when I do, it's not restful. My appetite is gone, and I’ve lost so much weight. I just feel utterly empty, drained of all energy and completely worthless. Every little mistake at work, even a tiny miscalculation with medication, feels like the end of the world, like I’m a complete fraud who shouldn’t be a nurse at all. But it’s more than just the sadness. Lately, I've had these really strange experiences. I’ll constantly smell this faint, rotting garbage odor, even though no one else seems to notice it and I've checked everywhere. It's so distracting, especially when I'm trying to focus at work. And I know it sounds crazy, but I truly believe I have a special gift, a sort of advanced intuition that allows me to predict patient outcomes just by looking at them. I even feel like I can influence their recovery with my thoughts. I know it isn't logical, but these thoughts and the smells only happen when I'm feeling this low, this utterly despondent. When I have a rare good day, they completely disappear.\\
        
        % \midrule

        % \multirow{2}{*}{\makecell[l]{D011\\(Generalized\\Anxiety Disorder)}}
        % & \multirow{2}{*}{\makecell[l]{D016\\(Obsessive\\Compulsive\\Disorder)}}
        % & \multirow{2}{*}{GPT}

        % & \textbf{\textit{Diagnosis.}} Generalized Anxiety Disorder \\
        % &
        % &
        % & \textbf{\textit{Discriminative Rule.}}
        % The focus of worry is on forthcoming problems, and the degree is excessive.\\
        % % \cline{4-5}
        % &
        % &
        % & \textbf{\textit{Scenario.}}
        % Honestly, it feels like my brain never shuts off. I’m worrying from the moment I wake up until I finally crash. It jumps from one thing to another—if the deliveries are late, I start thinking we’ll lose customers, then that the restaurant will go under, then I won’t be able to pay rent, and then I’m imagining my wife leaving me because I can’t provide. It’s always about what could go wrong next, like I’m constantly bracing for some disaster. It’s not just work, either. I worry about my parents’ health, about my wife driving to her job, about whether we’ll ever have enough saved if something happens. Even small things, like a slow night at the restaurant, can set off this whole spiral. My body feels like it’s stuck on “on.” My shoulders and neck are tight all the time, my back and legs ache like I’ve done heavy lifting even on my day off, and I’m exhausted but can’t really relax. This has been going on for at least the past six, seven months, pretty much every day. It’s not like random weird thoughts popping into my head—I’m just constantly, excessively worried about real-life things in the future. \\
        \bottomrule
    \end{tabularx}
\end{table*}

%% file: Texts/Appendix_4ResultandAnalysis.tex
\section{Supplementary Results \& Discussions\label{appen:analysis}}
\subsection{Disease-Specific Performance}
While Figure~\ref{fig:heatmap} displays results for representative models, we present the comprehensive heatmaps for all evaluated models below. The heatmaps for \textit{Type 1} (\autoref{fig:heatmap_type1}) and \textit{Type 2} (\autoref{fig:heatmap_type2}) illustrate accuracy by individual disease, while those for \textit{Type 3} (\autoref{fig:heatmap_type3}) and \textit{Type 4} (\autoref{fig:heatmap_type4}) show accuracy across differential diagnosis tasks.

\input{Figures/Appendix/diff_diag_error_sankey}
\clearpage
\input{Figures/Appendix/Heatmap/Accuracy_Heatmap_Type_I}
\input{Figures/Appendix/Heatmap/Accuracy_Heatmap_Type_II}
\input{Figures/Appendix/Heatmap/Accuracy_Heatmap_Type_III}
\input{Figures/Appendix/Heatmap/Accuracy_Heatmap_Type_IV}
\clearpage

% \section{Supplementary Discussions}

\subsection{\label{appen:metric_analysis}Error Analysis for Types 3 and 4 Using Quantitative Metrics}

This section provides a detailed analysis complementing \autoref{fig:error_analysis}. To strictly evaluate performance in multi-label scenarios (\textit{Types 3 and 4}), we employ Micro-average Precision, Recall, and F1-score.

Let $\mathcal{O}$ denote the set of options for a given problem, and $N$ be the total number of data samples. For the $i$-th sample, we define the set of ground-truth answers as $Y_i$ and the set of model predictions as $\hat{Y}_i$. Unlike macro-averaging, which calculates metrics per class before averaging, the micro-average approach first aggregates True Positives ($TP$), False Positives ($FP$), and False Negatives ($FN$) across the entire dataset, and then calculates Precision, Recall, and F1-score based on these global counts.

The global $TP$, $FP$, and $FN$ are defined as follows, where $\mathbb{I}(\cdot)$ is the indicator function that returns 1 if the condition is true and 0 otherwise:

\begin{equation}
\begin{aligned}
TP &= \sum_{i=1}^{N} \sum_{o \in \mathcal{O}} \mathbb{I}(o \in Y_i \land o \in \hat{Y}_i) \\
FP &= \sum_{i=1}^{N} \sum_{o \in \mathcal{O}} \mathbb{I}(o \notin Y_i \land o \in \hat{Y}_i) \\
FN &= \sum_{i=1}^{N} \sum_{o \in \mathcal{O}} \mathbb{I}(o \in Y_i \land o \notin \hat{Y}_i) 
\end{aligned}
\end{equation}

Based on these aggregated values, Micro-average Precision ($P_{micro}$), Recall ($R_{micro}$), and F1-score ($F1_{micro}$) are calculated as:

\begin{equation}
\begin{aligned}
P_{micro} &= \frac{TP}{TP + FP}, \qquad
R_{micro} &= \frac{TP}{TP + FN}, \qquad
F1_{micro} &= 2 \times \frac{P_{micro} \times R_{micro}}{P_{micro} + R_{micro}}.
\end{aligned}
\end{equation}

% \autoref{tab:type34_F1} presents these metrics for representative models. A consistent trend is observed where Recall remains relatively high, whereas Precision and F1-scores are significantly lower. When interpreted alongside the low exact-match accuracy in \autoref{tab:main_results}, this discrepancy quantifies the fundamental trade-off between diagnostic inclusivity and strictness. While models successfully identify the correct diagnoses (High Recall), they fail to exercise the exclusionary strictness required to rule out distractors (Low Precision). This indicates that the primary bottleneck in current LLMs is not merely a lack of clinical knowledge, but an inability to balance this trade-off, resulting in excessive inclusivity and logical confusion between the target diagnosis and clinically similar conditions.

\input{Tables/Appendix/Analysis/type34_F1}

\paragraph{Type 3.}
\autoref{tab:type34_F1} presents these metrics for representative models.
Open-source models achieve higher recall in the Type 3 setting ($\approx$ 0.79–0.81), whereas closed-source models exhibit substantially lower recall ($\approx$ 0.58–0.69). At the same time, closed models maintain very high precision ($\approx$ 0.94–0.96), indicating that their predictions are typically a strict subset of the correct answers and reflect a highly conservative selection strategy. This imbalance results in systematic underdiagnosis, which is consistent with the elevated underdiagnosis rates observed for closed models in \autoref{fig:error_analysis}. Open models, in contrast, trade precision ($\approx$ 0.82–0.87) for coverage, leading to fewer missed diagnoses but more frequent inclusion of irrelevant candidates.

\paragraph{Type 4.}
In the Type 4 setting, where exactly one diagnosis is correct, both open and closed models maintain consistently high recall ($\approx$ 0.93–0.96), suggesting that the correct diagnosis is almost always included in the response. However, precision diverges sharply: open models show low precision ($\approx$ 0.50–0.57), while closed models achieve higher but still imperfect precision ($\approx$ 0.71–0.82). Given the uniformly high recall, this precision gap implies that models frequently select additional, unnecessary diagnoses even when the task is unambiguous, resulting in systematic overdiagnosis. This pattern aligns with \autoref{fig:error_analysis}, where overdiagnosis errors dominate the error distribution in Type 4 tasks, confirming that the primary failure lies in regulating prediction cardinality rather than identifying the correct diagnosis.

Overall, these findings reveal a calibration failure in diagnostic strictness. The persistent pattern of high recall but low precision shows that models favor inclusive prediction strategies that ensure the correct diagnosis is selected, yet fail to reliably exclude clinically similar distractors. This imbalance reflects an inability to internally regulate the trade-off between inclusivity and exclusion, leading to systematic over-diagnosis in unambiguous cases and under-diagnosis in ambiguous ones. As a result, current LLMs struggle to infer appropriate diagnostic cardinality from evidence alone, limiting their reliability for diagnostic-reasoning evaluation under unconstrained answer cardinality.

\subsection{LLM Data Generation Bias Analysis (Analysis on Self-Recognition Bias)\label{appendix:bias}}
The dataset constructed in this study consists of augmented data generated partitions using the GPT, Gemini, and Qwen model families. Such a generation method may raise concerns regarding ``self-bias,'' where a specific model might recognize its own generated problems more effectively or exhibit relatively higher performance compared to problems generated by other models.

\input{Tables/Appendix/Analysis/self_recognition_bias}

To investigate this possibility, we compare the accuracy rates across different combinations of data generation models and evaluation models.
% 본 연구에서 구축한 데이터셋은 GPT, Gemini, Qwen 계열 모델을 활용하여 분할 생성된 증강 데이터로 구성되어 있으며, 이러한 생성 방식은 특정 모델의 생성 편향(self-bias)에 대한 우려를 야기할 수 있다. 즉, 각 모델이 스스로 생성한 문제를 타 모델이 생성한 문제보다 더 잘 인식하거나, 상대적으로 높은 성능을 보일 가능성이 존재한다.
% 이러한 가능성을 검증하기 위해, 데이터 생성 모델과 평가 모델 간의 정답률을 비교해보았다.
Table \ref{tab:Self_Recognition_Bias} presents the evaluation results of three models (GPT-5.1, Gemini-2.5-Pro, and Qwen3-235) serving as evaluators for the datasets generated by each respective model. The experimental results indicate that all evaluation models recorded consistent performance metrics regardless of the data source. 

% 표 \ref{tab:Self_Recognition_Bias}은 세 가지 모델(GPT-5.1, Gemini-2.5-Pro, Qwen3-235)을 평가 모델로 설정하여 각 모델이 생성한 데이터셋을 평가한 결과를 나타낸다. 실험 결과, 모든 평가 모델은 데이터의 생성 출처와 관계없이 일관된 성능 지표를 기록하였다.
Specifically, GPT-5.1 shows no significant performance discrepancy between its self-generated data (0.3549) and data generated by other models (0.3670, 0.3601). This trend is consistently observed for Gemini-2.5-Pro and Qwen3-235 as well. This implies that the models do not identify the source of the data to produce biased evaluations. Furthermore, it substantiates that the multi-model-based data augmentation technique employed in this study has reduced the likelihood that benchmark performance is driven by generator-specific stylistic artifacts.
% 구체적으로, GPT-5.1d은 자신이 생성한 데이터(0.3549)와 타 모델이 생성한 데이터(0.3670, 0.3601) 사이에서 유의미한 성능 차이를 보이지 않았다. 이러한 경향은 Gemini-2.5-Pro와 Qwen3-235에서도 동일하게 관찰되었다. 이는 각 모델이 데이터의 생성 주체를 식별하여 편향된 평가를 내리지 않음을 의미하며, 본 연구에서 사용된 다종 모델 기반 데이터 증강 기법이 특정 모델에 종속되지 않은 객관적인 데이터 품질을 확보했음을 방증한다.

%\section{Case Examples}

%\input{Tables/Appendix/Case_study/case_type1}

%% file: Figures/Appendix/diff_diag_error_sankey.tex
\begin{figure*}[t]
    \centering
    \includegraphics[width=0.5\linewidth]{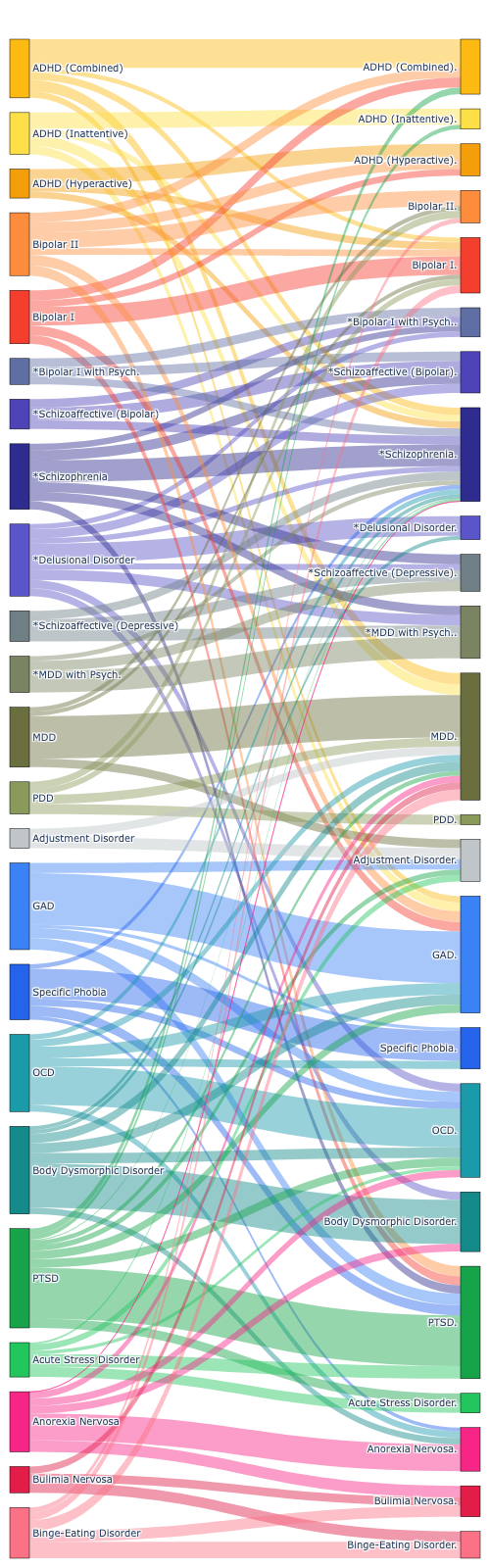}
    \caption{Confusion mapping between ground truth diagnoses (left) and model predictions (right) for \textit{Type 4} (Full version-). * denotes disorders with psychotic symptoms.}
    \label{fig:error_diff_diag_full}
\end{figure*}

%% file: Figures/Appendix/Heatmap/Accuracy_Heatmap_Type_I.tex
\begin{figure*}[h]
    \centering
    \includegraphics[width=\linewidth]{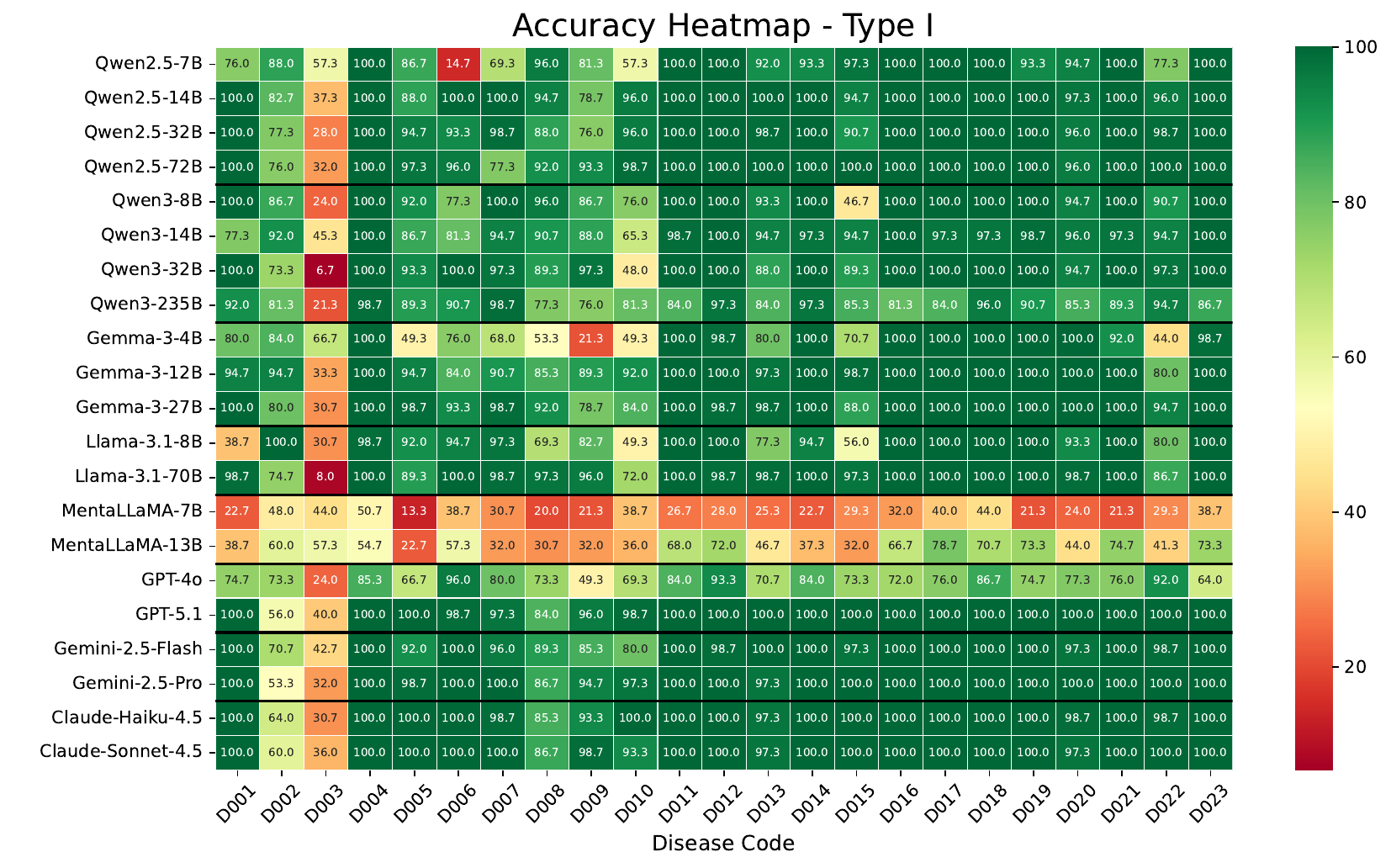}
    \caption{Accuracy Heatmap Type I\label{fig:heatmap_type1}}
    \vspace{-0.2in}
\end{figure*}

%% file: Figures/Appendix/Heatmap/Accuracy_Heatmap_Type_II.tex
\begin{figure*}[h]
    \centering
    \includegraphics[width=\linewidth]{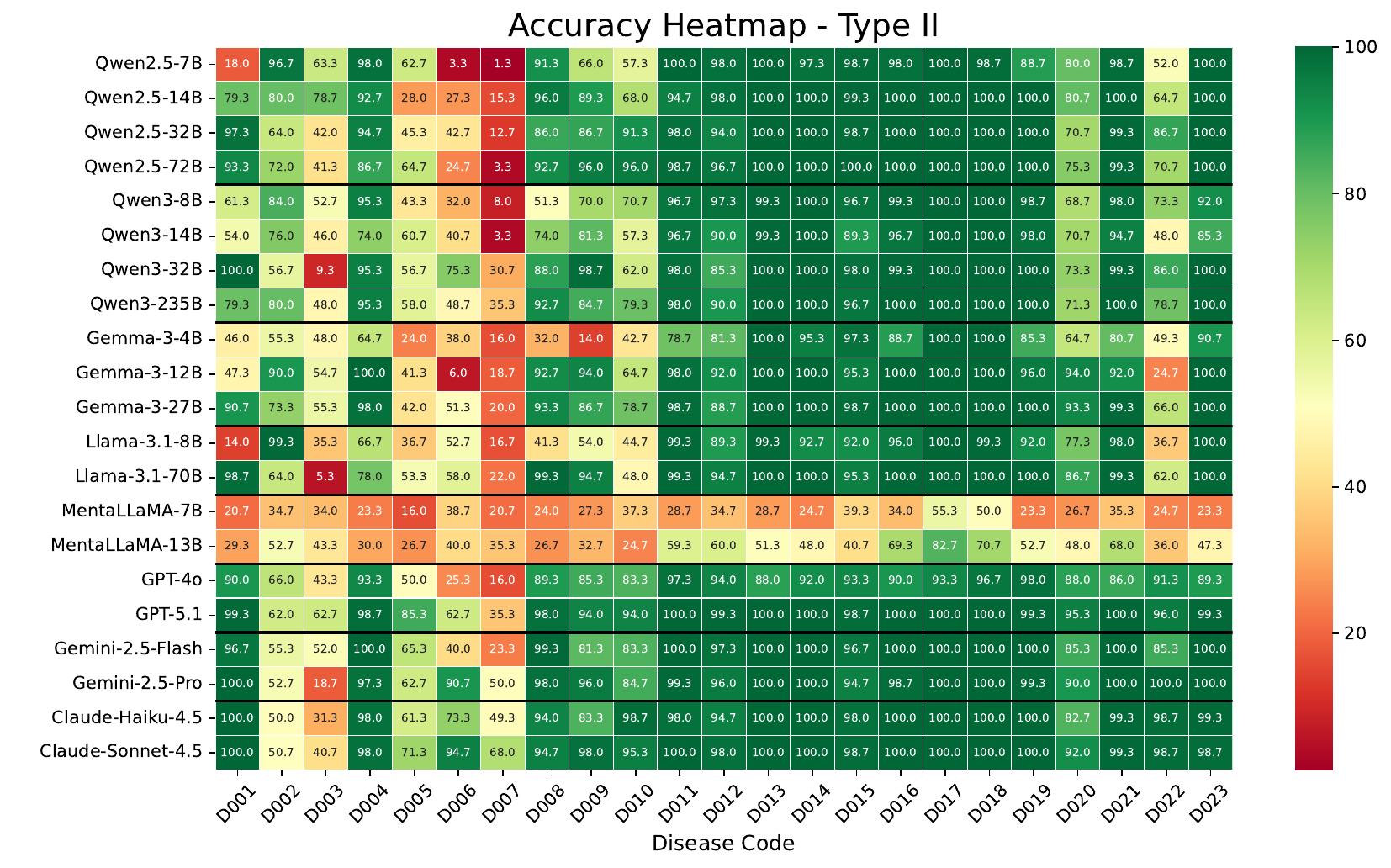}
    \caption{Accuracy Heatmap Type II\label{fig:heatmap_type2}}
    \vspace{-0.2in}
\end{figure*}

%% file: Figures/Appendix/Heatmap/Accuracy_Heatmap_Type_III.tex
\begin{figure*}[h]
    \centering
    \includegraphics[width=\linewidth]{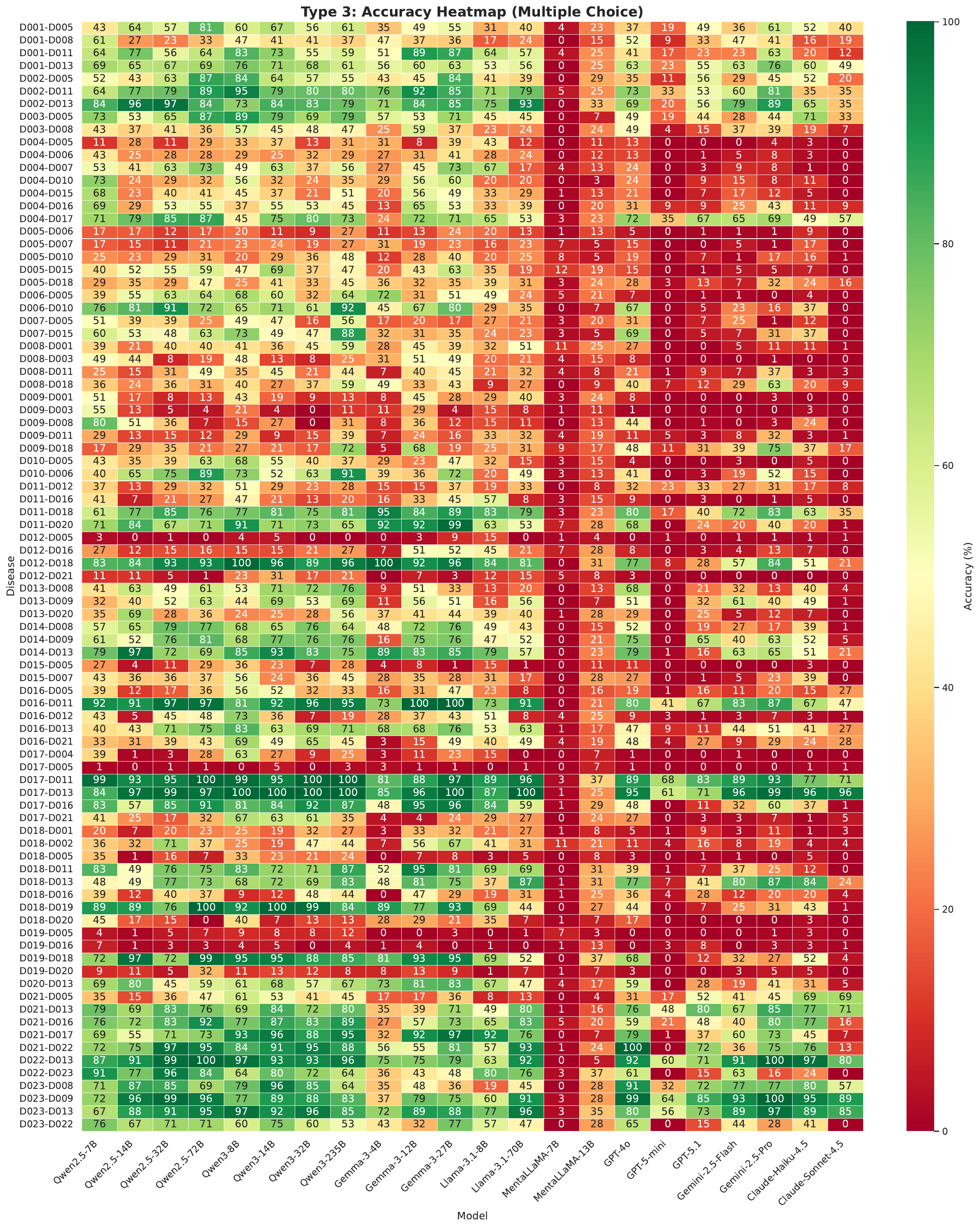}
    \caption{Accuracy Heatmap Type III\label{fig:heatmap_type3}}
    \vspace{-0.2in}
\end{figure*}

%% file: Figures/Appendix/Heatmap/Accuracy_Heatmap_Type_IV.tex
\begin{figure*}[h]
    \centering
    \includegraphics[width=\linewidth]{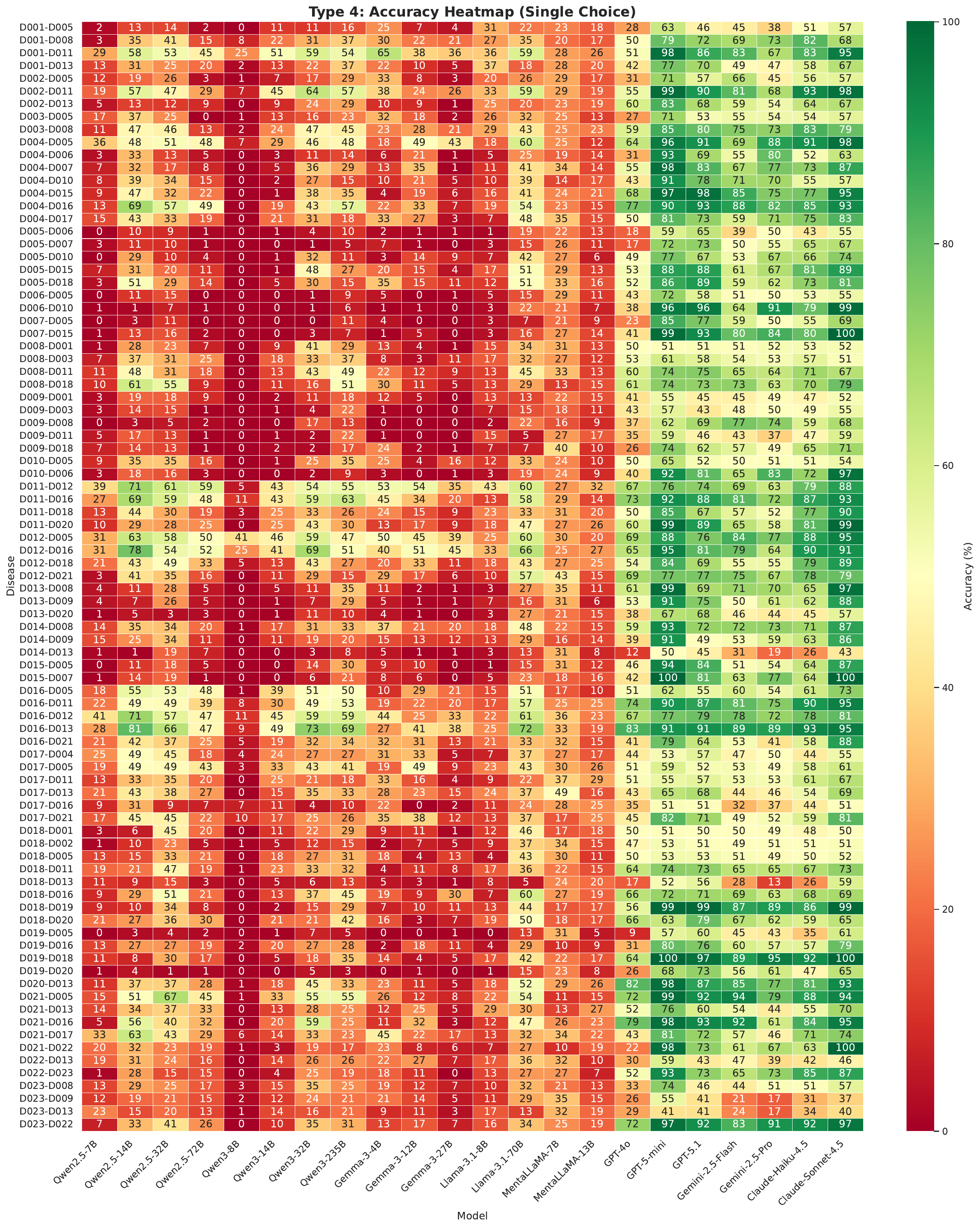}
    \caption{Accuracy Heatmap Type IV\label{fig:heatmap_type4}}
    \vspace{-0.2in}
\end{figure*}

%% file: Tables/Appendix/Analysis/type34_F1.tex
\begin{table}[ht!]
\centering
\caption{\label{tab:type34_F1}Performance comparison of various LLMs on Type 3 and Type 4 datasets using micro-averaged precision, recall, and F1.}
% \resizebox{\columnwidth}{!}{
\begin{tabular}{lcccccc}
\toprule
\multirow{2}{*}{\textbf{Model}} & \multicolumn{3}{c}{\textbf{Type 3}} & \multicolumn{3}{c}{\textbf{Type 4}} \\
\cmidrule(lr){2-4} \cmidrule(lr){5-7}
 & \textbf{Precision} & \textbf{Recall} & \textbf{F1} & \textbf{Precision} & \textbf{Recall} & \textbf{F1} \\
\midrule
Qwen2.5-72B       & 0.8554 & 0.7899 & 0.7939 & 0.5374 & 0.9610 & 0.6780 \\
Qwen3-235B        & 0.8708 & 0.7930 & 0.8133 & 0.5703 & 0.9584 & 0.7059 \\
Gemma3-27B        & 0.8251 & 0.8088 & 0.7914 & 0.5074 & 0.9601 & 0.6531 \\
GPT-5.1           & 0.9389 & 0.6259 & 0.7245 & 0.7749 & 0.9426 & 0.8403 \\
Claude-sonnet-4.5 & 0.9581 & 0.5753 & 0.6901 & 0.8178 & 0.9256 & 0.8550 \\
Gemini-2.5-pro    & 0.9253 & 0.6900 & 0.7676 & 0.7171 & 0.9344 & 0.7992 \\
\bottomrule
\end{tabular}
% }

\end{table}

%% file: Tables/Appendix/Analysis/self_recognition_bias.tex
\begin{table}[h]
\centering
\caption{\label{tab:Self_Recognition_Bias}Cross-evaluation results between different LLMs to check for self-recognition bias. The consistent values across columns for each row indicate minimal source-specific bias.}
% \resizebox{\columnwidth}{!}{
\begin{tabular}{lccc}
\toprule
Model & GPT-5.1& Gemini-2.5-Pro& Qwen3-235B\\
\midrule
GPT-5.1         & 0.3549 & 0.3670 & 0.3601 \\
Gemini-2.5-Pro  & 0.4222 & 0.4370 & 0.4332 \\
Qwen3-235B       & 0.5856 & 0.5972 & 0.5812 \\
\bottomrule
\end{tabular}
% }
\end{table}

%% file: Texts/Appendix_3Experiments.tex
\section{Details for Experimental Setup \label{appen:setup}}
\subsection{Versions of Models\label{appen:models}}
For closed-source models, we employ GPT 5.1~\cite{openai_gpt5_1}, GPT 5-mini~\cite{openai_gpt5_mini}, GPT 4o~\cite{openai2024gpt4ocard}, Gemini 2.5-flash, Gemini 2.5-pro~\cite{comanici2025gemini}, Claude Haiku-4.5~\cite{Anthropic2025Haiku4_5}, and Claude Sonnet-4.5~\cite{Anthropic2025Sonnet4_5}. For open-source models, we employ Qwen 3~\cite{qwen3technicalreport}, Qwen 2.5~\cite{qwen2025qwen25technicalreport}, Gemma 3~\cite{team2025gemma}, Llama-3.1~\cite{grattafiori2024llama}, and MentaLLaMA~\cite{yang2024mentallama}. The specific model versions and their Hugging Face identifiers are listed in \autoref{tab:model_versions}.
\input{Tables/Appendix/model_versions}

\subsection{Experimental Settings \label{appen:settings}}
The inference environments were configured based on the model access methods. The GPT series were accessed via the OpenAI Platform, while the Gemini and Claude series were utilized through the OpenRouter Platform. 
In contrast, all open-weights models, including Qwen3-235B-A22B, were hosted in a self-managed environment. We performed inference using the vLLM library on an AWS p5en.48xlarge instance equipped with 8 NVIDIA H200 GPUs.
For all open-source models, we adopted the default generation configuration provided by each model's respective provider, without any additional tuning of decoding hyperparameters.

\subsection{Details of Instruction Prompts\label{appen:prompts}}
We present the specific prompts used for dataset generation and model evaluation. For dataset generation, the system prompts for Types 1 through 4 are listed in \autoref{tab:system_prompt_type1}--\ref{tab:system_prompt_type4}, respectively. Regarding user prompts, Types 1 and 2 utilize the prompt in \autoref{tab:user_prompt_type12}, while Types 3 and 4 use those in \autoref{tab:user_prompt_type3} and \autoref{tab:user_prompt_type4}. For model evaluation, we assess Types 1 and 2 using the `Single' prompt in \autoref{tab:evaluation_prompt_single}, and Types 3 and 4 using the `Hybrid' prompt (\autoref{tab:evaluation_prompt_hybrid}). We also utilize the `Multiple' prompt (\autoref{tab:evaluation_prompt_multiple}) for analysis.
\clearpage

\input{Tables/Appendix/prompts/system_prompt_type1}

\input{Tables/Appendix/prompts/system_prompt_type2}
\input{Tables/Appendix/prompts/system_prompt_type3}
\input{Tables/Appendix/prompts/system_prompt_type4}

\input{Tables/Appendix/prompts/user_prompt_type12}

\input{Tables/Appendix/prompts/user_prompt_type3}

\input{Tables/Appendix/prompts/user_prompt_type4}

\input{Tables/Appendix/prompts/evaluation_prompt_single}

\input{Tables/Appendix/prompts/evaluation_prompt_hybrid}
\input{Tables/Appendix/prompts/evaluation_prompt_multiple}

%% file: Tables/Appendix/model_versions.tex
% \begin{table}[h]
% \centering
% \resizebox{\columnwidth}{!}{
% \footnotesize
% \begin{tabular}{@{}llp{3.5cm}@{}}
% \toprule
% \textbf{Model} & \textbf{Hugging Face Identifier} \\
% \midrule
% Qwen2.5-7B & \texttt{Qwen/Qwen2.5-7B-Instruct} \\
% Qwen2.5-14B & \texttt{Qwen/Qwen2.5-14B-Instruct} \\
% Qwen2.5-32B & \texttt{Qwen/Qwen2.5-32B-Instruct} \\
% Qwen2.5-72B & \texttt{Qwen/Qwen2.5-72B-Instruct} \\
% \midrule
% Qwen3-8B & \texttt{Qwen/Qwen3-8B} \\
% Qwen3-14B & \texttt{Qwen/Qwen3-14B} \\
% Qwen3-32B & \texttt{Qwen/Qwen3-32B} \\
% Qwen3-235B & \texttt{Qwen/Qwen3-235B-A22B} \\
% \midrule
% Gemma-3-4B & \texttt{google/gemma-3-4b-it} \\
% Gemma-3-12B & \texttt{google/gemma-3-12b-it} \\
% Gemma-3-27B & \texttt{google/gemma-3-27b-it} \\
% \midrule
% Llama-3.1-8B & \texttt{meta-llama/Llama-3.1-8B-Instruct} \\
% Llama-3.1-70B & \texttt{meta-llama/Llama-3.1-70B-Instruct} \\
% \midrule
% MentaLLaMA-7B & \texttt{klyang/MentaLLaMA-chat-7B} \\
% MentaLLaMA-13B & \texttt{klyang/MentaLLaMA-chat-13B} \\
% \bottomrule
% \end{tabular}
% }
% \caption{Model versions used in experiments}
% \label{tab:model_versions}
% \end{table}

\begin{table}[h]
\centering
\caption{\label{tab:model_versions}Model versions used in experiments}
\resizebox{\columnwidth}{!}{
\footnotesize
\begin{tabular}{ll|ll}
\toprule
\textbf{Model} & \textbf{Hugging Face Identifier}
 & \textbf{Model} & \textbf{Hugging Face Identifier}\\
\midrule
Qwen2.5-7B & \texttt{Qwen/Qwen2.5-7B-Instruct} & Gemma-3-4B & \texttt{google/gemma-3-4b-it} \\
Qwen2.5-14B & \texttt{Qwen/Qwen2.5-14B-Instruct} & Gemma-3-12B & \texttt{google/gemma-3-12b-it} \\
Qwen2.5-32B & \texttt{Qwen/Qwen2.5-32B-Instruct} & Gemma-3-27B & \texttt{google/gemma-3-27b-it} \\
Qwen2.5-72B & \texttt{Qwen/Qwen2.5-72B-Instruct} \\
\midrule
Qwen3-8B & \texttt{Qwen/Qwen3-8B} & Llama-3.1-8B & \texttt{meta-llama/Llama-3.1-8B-Instruct} \\
Qwen3-14B & \texttt{Qwen/Qwen3-14B} & Llama-3.1-70B & \texttt{meta-llama/Llama-3.1-70B-Instruct} \\
\cmidrule(lr){3-4}
Qwen3-32B & \texttt{Qwen/Qwen3-32B} & MentaLLaMA-7B & \texttt{klyang/MentaLLaMA-chat-7B} \\
Qwen3-235B & \texttt{Qwen/Qwen3-235B-A22B} & MentaLLaMA-13B & \texttt{klyang/MentaLLaMA-chat-13B} \\
\bottomrule
\end{tabular}
}

\end{table}

%% file: Tables/Appendix/prompts/system_prompt_type1.tex
\begin{table*}[t] 
    \centering
    \scriptsize
    \caption{System Prompt for Type 1 Generation \label{tab:system_prompt_type1}}
    % 너비 기준을 \linewidth 대신 \textwidth로 바꾸는 것을 권장합니다.
    \begin{tcolorbox}[title={System Prompt for Type 1 Generation}, width=0.95\textwidth]
        \textbf{SYSTEM:}
        \begin{lstlisting}[basicstyle=\scriptsize\ttfamily, breaklines=true]
    You are a psychiatrist. 
    After interviewing a patient, document the case presentation in medical chart format.

    # Goal
    Create a professional **Clinical Case Summary** (Third-person perspective) based on the target symptoms and demographics.

    # Guidelines

    **Perspective:**
    - Write in third-person
    - Refer to the patient as "The patient" or "A [Age]-year-old [Gender]"

    **Patient Demographics:**
    - Include age, gender, marital status, occupation in the opening
    - Avoid temporal modifiers (e.g., "recently retired", "newly divorced")
    - IMPORTANT: Provided demographics are NOT stressors or causes of symptoms
    - For potentially stressor-like demographics (e.g., retirement, divorce, bereavement):
        a. Add a statement showing clear temporal separation between the demographic event and symptom onset
        b. The demographic event should have occurred significantly earlier than symptom onset

    **Terminology & Tone:**
    - Use standard medical terminology (avoid patient slang)
    - Maintain an objective, clinical tone

    **Constraints:**
    - Do not reveal the diagnosis name
    - Do not directly quote or copy the provided symptom descriptions

    **Requirements:**
    - Include the duration in clinical format
    - Incorporate all provided symptom categories as clinical observations
    - Keep the summary concise (approximately 150-200 words)

        \end{lstlisting}

    \end{tcolorbox}
    
\end{table*}

%% file: Tables/Appendix/prompts/system_prompt_type2.tex
\begin{table*}[t] 
    \centering
    \scriptsize
    \caption{System Prompt for Type 2 Generation \label{tab:system_prompt_type2}}
    % 너비 기준을 \linewidth 대신 \textwidth로 바꾸는 것을 권장합니다.
    \begin{tcolorbox}[title={System Prompt for Type 2 Generation}, width=0.95\textwidth]
        \textbf{SYSTEM:}
        \begin{lstlisting}[basicstyle=\scriptsize\ttfamily, breaklines=true]
    You are a patient experiencing mental health issues. 
    Describe your symptoms during a clinical interview with a psychiatrist.

    # Goal
    Create a realistic **Patient's Subjective Description** (Vignette) based on the target symptoms and demographics.

    # Guidelines

    **Perspective:**
    - Write in first-person ("I")

    **Demographic & Persona Integration (CRITICAL):**
    - **Adopt the Persona:** Use vocabulary, sentence structure, and concerns appropriate for the provided **Age, Gender, and Occupation**.
    - **Contextualize Symptoms:** Weave the symptoms into the specific lifestyle context.

    **Constraints:**
    - DO NOT explicitly state the diagnosis name
    - DO NOT list demographics robotically; show them through context.
	- DO NOT directly quote or copy the provided symptom descriptions

    **Requirements:**
    - Include the duration naturally in your description
    - Incorporate all provided symptom categories
    - Keep your description concise (approximately 150-200 words)

        \end{lstlisting}

    \end{tcolorbox}
    
\end{table*}

%% file: Tables/Appendix/prompts/system_prompt_type3.tex
\begin{table*}[t] 
    \centering
    \scriptsize
    \caption{System Prompt for Type 3 Generation \label{tab:system_prompt_type3}}
    % 너비 기준을 \linewidth 대신 \textwidth로 바꾸는 것을 권장합니다.
    \begin{tcolorbox}[title={System Prompt for Type 3 Generation}, width=0.95\textwidth]
        \textbf{SYSTEM:}
        \begin{lstlisting}[basicstyle=\scriptsize\ttfamily, breaklines=true]
    You are a patient experiencing mental health issues. 
    Describe your symptoms during a clinical interview with a psychiatrist.

    # Goal
    Create a realistic **Patient's Subjective Description** (Vignette) based on the target symptoms, demographics, and additional condition (if provided).

    # Guidelines

    **Perspective:**
    - Write in first-person ("I")

    **Demographic & Persona Integration (CRITICAL):**
    - **Adopt the Persona:** Use vocabulary, sentence structure, and concerns appropriate for the provided **Age, Gender, and Occupation**.
    - **Contextualize Symptoms:** Weave the symptoms into the specific lifestyle context.

    **Symptom Integration (CRITICAL):**
    - **Core Symptoms:** You MUST include ALL provided symptoms comprehensively in your description.
    - **Additional Condition (If Provided):** 
      - If an additional condition is specified, you MUST also include manifestations of this condition naturally within your description.
      - The additional symptoms should feel concurrent and interwoven with core symptoms, not listed separately.
      - The additional symptoms should NOT dominate; they are co-occurring, not primary.
      - If no additional condition is provided, generate the vignette using ONLY the core symptoms.

    **Constraints:**
    - DO NOT explicitly state the diagnosis name
    - DO NOT list demographics robotically; show them through context.
    - DO NOT directly quote or copy the provided symptom descriptions

    **Requirements:**
    - Include all provided symptoms, ensuring each is represented according to its meaning
    - Include the duration naturally in your description
    - The generated description should be ambiguous enough that both disease could be considered as potential diagnoses
    - Keep your description concise (approximately 150-200 words)

        \end{lstlisting}

    \end{tcolorbox}
    
\end{table*}

%% file: Tables/Appendix/prompts/system_prompt_type4.tex
\begin{table*}[t] 
    \centering
    \scriptsize
    \caption{System Prompt for Type 4 Generation \label{tab:system_prompt_type4}}
    % 너비 기준을 \linewidth 대신 \textwidth로 바꾸는 것을 권장합니다.
    \begin{tcolorbox}[title={System Prompt for Type 4 Generation}, width=0.95\textwidth]
        \textbf{SYSTEM:}
        \begin{lstlisting}[basicstyle=\scriptsize\ttfamily, breaklines=true]
    You are a patient experiencing mental health issues. 
    Describe your symptoms during a clinical interview with a psychiatrist.

    # Goal
    Create a realistic **Patient's Subjective Description** (Vignette) based on the target symptoms, demographics, additional condition (if provided), and provided rule.

    # Guidelines

    **Perspective:**
    - Write in first-person ("I")

    **Demographic & Persona Integration (CRITICAL):**
    - **Adopt the Persona:** Use vocabulary, sentence structure, and concerns appropriate for the provided **Age, Gender, and Occupation**.
    - **Contextualize Symptoms:** Weave the symptoms into the specific lifestyle context.

    **Symptom Integration (CRITICAL):**
    - **Core Symptoms:** You MUST include ALL provided symptoms comprehensively in your description.
    - **Additional Condition (If Provided):** 
      - If an additional condition is specified, you MUST also include manifestations of this condition naturally within your description.
      - The additional symptoms should feel concurrent and interwoven with core symptoms, not listed separately.
      - The additional symptoms should NOT dominate; they are co-occurring, not primary.
      - If no additional condition is provided, generate the vignette using ONLY the core symptoms.
    - **Rule Application (MOST CRITICAL):**
      - You will be provided with a specific rule for the target diagnosis.
      - Your description MUST clearly and unambiguously satisfy this rule.
      - Refer to the provided additional condition and key difference to naturally and explicitly incorporate the rule into your description.

    **Constraints:**
    - DO NOT explicitly state the diagnosis name
    - DO NOT list demographics robotically; show them through context
    - DO NOT directly quote or copy the provided symptom descriptions

    **Requirements:**
    - Include all provided symptoms, ensuring each is represented according to its description
    - Include the duration naturally in your description
    - Explicitly incorporate the provided rule into your description so that it clearly points to the target diagnosis 
    - Keep your description concise (approximately 150-200 words)

        \end{lstlisting}

    \end{tcolorbox}
    
\end{table*}

%% file: Tables/Appendix/prompts/user_prompt_type12.tex
\begin{table*}[t] 
    \centering
    \scriptsize
    \caption{User Prompt for Type 1 \& 2 Generation \label{tab:user_prompt_type12}}
    % 너비 기준을 \linewidth 대신 \textwidth로 바꾸는 것을 권장합니다.
    \begin{tcolorbox}[title={User Prompt for Type 1 \& 2 Generation}, width=0.9\textwidth]
        \textbf{User:}
        \begin{lstlisting}[basicstyle=\scriptsize\ttfamily, breaklines=true]
    **Patient Demographics:**
    - **Age:** {Age}
    - **Gender:** {Gender}
    - **Occupation:** {Occupation}
    - **Marital Status:** {Marital Status}

    **Diagnosis Context:**
    - **Target Disease (Correct Answer):** {Disease Name}

    **Clinical Presentation (Symptoms to Integrate):**
    **1. {Symptom Group Name} Requirements:**
    - {Symptom Name} ({Symptom Description})
    - {Symptom Name} ({Symptom Description})
    Duration of {Symptom Group Name}: {Duration}
    Functional Impairment Due to {Symptom Group Name}: {True/False}
    
    **2. {Symptom Group Name} Requirements:**
    ...

    **Critical Diagnostic Criteria:**
    - Duration: {Total Duration}
    - Functional Impairment: {True/False}
    - Psychosocial Stressor: {True/False}
    - Traumatic Stressor: {True/False}
    - Additional Requirements:
      1) {Requirement 1}
      2) {Requirement 2}
      ...

        \end{lstlisting}

    \end{tcolorbox}
    
\end{table*}

%% file: Tables/Appendix/prompts/user_prompt_type3.tex
\begin{table*}[t] 
    \centering
    \scriptsize
    \caption{User Prompt for Type 3 Generation \label{tab:user_prompt_type3}}
    % 너비 기준을 \linewidth 대신 \textwidth로 바꾸는 것을 권장합니다.
    \begin{tcolorbox}[title={User Prompt for Type 3 Generation}, width=0.9\textwidth]
        \textbf{User:}
        \begin{lstlisting}[basicstyle=\scriptsize\ttfamily, breaklines=true]
    **Patient Demographics:**
    - **Age:** {Age}
    - **Gender:** {Gender}
    - **Occupation:** {Occupation}
    - **Marital Status:** {Marital Status}

    **Diagnosis Context (Ambiguous / Insufficient Evidence):**
    - **Suspected Condition A:** {Disease A Name}
    - **Suspected Condition B:** {Disease B Name}
    - **Additional Condition:** {Additional Condition}
    *Note: The patient presents with symptoms relevant to both but LACKS definitive criteria for either.*

    **Clinical Presentation (Symptoms to Integrate):**
    **1. {Symptom Group Name} Requirements:**
    - {Symptom Name} ({Symptom Description})
    - {Symptom Name} ({Symptom Description})
    Duration of {Symptom Group Name}: {Duration}
    Functional Impairment Due to {Symptom Group Name}: {True/False}
    
    **2. {Symptom Group Name} Requirements:**
    ...

    **Ambiguity Logic Constraints (CRITICAL - The "Void" Logic):**
    - **Goal:** Create a scenario where a definitive diagnosis is IMPOSSIBLE based on the text provided.
    - **Missing Evidence for {Disease A Name}:** {Rule A} -> **MUST be vague or explicitly shorter/absent.**
    - **Missing Evidence for {Disease B Name}:** {Rule B} -> **MUST be vague or unconfirmed.**
    - **Constraint:** DO NOT include any pathognomonic features (determining symptoms) that would confirm A or B. Keep the severity and duration strictly in the "gray zone".

        \end{lstlisting}

    \end{tcolorbox}
    
\end{table*}

%% file: Tables/Appendix/prompts/user_prompt_type4.tex
\begin{table*}[t] 
    \centering
    \scriptsize
    \caption{User Prompt for Type 4 Generation \label{tab:user_prompt_type4}}
    % 너비 기준을 \linewidth 대신 \textwidth로 바꾸는 것을 권장합니다.
    \begin{tcolorbox}[title={User Prompt for Type 4 Generation}, width=0.9\textwidth]
        \textbf{User:}
        \begin{lstlisting}[basicstyle=\scriptsize\ttfamily, breaklines=true]
    **Patient Demographics:**
    - **Age:** {Age}
    - **Gender:** {Gender}
    - **Occupation:** {Occupation}
    - **Marital Status:** {Marital Status}

    **Diagnosis Context:**
    - **Target Disease (Correct Answer):** {Disease A Name}
    - **Differential Diagnosis (Main Distractor):** {Disease B Name}

    **Clinical Presentation (Symptoms to Integrate):**
    **1. {Symptom Group Name} Requirements:**
    - {Symptom Name} ({Symptom Description})
    - {Symptom Name} ({Symptom Description})
    Duration of {Symptom Group Name}: {Duration}
    Functional Impairment Due to {Symptom Group Name}: {True/False}
    
    **2. {Symptom Group Name} Requirements:**
    ...

    **Critical Diagnostic Criteria:**
    - Duration: {Total Duration}
    - Functional Impairment: {True/False}
    - Psychosocial Stressor: {True/False}
    - Traumatic Stressor: {True/False}
    - Additional Requirements:
      1) {Requirement 1}
      2) {Requirement 2}
      ...

    **Differential Logic Constraints (CRITICAL):**
    - **Additional condition:** {Additional Condition}
    - **Key Difference:** {Key Difference}
    - **Rule for Target ({Disease A Name}):** {Rule A}
    - **Rule for Differential ({Disease B Name}):** {Rule B}

        \end{lstlisting}

    \end{tcolorbox}
    
\end{table*}

%% file: Tables/Appendix/prompts/evaluation_prompt_single.tex
\begin{table*}[t] 
    \centering
    \scriptsize
    % 너비 기준을 \linewidth 대신 \textwidth로 바꾸는 것을 권장합니다.
    \caption{Evaluation Prompt for `Single' \label{tab:evaluation_prompt_single}}
    \begin{tcolorbox}[title={Evaluation Prompt for `Single'}, width=0.95\textwidth]
        \textbf{User:}
        \begin{lstlisting}[basicstyle=\scriptsize\ttfamily, breaklines=true]
    {Question Text}

    Q: What is the most likely diagnosis?
    Select the single best answer from the options (A, B, C, D).
    Do NOT provide any explanation, reasoning, or introduction. Just the letter.
    **Output Examples:**
    - Single Answer: A.
    
    {Options List}
    Answer:
        \end{lstlisting}

    \end{tcolorbox}
    
\end{table*}

%% file: Tables/Appendix/prompts/evaluation_prompt_hybrid.tex
\begin{table*}[t] 
    \centering
    \scriptsize
    \caption{Evaluation Prompt for `Hybrid' \label{tab:evaluation_prompt_hybrid}}
    % 너비 기준을 \linewidth 대신 \textwidth로 바꾸는 것을 권장합니다.
    \begin{tcolorbox}[title={Evaluation Prompt for `Hybrid'}, width=0.95\textwidth]
        \textbf{User:}
        \begin{lstlisting}[basicstyle=\scriptsize\ttfamily, breaklines=true]
    {Question Text}

    Select **one or more** applicable answers from the options (A, B, C, D) based on the patient's presentation.
    You must determine whether a **single** diagnosis or **multiple** diagnoses are correct.
    
    **Output Rules:**
    1. If only **one** option is correct, output just the letter (e.g., A).
    2. If **multiple** options are correct, separate them with " & " (e.g., A & B).
    3. Do NOT provide any explanation, reasoning, or introduction. Just the letters.
    
    **Output Examples:**
    - Single Answer: A.
    - Multiple Answers: A & B.
    
    {Options List}
    Answer:
        \end{lstlisting}

    \end{tcolorbox}
    
\end{table*}

%% file: Tables/Appendix/prompts/evaluation_prompt_multiple.tex
\begin{table*}[t] 
    \centering
    \scriptsize
    \caption{Evaluation Prompt for `Multiple' \label{tab:evaluation_prompt_multiple}}
    % 너비 기준을 \linewidth 대신 \textwidth로 바꾸는 것을 권장합니다.
    \begin{tcolorbox}[title={Evaluation Prompt for `Multiple'}, width=0.95\textwidth]
        \textbf{User:}
        \begin{lstlisting}[basicstyle=\scriptsize\ttfamily, breaklines=true]
    {Question Text}

   Q: Which of the following diagnoses are consistent with the patient's presentation? (Select all that apply)
    Do NOT provide any explanation, reasoning, or introduction. Just the letter.
    **Output Examples:**
    - Single Answer: A.
    - Multiple Answers: A & B.
    
    {Options List}
    Answer:
        \end{lstlisting}

    \end{tcolorbox}
    
\end{table*}

%% file: checklist.tex
\section*{NeurIPS Paper Checklist}

\begin{enumerate}

\item {\bf Claims}
    \item[] Question: Do the main claims made in the abstract and introduction accurately reflect the paper's contributions and scope?
    \item[] Answer: \answerYes{} % Replace by \answerYes{}, \answerNo{}, or \answerNA{}.
    \item[] Justification: See Section~\ref{sec:intro}.%The main claims written in Abstract and the contributions in Introduction are aligned.

\item {\bf Limitations}
    \item[] Question: Does the paper discuss the limitations of the work performed by the authors?
    \item[] Answer: \answerYes{} % Replace by \answerYes{}, \answerNo{}, or \answerNA{}.
    \item[] Justification: See Appendix~\ref{limitation}%We include the Limitations section in the main text.

\item {\bf Theory assumptions and proofs}
    \item[] Question: For each theoretical result, does the paper provide the full set of assumptions and a complete (and correct) proof?
    \item[] Answer: \answerNA{} % Replace by \answerYes{}, \answerNo{}, or \answerNA{}.
    \item[] Justification: Our work does not deal with theoretical study.

    \item {\bf Experimental result reproducibility}
    \item[] Question: Does the paper fully disclose all the information needed to reproduce the main experimental results of the paper to the extent that it affects the main claims and/or conclusions of the paper (regardless of whether the code and data are provided or not)?
    \item[] Answer: \answerYes{} % Replace by \answerYes{}, \answerNo{}, or \answerNA{}.
    \item[] Justification: We provide detailed experimental setups (models, GPU configurations, API platforms, and prompts for data generation and evaluation tasks) in \autoref{appen:setup}. Our experimental code is available on GitHub (\url{https://github.com/HoyunS/MentalBench}), and the dataset is available on Hugging Face (\url{https://hf.co/datasets/hysong/MentalBench}).

\item {\bf Open access to data and code}
    \item[] Question: Does the paper provide open access to the data and code, with sufficient instructions to faithfully reproduce the main experimental results, as described in supplemental material?
    \item[] Answer: \answerYes{} % Replace by \answerYes{}, \answerNo{}, or \answerNA{}.
    \item[] Justification: We provide open access to the code via GitHub (\url{https://github.com/HoyunS/MentalBench}). Our benchmark is also publicly available on HuggingFace (\url{https://hf.co/datasets/hysong/MentalBench}).

\item {\bf Experimental setting/details}
    \item[] Question: Does the paper specify all the training and test details (e.g., data splits, hyperparameters, how they were chosen, type of optimizer) necessary to understand the results?
    \item[] Answer: \answerYes{} % Replace by \answerYes{}, \answerNo{}, or \answerNA{}.
    \item[] Justification: Our work does not contain any training process. However, for the evaluation, we provide statistics for our benchmark and the experimental setups used (See \autoref{appen:setup}).

\item {\bf Experiment statistical significance}
    \item[] Question: Does the paper report error bars suitably and correctly defined or other appropriate information about the statistical significance of the experiments?
    \item[] Answer: \answerYes{} % Replace by \answerYes{}, \answerNo{}, or \answerNA{}.
    \item[] Justification: Detailed results are in \autoref{appen:analysis}. We provide results for all models and evaluation types via tables and heatmaps.

\item {\bf Experiments compute resources}
    \item[] Question: For each experiment, does the paper provide sufficient information on the computer resources (type of compute workers, memory, time of execution) needed to reproduce the experiments?
    \item[] Answer: \answerYes{} % Replace by \answerYes{}, \answerNo{}, or \answerNA{}.
    \item[] Justification: We provide detailed experimental setups (GPU configurations and API platforms) in \autoref{appen:setup}.
    
\item {\bf Code of ethics}
    \item[] Question: Does the research conducted in the paper conform, in every respect, with the NeurIPS Code of Ethics \url{https://neurips.cc/public/EthicsGuidelines}?
    \item[] Answer: \answerYes{} % Replace by \answerYes{}, \answerNo{}, or \answerNA{}.
    \item[] Justification: The research adheres to the NeurIPS Code of Ethics.

\item {\bf Broader impacts}
    \item[] Question: Does the paper discuss both potential positive societal impacts and negative societal impacts of the work performed?
    \item[] Answer: \answerYes{} % Replace by \answerYes{}, \answerNo{}, or \answerNA{}.
    \item[] Justification: See Appendix~\ref{ethics} (Ethical Considerations).
    
\item {\bf Safeguards}
    \item[] Question: Does the paper describe safeguards that have been put in place for responsible release of data or models that have a high risk for misuse (e.g., pre-trained language models, image generators, or scraped datasets)?
    \item[] Answer: \answerYes{} % Replace by \answerYes{}, \answerNo{}, or \answerNA{}.
    \item[] Justification: See Appendix~\ref{ethics} (Ethical Considerations).

\item {\bf Licenses for existing assets}
    \item[] Question: Are the creators or original owners of assets (e.g., code, data, models), used in the paper, properly credited and are the license and terms of use explicitly mentioned and properly respected?
    \item[] Answer: \answerYes{} % Replace by \answerYes{}, \answerNo{}, or \answerNA{}.
    \item[] Justification: We cite the original papers and datasets, including the versions used.

\item {\bf New assets}
    \item[] Question: Are new assets introduced in the paper well documented and is the documentation provided alongside the assets?
    \item[] Answer: \answerYes{} % Replace by \answerYes{}, \answerNo{}, or \answerNA{}.
    \item[] Justification: We provide the new assets together with detailed documentation in \autoref{sec:3_knowledge graph} and \autoref{sec:4_mentalbench}.

\item {\bf Crowdsourcing and research with human subjects}
    \item[] Question: For crowdsourcing experiments and research with human subjects, does the paper include the full text of instructions given to participants and screenshots, if applicable, as well as details about compensation (if any)? 
    \item[] Answer: \answerYes{} % Replace by \answerYes{}, \answerNo{}, or \answerNA{}.
    \item[] Justification: We did the human validation with two clinical experts. We include information about expert validation in Appendix \ref{appendix: expert validation}.

\item {\bf Institutional review board (IRB) approvals or equivalent for research with human subjects}
    \item[] Question: Does the paper describe potential risks incurred by study participants, whether such risks were disclosed to the subjects, and whether Institutional Review Board (IRB) approvals (or an equivalent approval/review based on the requirements of your country or institution) were obtained?
    \item[] Answer: \answerNA{} % Replace by \answerYes{}, \answerNo{}, or \answerNA{}.
    \item[] Justification: We did not do crowdsourcing, but validated with expert authors. For this reason, we do not obtain IRB approval.

\item {\bf Declaration of LLM usage}
    \item[] Question: Does the paper describe the usage of LLMs if it is an important, original, or non-standard component of the core methods in this research? Note that if the LLM is used only for writing, editing, or formatting purposes and does \emph{not} impact the core methodology, scientific rigor, or originality of the research, declaration is not required.
    %this research? 
    \item[] Answer: \answerYes{} % Replace by \answerYes{}, \answerNo{}, or \answerNA{}.
    \item[] Justification: We create our benchmark via LLM synthesis. Therefore, we describe the use of LLMs in our paper. The used models are in Appendix~\ref{appen:models}.

\end{enumerate}